\documentclass[11pt]{article}

\usepackage[preprint]{acl}
\usepackage{times}
\usepackage{latexsym}

\usepackage[T1]{fontenc}

\usepackage[utf8]{inputenc}

\usepackage{microtype}

\usepackage{inconsolata}

\usepackage{graphicx}

\usepackage{makecell} 
\usepackage{multirow}
\usepackage{tabularx}
\usepackage{caption} 
\usepackage{amsmath}
\usepackage{array}
\usepackage[table]{xcolor}
\usepackage{float}  
\usepackage{cuted}

\usepackage{booktabs}

\usepackage{url}
\usepackage{needspace} 
\usepackage{placeins}

\usepackage{rotating} 

\usepackage{xstring}

\usepackage{tikz}
\usetikzlibrary{arrows.meta}

\makeatletter
\let\old@autoref\autoref

\protected\def\autoref#1{%
  \IfBeginWith{#1}{app_}{%
    \hyperref[#1]{Appendix~\ref*{#1}}%
  }{%
    \old@autoref{#1}%
  }%
}
\makeatother

\providecolor{englishblue}{HTML}{1F77B4}
\providecolor{multiorange}{HTML}{FF7F0E}
\definecolor{englishbluebg}{HTML}{E4EFF6} 
\definecolor{multiorangebg}{HTML}{FFDFC3}

 \usepackage[most]{tcolorbox}
\newtcolorbox{promptbox}[1]{%
  colback=gray!5,
  colframe=black!60,
  fonttitle=\bfseries,
  title=#1,
  breakable,
  enhanced,
  sharp corners,
  boxrule=0.5pt,
  left=4pt, right=4pt, top=4pt, bottom=4pt,
}

\tcbset{
colback=gray!5,
colframe=black!50,
boxrule=0.5pt,
arc=2pt,
outer arc=2pt,
left=6pt,
right=6pt,
top=6pt,
bottom=6pt
}

\newcommand{\green}[1]{\textcolor{green!50!black}{#1}}
\newcommand{\red}[1]{\textcolor{red!70!black}{#1}}

\title{The Harder Text Embedding Benchmark (HTEB):\\ Beyond One-dimensional Static Robustness}

\author{
Manuel Frank \\
Department of Computer Science \\
Munster Technological University \\
\texttt{Manuel.Frank@zohomail.eu} \\\And
Haithem Afli \\
Department of Computer Science \\
Munster Technological University \\
\texttt{Haithem.Afli@mtu.ie} \\
}

\begin{document}
\maketitle
\begin{abstract}
    Embedding benchmarks like MTEB report a single score per model, implicitly treating robustness as a static, scalar property. We argue that embedding robustness is multidimensional, since models respond differently to different types of variation, and requires dynamic evaluation to expose failures hidden by static benchmarks. We introduce the Harder Text Embedding Benchmark (HTEB), a dynamic evaluation framework that challenges model robustness along three practically interpretable axes (Lexical/Stylistic, Length and Language) by stochastically transforming inputs at evaluation time with an LLM. Evaluating 16 open-weight embedding models on 32 datasets covering 42 languages under transformations validated by 4,800 individual human ratings on an English subsample, supplemented by a Spanish-source evaluation and an exploratory STS-B study of pair-level label preservation, we find three patterns: (1) Models exhibit specific, partly decoupled robustness profiles across axes. (2) Across three model families, scale increases absolute scores but does not close the gap between original and transformed evaluations. Here, scaling tends to improve specifically the Language axis. (3) English datasets are more sensitive to HTEB transformations than multilingual datasets. This demonstrates that HTEB identifies strengths and weaknesses of models along deployment-relevant axes, challenging current embedding benchmarks and arguing for multidimensional, dynamic robustness evaluation. We make the code to run HTEB publicly available\footnote{\url{https://github.com/ThisIsManuel/HTEB}}.
\end{abstract}
%

\section{Introduction}
Text embeddings are commonly evaluated with benchmarks such as MTEB~\citep{Muennighoff2023_MTEB_EACL} and MMTEB~\citep{Enevoldsen2024_MMTEB_ICLR}.
While benchmark limitations have a long history in NLP~\citep{Dhar2026_EvaluationRevisited}, recent work raises specific concerns about the robustness and generalisability of static embedding benchmarks~\citep{Assadi2025_HUME_ICLR, Goel2025_SAGE_NeurIPS, frank-afli-2026-pteb}; we argue that embedding robustness is multidimensional and requires dynamic evaluation.

\begin{figure}
    \centering
    \includegraphics[width=1\linewidth]{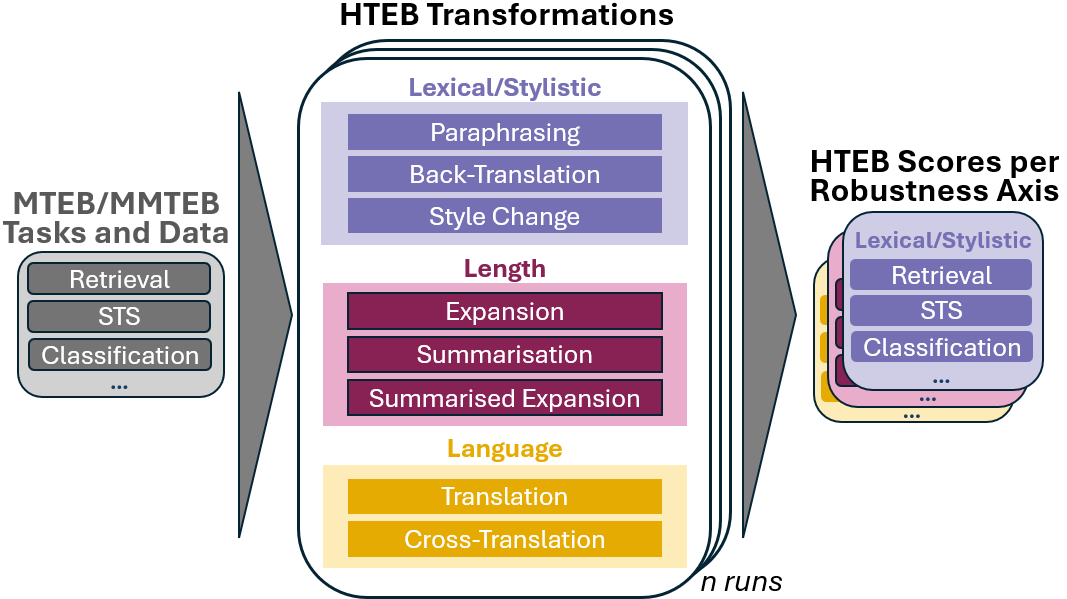}
    \caption{(M)MTEB datasets are transformed over n runs using eight transformations that generate lexical-/stylistic-, length- or language-related variations.}
    \label{fig:hteb_approach}
\end{figure}
\begin{figure*}
    \centering
    \includegraphics[width=0.8\linewidth]{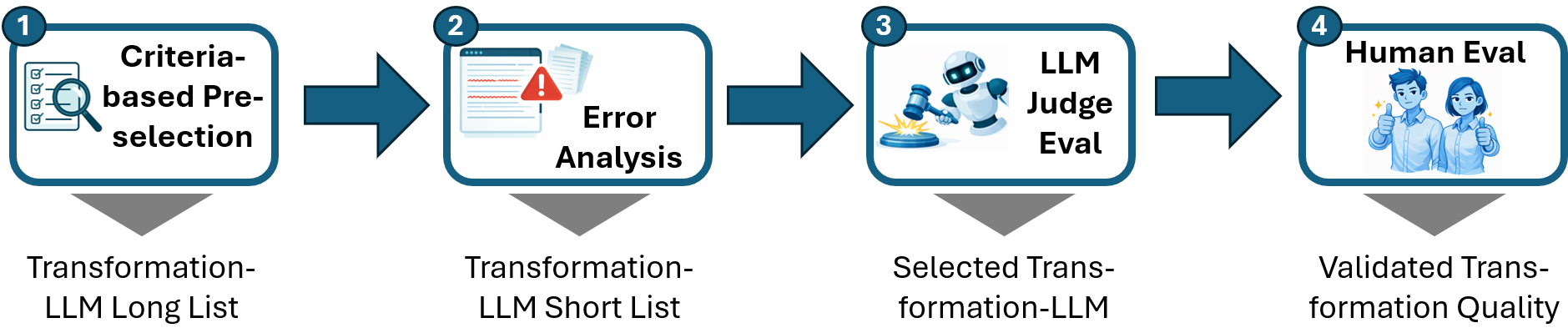}
        \caption{Four-step method to select the HTEB transformation model.}
    \label{fig:HTEB_method}
\end{figure*}

Recent work has begun to address these limitations. SAGE~\citep{Goel2025_SAGE_NeurIPS} tests robustness to controlled perturbations and noise injections on English datasets. PTEB~\citep{frank-afli-2026-pteb} stochastically paraphrases 20 MTEB/MMTEB datasets including 25 languages at evaluation time, revealing degradation even in state-of-the-art embedding models. FLUKE~\citep{otmakhova-etal-2026-fluke} evaluates the downstream robustness through minimal linguistic modifications for the English language to find task-specific brittleness. Building on this line of work, we argue that PTEB's dynamic, LLM-based evaluation-time paradigm is a promising direction, but that robustness evaluation must be broadened beyond paraphrasing and structured along interpretable axes: aggregate robustness scores can still obscure important model differences since robustness is multidimensional. We therefore introduce the Harder Text Embedding Benchmark (HTEB), which generates transformations along three deployment-oriented axes: Lexical/Stylistic, Length, and Language (\autoref{fig:hteb_approach}). Importantly, HTEB's axes are deployment-oriented rather than universal robustness requirements. In particular, the Language axis measures sensitivity to changes in the language used at deployment.

In a concurrent work, CS-MTEB~\cite{Zeng2026_CS-MTEB} evaluates embedding model robustness to LLM-generated code-switching across multiple MTEB datasets. CS-MTEB follows the same evaluation method as the earlier PTEB protocol, replacing PTEB's paraphrasing with code-switching. At a broader methodological level, \citet{Afli2026_MD} develop a distribution-based evaluation protocol for NLP based on HTEB and related work. HTEB is aligned with this but focuses on the embedding setting: it provides a concrete empirical benchmark with deployment-motivated transformations, human validation, repeated runs, and multi-dimensional robustness analyses.

In summary, this work makes two main contributions: (1) We introduce HTEB, a dynamic benchmark that applies eight LLM-generated transformations along three axes, validated by 4,800 individual human ratings based on HTEB's English datasets supported by a Spanish-source evaluation and an exploratory STS-B study of pair-level label preservation. (2) Evaluating 16 embedding models on 32 datasets covering 42 languages, we show scalar rankings hide model-specific, partly decoupled robustness profiles, and scaling changes axis-specific robustness profiles but does not generally improve robustness in three model families. These findings challenge single-score embedding rankings and motivate per-axis robustness profiles for benchmarks. 

\section{Method}%
\label{sec:method}

\subsection{Models}
\label{sec:method_models}
 
\textbf{Generative Models.} The selection of generative models follows the process in \autoref{fig:HTEB_method}. We select the LLM used to generate eight transformations by filtering open-weight vLLM-compatible candidates suitable for consumer-grade hardware (see hardware specification in \autoref{app_hardware}), measuring transformation error rates on STS-B~\citep{SemEval2017_Task1}, and comparing the five lowest-error models with a crossover LLM-judge setup. The transformations are briefly described in \autoref{tab:transformations}. Prompts are shown in \autoref{app_prompts}; an example for each transformation in \autoref{tab:transformation_examples} of \autoref{app_examples}. \\

\noindent\textbf{Embedding Models.} We evaluate open-weight SentenceTransformers-compatible~\citep{SBERT_Reimers2019, Reimers2020_multilingual-sbert} embedding models spanning model size, MTEB performance, and language coverage. All original and transformed scores are computed with our own evaluation pipeline, so scores may differ from public MTEB scores. Although we do not advocate limiting the evaluation to a single score, we still report total HTEB scores as an average over the axis scores.

\subsection{Tasks, Datasets and Metrics}

We evaluate 11 models on 19 English datasets across eight tasks and 9 models on 13 multilingual datasets across seven tasks\footnote{ Instead of a third Semantic Textual Similarity (STS) dataset, we include a Semantic Textual Relatedness (STR) dataset to capture broader relatedness~\citep{abdallaWhatMakesSentences2023a}.}, with four models evaluated in both settings. The multilingual subset covers 42 languages, including low-resource African languages alongside mid- and high-resource languages. See \autoref{tab:english_datasets} and \autoref{tab:multilingual_datasets} in \autoref{app_datasets} for details regarding the datasets and \autoref{app_metrics} for the metrics applied.

\subsection{Statistical Analysis}

We apply non-parametric inferential statistics throughout this work, as normality cannot be assumed for NLP evaluations~\citep{sogaardEstimatingEffectSize2013,Dror2018_HitchhikersGuide}. For LLM judge score comparison during transformation model selection, we use bootstrapping~\citep{Efron1994_Bootstrap}, with observations defined as sentence-pair scores averaged across judges and transformations. 

For the embedding model evaluation, we conduct three runs per model using different random seeds to mitigate single-run unreliability~\citep{Reimers2017_ReportingScoreDistributions, Reimers2018, Madaan2024_QuantifyingVariance}. Each run is a seeded execution of the HTEB transformation pipeline including selection of target or intermediate languages where applicable. For all reported results and inferential tests, we average scores over runs because runs are stochastically dependent repeated evaluations of the same dataset-model-transformation combination and therefore should not be treated as independent observations. Using per-dataset averages as observations, we test per-transformation dataset-level paired score deltas using the Wilcoxon signed-rank tests~\citep{Wilcoxon1945} and Hodges--Lehmann shifts with 95\% CIs. We use the Wilcoxon signed-rank test due to bootstrap instability at smaller sample sizes~\citep{Sogaard2014_WhatsPvalueNLP, hesterbergWhatTeachersShould2015}. 

The p-values for the selection of the transformation model and the evaluation of the embedding models are Holm-corrected~\citep{Holm1979} to control for multiple comparisons, an important problem when evaluating models on multiple datasets~\citep{Dror2017_Replicability-Analysis}. An overview of the main statistical analyses is provided in \autoref{tab:stats_overview} in \autoref{app_statistical_analysis}.
 
\begin{table}[t]
    \small
    \centering
    \begin{tabular}{@{}p{2.58cm}p{4.9cm}@{}}
        \toprule
        \textbf{Transformation} & \textbf{Description} \\
        \midrule
        \textbf{Lexical/Stylistic} & \\
        \quad Paraphrasing      & Rephrase preserving meaning \\
        \quad Backtranslation    & Translate to random language and back \\
        \quad Style Change       & Convert formal to informal language and vice versa\\
        \midrule
        \textbf{Length} & \\
        \quad Expansion          & Elaborate preserving core meaning \\
        \quad Summarisation      & Compress preserving core meaning \\
        \quad Summ.\ Expansion   & Expand then summarise  \\
        \midrule
        \textbf{Language} & \\
        \quad Translation        & Translate each text to the same random target language \\
        \quad Cross-Translation  & Translate each text to independent random languages \\
        \bottomrule
    \end{tabular}
    \caption{HTEB's eight transformations grouped by robustness dimension.}
    \label{tab:transformations}
\end{table}

\begin{figure*}
    \centering
    \includegraphics[width=1\linewidth]{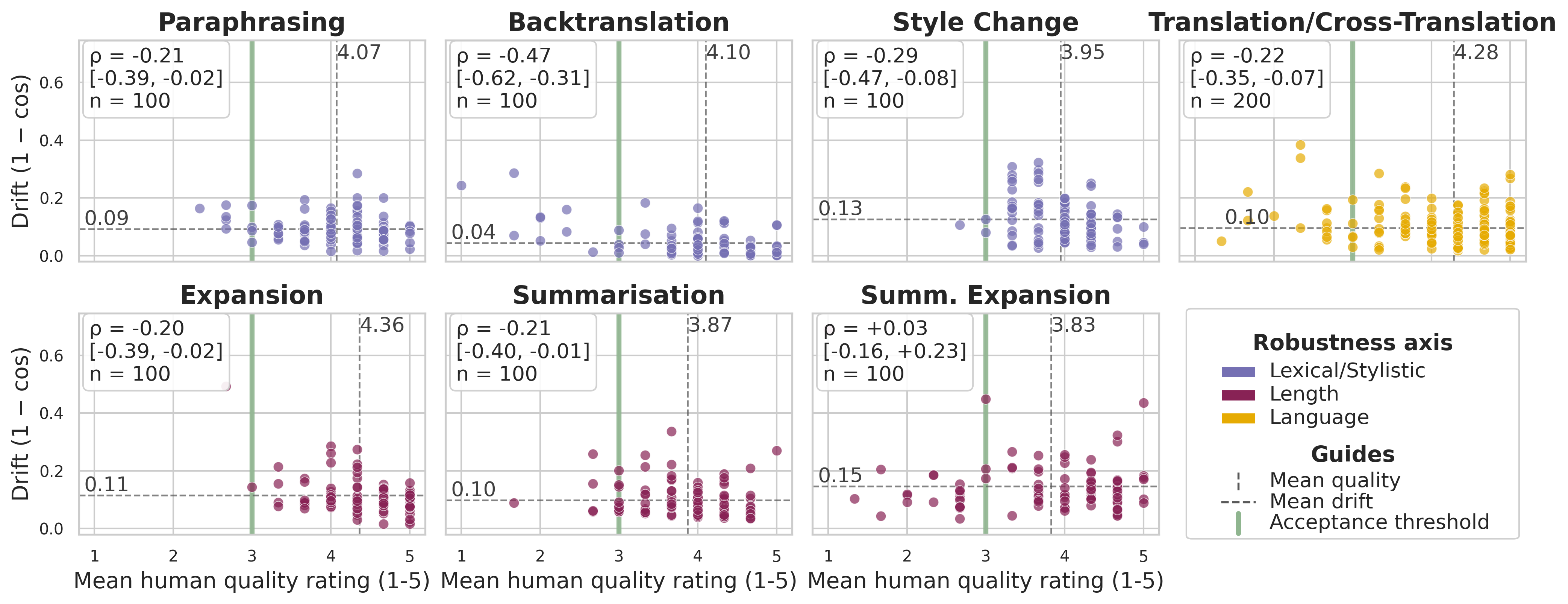}
    \caption{Per-item embedding drift versus mean human transformation quality rating, by transformation. Each dot is one text pair (original, transformed) from the human-evaluation study; mean quality (1–5 Likert) is averaged across raters, drift is $1 - \cos(emb(original), emb(transformed))$ under Jina-Embeddings-v5-Small. Each panel denotes $\rho$ as Spearman's $\rho$ with its 95\% bootstrap CI and the number of sentence pairs n.}
    \label{fig:quality_vs_drift}
\end{figure*}

\subsection{Human Evaluation}

We only include transformations that receive sufficiently high ratings from human annotators. Specifically, we set the acceptance threshold at a mean score of~3 on a 1-to-5 Likert scale, corresponding to "acceptable". This helps reduce the risk that observed performance drops are primarily driven by semantic distortion introduced by the transformations. All ratings are collected along two criteria:

\textit{1. Transformation quality:} The degree to which the output performs the intended transformation while preserving the meaning of the original text.\footnote{For length-increasing transformations, meaning preservation is interpreted as preservation of the original core meaning.}

\textit{2. Fluency:} The degree to which a text reads as well-formed~\cite{White1994_ARPA_MT_Evaluation}.\\

For each of the eight transformations, three annotators evaluate 100 text pairs, yielding $2 \times 8 \times 100 \times 3 = 4{,}800$ human ratings in total. The human-evaluation sample is randomly drawn across HTEB's
English datasets. Thus, human ratings directly validate the transformations applied to English input, including translation-based transformations. We measure inter-rater agreement using Gwet's AC$_2$~\citep{Gwet2008} with ordinal weights since, unlike Krippendorff's $\alpha$, it remains stable when ratings cluster closely around a single value. \autoref{fig:google_form} in \autoref{app_human_eval} shows an example of the Google Forms used to collect the data. 

Translation and Cross-Translation are evaluated jointly (together 200 samples) since the annotation unit is an individual sentence and its transformation, not a task-specific text pair. As an exploratory analysis, we include an additional human evaluation for Spanish as a non-English source language. 

Moreover, since transformation quality and fluency only provide indirect evidence for label preservation, we conduct an exploratory human validation study using the STS-B dataset and one transformation per axis: paraphrasing, summarised expansion and translation. Based on this, we compare the similarity ratings assigned to original and transformed sentences and analyse the differences between them. See \autoref{app_human_eval} for more details.

\section{Experimental Setup}%

\paragraph{Generative Models.} Following \autoref{fig:HTEB_method}, we
consider 16 open generative LLMs from the Apriel-1.5/-1.6, Gemma-3~\citep{gemma3}, Ministral-3~\citep{mistral3}, Olmo-3~\citep{Olmo3_2025}, and Qwen3~\citep{qwen3} families (see \autoref{app_gen_models} for a complete list of models). All transformation prompts are listed in \autoref{app_prompts}.
\paragraph{Embedding Models.} Following the criteria described, we evaluate 11 models on 19 English datasets across eight tasks and 9 models on 13 multilingual datasets across seven tasks, with 4 models evaluated in both settings (16 unique models). See \autoref{tab:embedding_models} in \autoref{app_emb_models} for the full list. Separately, we conduct a scale ablation over 14 checkpoints from three model families including three checkpoints that also appear in the main benchmark.

\section{Experimental Results}
This section presents our empirical results. We first describe the transformation model selection results, comprising error analysis, LLM evaluation, and human evaluation (see \autoref{fig:HTEB_method}). Using the selected transformation model, we then evaluate embedding models on HTEB.
\subsection{Transformation Model Selection }

\subsubsection{Error Analysis and LLM Judge Eval}

Gemma-3-27B-int4-AWQ (3.0\%), Qwen3-8B-AWQ (5.8\%), Ministral-3-14B-Instruct-2512 (6.5\%), Olmo-3-7B-Instruct (7.0\%), and Qwen3-8B (7.1\%) have the lowest total error rates (\autoref{tab:error_rates} in \autoref{app_err_rate}). Notably, 4-bit activation-aware weight quantisation (AWQ)~\citep{Lin20204_AWQ} reduces error rates for Qwen3-8B but not the 4B model, and all Qwen3 models with 4B or 8B parameters outperform Qwen3-14B-AWQ. See \autoref{app_err_rate} and \autoref{tab:error_rates} for a breakdown of the 10 error types.

Using each of the five best models as both a transformation model and LLM judge, we found that Gemma-3-27B-int4-AWQ received the highest scores across LLM judges, with the pairwise differences against all models reaching statistical significance (Holm-corrected bootstrap p-values $<0.001$; see \autoref{tab:llm_judge_scores} in \autoref{app_llm_judge_eval}). Gemma-3-27B-int4-AWQ is therefore used as HTEB's main transformation model. 

\begin{table}[t]
\centering
\small
\begin{tabular}{lcccc}
\toprule
 & \multicolumn{2}{c}{\textbf{Quality}} & \multicolumn{2}{c}{\textbf{Fluency}} \\
\cmidrule(lr){2-3} \cmidrule(lr){4-5}
\textbf{Transformation} & \textbf{Mean} & \textbf{AC$_2$} & \textbf{Mean} & \textbf{AC$_2$} \\
\midrule
\textbf{Lexical/Stylistic} & \textbf{4.04{\scriptsize $\pm$1.04}} & \textbf{0.62} & \textbf{4.32{\scriptsize $\pm$0.87}} & \textbf{0.71} \\
\quad Paraphrasing & 4.07{\scriptsize $\pm$1.07} & 0.57 & 4.04{\scriptsize $\pm$0.90} & 0.57 \\
\quad Backtranslation & 4.10{\scriptsize $\pm$1.16} & 0.68 & 4.40{\scriptsize $\pm$0.99} & 0.69 \\
\quad Style Change & 3.95{\scriptsize $\pm$0.86} & 0.67 & 4.52{\scriptsize $\pm$0.63} & 0.88 \\
\midrule
\textbf{Length} & \textbf{4.02{\scriptsize $\pm$1.04}} & \textbf{0.67} & \textbf{4.66{\scriptsize $\pm$0.66}} & \textbf{0.92} \\
\quad Expansion & 4.36{\scriptsize $\pm$0.79} & 0.81 & 4.82{\scriptsize $\pm$0.39} & 0.95 \\
\quad Summarisation & 3.87{\scriptsize $\pm$0.96} & 0.67 & 4.20{\scriptsize $\pm$0.89} & 0.71 \\
\quad Summ.\ Expan. & 3.83{\scriptsize $\pm$1.24} & 0.56 & 4.97{\scriptsize $\pm$0.18} & 1.00 \\
\midrule
\textbf{Language} & \textbf{4.28{\scriptsize $\pm$1.03}} & \textbf{0.77} & \textbf{4.20{\scriptsize $\pm$1.00}} & \textbf{0.72} \\
\quad (Cross-)Transl. & 4.28{\scriptsize $\pm$1.03} & 0.77 & 4.20{\scriptsize $\pm$1.00} & 0.72 \\
\midrule
\textbf{Overall} & \textbf{4.09{\scriptsize $\pm$1.04}} & \textbf{0.68} & \textbf{4.42{\scriptsize $\pm$0.86}} & \textbf{0.80} \\
\bottomrule
\end{tabular}
\caption{Mean human ratings (with standard dev. (SD)) and inter-annotator agreement per transformation.}
\label{tab:human_eval_per_language}
\end{table}

\subsubsection{Human Validation}

\paragraph{Quality and Fluency.}

All transformations exceed the acceptance threshold for quality and fluency, with a higher mean rating for fluency (\autoref{tab:human_eval_per_language}). At the axis level, the Language transformations receive the highest quality ratings (mean 4.28). At the transformation level, Expansion has the highest quality score (mean 4.36), indicating that annotators judged it to preserve the original core meaning despite substantial additions. We generally observe moderate-to-high inter-rater agreement, with the lowest agreement for Paraphrasing on Fluency and for Summarised Expansion and Paraphrasing on Quality but essentially perfect agreement on the Fluency of Summarised Expansion.

~\autoref{fig:quality_vs_drift} reveals three distinct patterns in how embedding drift relates to human quality ratings. Most transformations cluster around $\rho \approx -0.20$, a weak negative correlation indicating that human-perceived quality explains only a small share of the variance in embedding drift: embeddings shift substantially even for transformations annotators rate as faithful.
Backtranslation shows stronger correlation ($\rho = -0.47$), consistent with the finding that 5.51\% of backtranslations were identical with the input (compared to only 0.75\% across all transformations). Summarised Expansion is an outlier in the other direction ($\rho = +0.03$), with drift and quality effectively uncorrelated. This indicates that HTEB's measured performance drops are not primarily artefacts of transformation quality.

To test whether the transformation validation generalises beyond English-source inputs, we repeated the quality and fluency evaluation using Spanish as the source language (see \autoref{tab:spanish_human_evaluation}). All eight transformations exceeded the pre-specified mean threshold of 3, with overall means of 4.03 for quality and 4.34 for fluency. Agreement was generally moderate to high, although quality agreement for Summarised Expansion was low ($AC_2 = 0.20$).

\begin{table}[t]
\centering
\small
\begin{tabular}{lcccc}
\toprule
 & \multicolumn{2}{c}{\textbf{Quality}} & \multicolumn{2}{c}{\textbf{Fluency}} \\
\cmidrule(lr){2-3} \cmidrule(lr){4-5}
\textbf{Transformation} & \textbf{Mean} & \textbf{AC$_2$} & \textbf{Mean} & \textbf{AC$_2$} \\
\midrule
\textbf{Lexical/Stylistic} & \textbf{3.79} & \textbf{0.58} & \textbf{4.09} & \textbf{0.61} \\
\quad Paraphrasing & 4.10 & 0.67 & 4.20 & 0.56 \\
\quad Backtranslation & 3.63 & 0.40 & 4.10 & 0.58 \\
\quad Style Change & 3.63 & 0.74 & 3.98 & 0.74 \\
\midrule
\textbf{Length} & \textbf{3.98} & \textbf{0.45} & \textbf{4.51} & \textbf{0.85} \\
\quad Expansion & 4.07 & 0.49 & 4.30 & 0.75 \\
\quad Summarisation & 3.97 & 0.68 & 4.57 & 0.87 \\
\quad Summ.\ Expan. & 3.90 & 0.20 & 4.65 & 0.90 \\
\midrule
\textbf{Language} & \textbf{4.24} & \textbf{0.79} & \textbf{4.40} & \textbf{0.82} \\
\quad (Cross-)Transl. & 4.24 & 0.79 & 4.40 & 0.82 \\
\midrule
\textbf{Overall} & \textbf{4.03} & \textbf{0.63} & \textbf{4.34} & \textbf{0.77} \\
\bottomrule
\end{tabular}
\caption{Mean Spanish source-language human ratings and inter-annotator agreement per transformation.}
\label{tab:spanish_human_evaluation}
\end{table}

\paragraph{Label Preservation.}

\autoref{tab:label_preservation} shows the results of the exploratory analysis of human similarity ratings for original and transformed STS-B text pairs on the STS-B-specific 0-to-5 similarity scale. Mean similarity ratings for original and transformed pairs are 2.36 and 2.28 for Paraphrasing, 2.81 and 2.75 for Summarised Expansion and 2.49 and 2.39 for Translation (four target languages). The corresponding mean absolute differences are 0.65, 0.41, and 0.45 with 88\%, 96\%, and 90\% of paired ratings differing by no more than one point.

\begin{table}[t]
    \centering
    \small
    \setlength{\tabcolsep}{3.5pt}
    \begin{tabular}{lccccc}
    \toprule
    \textbf{Transformation} & \textbf{Orig.} & \textbf{Trans.} &
    \textbf{Gold} & \textbf{MAD} & $\boldsymbol{\leq 1}$ \\
    \midrule
    Paraphrasing & 2.36 & 2.28 & 2.37 & 0.65 & 88\% \\
    Summ.\ Expansion & 2.81 & 2.75 & 2.37 & 0.41 & 96\% \\
    Translation & 2.49 & 2.39 & 2.38 & 0.45 & 90\% \\
    \bottomrule
    \end{tabular}
    \caption{Mean human similarity ratings for transformed and original STS-B sentence pairs and corresponding gold labels. MAD is the mean absolute difference between paired transformed and original ratings; $\leq 1$ is the corresponding percentage of ratings differing by no more than one point. Translation includes Arabic, French, German, and Spanish as target languages.}
    \label{tab:label_preservation}
\end{table}

\begin{figure*}
    \centering
    \includegraphics[width=1\linewidth]{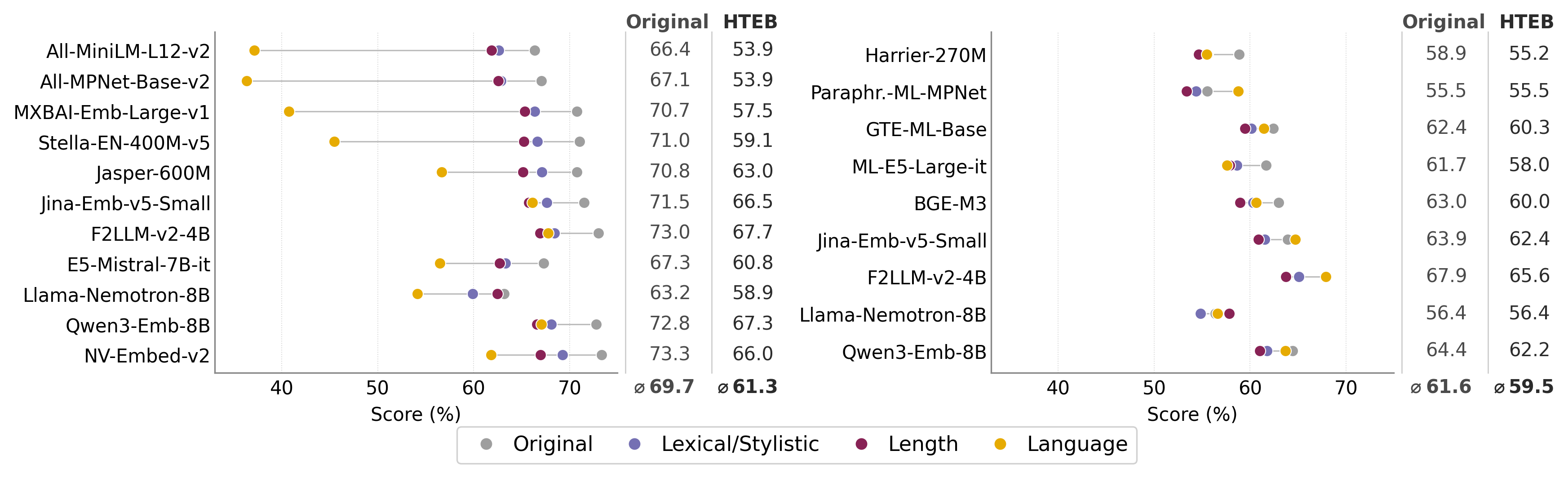}
    \caption{\emph{Left: }Embedding model performance on \textcolor{englishblue}{\textbf{English}} data (19 datasets, 3 runs). \emph{Right:} Embedding model performance on \textcolor{multiorange}{\textbf{multilingual}} data (13 datasets, 3 runs). \emph{Both:} Right columns report Original and total HTEB score with ø denoting the average. Exact scores per axis and model can be found in \autoref{tab:robustness-axis-dumbbell-joint-horizontal-english} and \autoref{tab:robustness-axis-dumbbell-joint-horizontal-multilingual} in \autoref{app_overall_results_detailed}.} 
    \label{fig:english_multilingual_total_results}
\end{figure*}

\subsection{HTEB Embedding Model Evaluation}

\subsubsection{English Evaluation}
\label{sec:english_eval}

\autoref{fig:english_multilingual_total_results} (left) shows the performance of the embedding models on the English datasets, with HTEB scores broken down by robustness axis. All models show overall drops. F2LLM-v2-4B overtakes NV-Embed-v2 under HTEB, moving from second to first place, while NV-Embed-v2 drops from first to fourth. The analysis shows that axis-level behaviour varies across models. Language transformations produce the largest aggregate drops, particularly for English-only models. These drops measure sensitivity to a change in deployment language and should not be interpreted as a lack of robustness within a strictly English-only setting. The size of the language-axis drop, however, differs across models: All-MiniLM-L12-v2 and MXBAI-Embed-Large-v1 show large drops on this axis, Jasper-600M shows a smaller drop despite being bilingual (English/Chinese) rather than multilingual, and Llama-Nemotron-8B shows only a minimal drop on the Length axis while length transformations are disruptive for all other models. The average performance drop across models is 8.4 pp.

\begin{figure}
    \centering
    \includegraphics[width=1\linewidth]{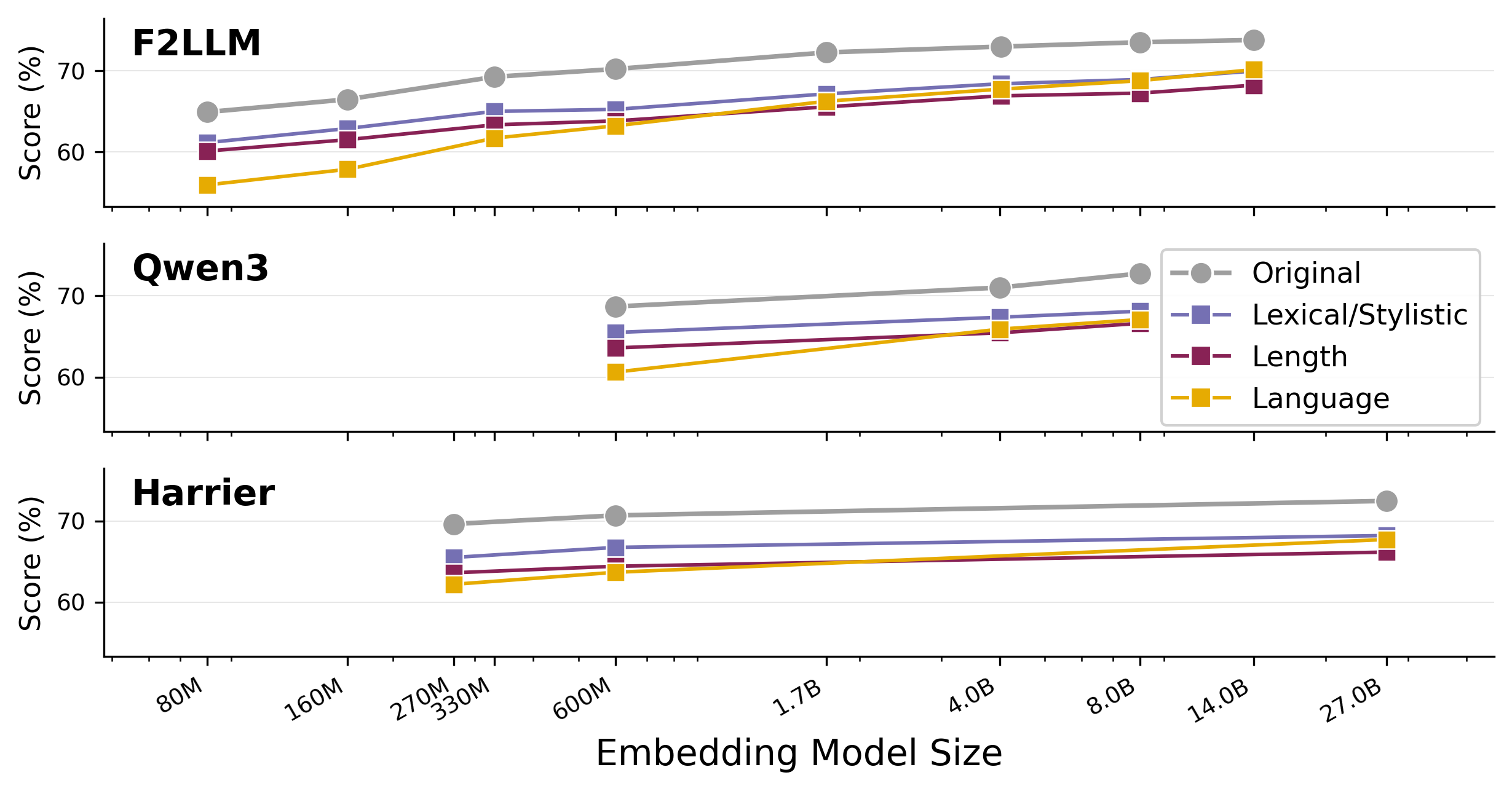}
    \caption{Size ablations for Qwen3-Embedding, F2LLM, and Harrier averaged over 19 English datasets and 3 runs. Scores are decomposed by robustness axis.}
    \label{fig:ablation_per_dimension}
\end{figure}

In the ablation study, we evaluate embedding models from the F2LLM, Qwen3, and Harrier families at different parameter counts on HTEB's English datasets.  \autoref{fig:ablation_per_dimension} shows that the gap between the Original score and the overall HTEB score persists across model sizes, remaining around five points in most cases. Across families, model size disproportionately improves Language axis scores.

\subsubsection{Multilingual Evaluation}

On the \textcolor{multiorange}{\textbf{multilingual}} data, most average scores drop under HTEB's transformations, though by a smaller margin compared to English (about -2.1 pp vs. -8.4 pp on average, \autoref{fig:english_multilingual_total_results}). Paraphrase-Multilingual-MPNet and Llama-Nemotron-8B are the exceptions that show effectively no average score change. For Paraphrase-Multilingual-MPNet this is due to the Language axis but Llama-Nemotron-8B offsets drops by a net positive on the Length axis.

\begin{figure*}
    \centering
    \includegraphics[width=1.0\linewidth]{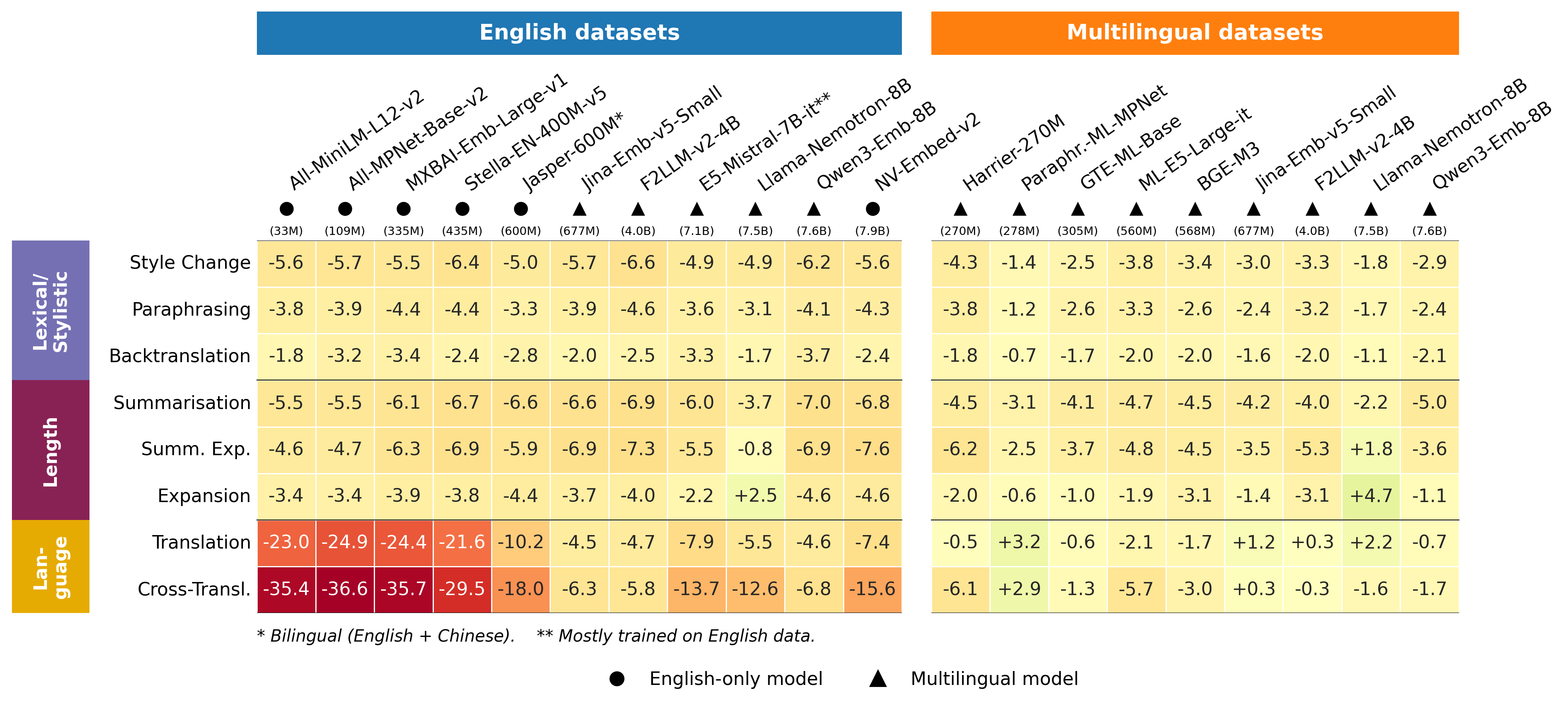}
    \caption{Mean score delta ($\Delta = \text{HTEB} - \text{Original}$) per embedding model (model size in parentheses)
        and transformation, averaged over datasets and seeds. Negative values (red)
        indicate degradation under HTEB's transformations; positive values (green) indicate
        improvement.
    }
    \label{fig:model_transform_heatmap}
\end{figure*}

Holding models and items constant, the smaller multilingual performance drop replicates on a German translation of STS-B generated with Gemma-3: aggregate drop vs. original scores is -14.9 on STS-B and -10.2 on STS-B-German (\autoref{tab:stsb_german_per_model_grouped} in \autoref{sec:sts_deu}). The gap shrinks but persists when restricted to multilingual models (-7.7 on English STS-B vs. -6.1 on STS-B-German).

Additionally, \autoref{fig:hteb_drop_vs_original_score} (\autoref{app_eng_vs_multilingual}) plots HTEB drop vs. original score at the dataset-model-transformation level for English and multilingual evaluation, showing that the English-multilingual asymmetry is not limited to a few outliers.

Excluding the Language axis reduces the all-model English drop from 8.41 to 4.56 points, compared with 2.67 points on the multilingual datasets. Restricting the comparison to the four shared models yields drops of 5.02 versus 1.51 points with the Language axis and 4.37 versus 2.31 points without it (\autoref{tab:hteb_gap_summary}). Their mean relative full-HTEB drops are 7.16\% and 2.30\%, respectively; per-model results are reported in \autoref{tab:shared_model_relative_drops} in the Appendix.

\begin{table}[t]
\centering
\small
\begin{tabular}{lccc}
\toprule
\textbf{Evaluation condition} & \textcolor{englishblue}{\textbf{English}} & \textcolor{multiorange}{\textbf{Multilingual}} & \textbf{Gap} \\
\midrule
\multicolumn{4}{c}{\textit{All models}} \\
\midrule
\quad HTEB w/ Lang. & -8.41 & -2.04 & 6.37 \\
\quad HTEB w/o Lang. & -4.56 & -2.67 & 1.89 \\
\midrule
\multicolumn{4}{c}{\textit{Shared multilingual models}} \\
\midrule
\quad HTEB w/ Lang. & -5.02 & -1.51 & 3.52 \\
\quad HTEB w/o Lang. & -4.37 & -2.31 & 2.06 \\
\bottomrule
\end{tabular}
\caption{Absolute HTEB performance drops under English and multilingual evaluation settings (HTEB score minus original score). "All models" refers to the models evaluated under the respective language setting, i.e., the 11 models listed on the left for \textcolor{englishblue}{\textbf{English}} and the 9 models on the right for \textcolor{multiorange}{\textbf{Multilingual}} in \autoref{fig:english_multilingual_total_results}. "Shared multilingual models" includes Jina-Embeddings-v5-Text-Small, F2LLM-v2-4B, Llama-Nemotron-Embed-8B, and Qwen3-Embedding-8B (see \autoref{tab:shared_model_relative_drops} in the Appendix for details).}
\label{tab:hteb_gap_summary}
\end{table}

\subsubsection{Overall HTEB Results}
\label{sec:overall_hteb_results}
\autoref{fig:model_transform_heatmap} decomposes performance differences to the transformation level. Three patterns stand out: (1) Cross-Translation is more disruptive than Translation in most cases. (2) Only Paraphrase-Multilingual-MPNet and Jina-Embeddings-v5-Small show positive deltas on both Language transformations in the multilingual setting despite Jina having negative deltas on both transformations in the English setting (-4.5 pp Translation, -6.3 pp Cross-Translation). (3) Llama-Nemotron-8B is the only model with positive deltas on most length-increasing transformations, although the effect is larger on multilingual data (Expansion +4.7 pp, Summarised Expansion +1.8 pp) than on English (Expansion +2.5 pp, Summarised Expansion -0.8 pp), consistent with generally larger performance drops on English data for this axis. While English-only models show the largest drops on the Language axis, multilingual models are not uniformly language robust either: Nemotron drops 12.6 pp on Cross-Translation on English data while gaining 2 pp on Translation on multilingual data. For results per dataset, please refer to \autoref{app_dataset_level_results}.

Ranking the three axes by absolute disruption within each model, Language is the most disruptive axis on English data for eight of eleven models with three models being more sensitive to changes in length. On multilingual data, Length is most disruptive for six of nine models, Language for two, and Lexical/Stylistic for one model. For an additional visualisation, see \autoref{app_axis_ranking}.

To check whether per-axis profiles are model properties or dataset-selection artefacts, we randomly partitioned datasets into task--stratified halves and correlated the resulting model rankings (\autoref{tab:half_split_per_axis_compact}). Median $\rho$ ranges from 0.82 to 0.96 on English and 0.80 to 0.93 on multilingual, indicating that per-axis rankings are stable across task-stratified partitions of the included datasets and are not overly sensitive to a particular subset of HTEB's datasets.

\begin{table}[t]
\small
\centering
\setlength{\tabcolsep}{6pt}
\begin{tabular}{lcc}
\toprule
\textbf{Axis} & \textcolor{englishblue}{\textbf{English}} & \textcolor{multiorange}{\textbf{Multilingual}} \\
\midrule
\textbf{Lexical / Stylistic} & +0.92 & +0.82 \\
\textbf{Length} & +0.82 & +0.80 \\
\textbf{Language} & +0.96 & +0.93 \\
\bottomrule
\end{tabular}
\caption{Split-half reliability of per-axis robustness profiles: Spearman $\rho$ medians over $1000$ stratified random dataset splits for \textcolor{englishblue}{\textbf{English}} and \textcolor{multiorange}{\textbf{Multilingual}} data.}
\label{tab:half_split_per_axis_compact}
\end{table}

\begin{figure}[!tb]
    \centering
    \includegraphics[width=1.0\linewidth]{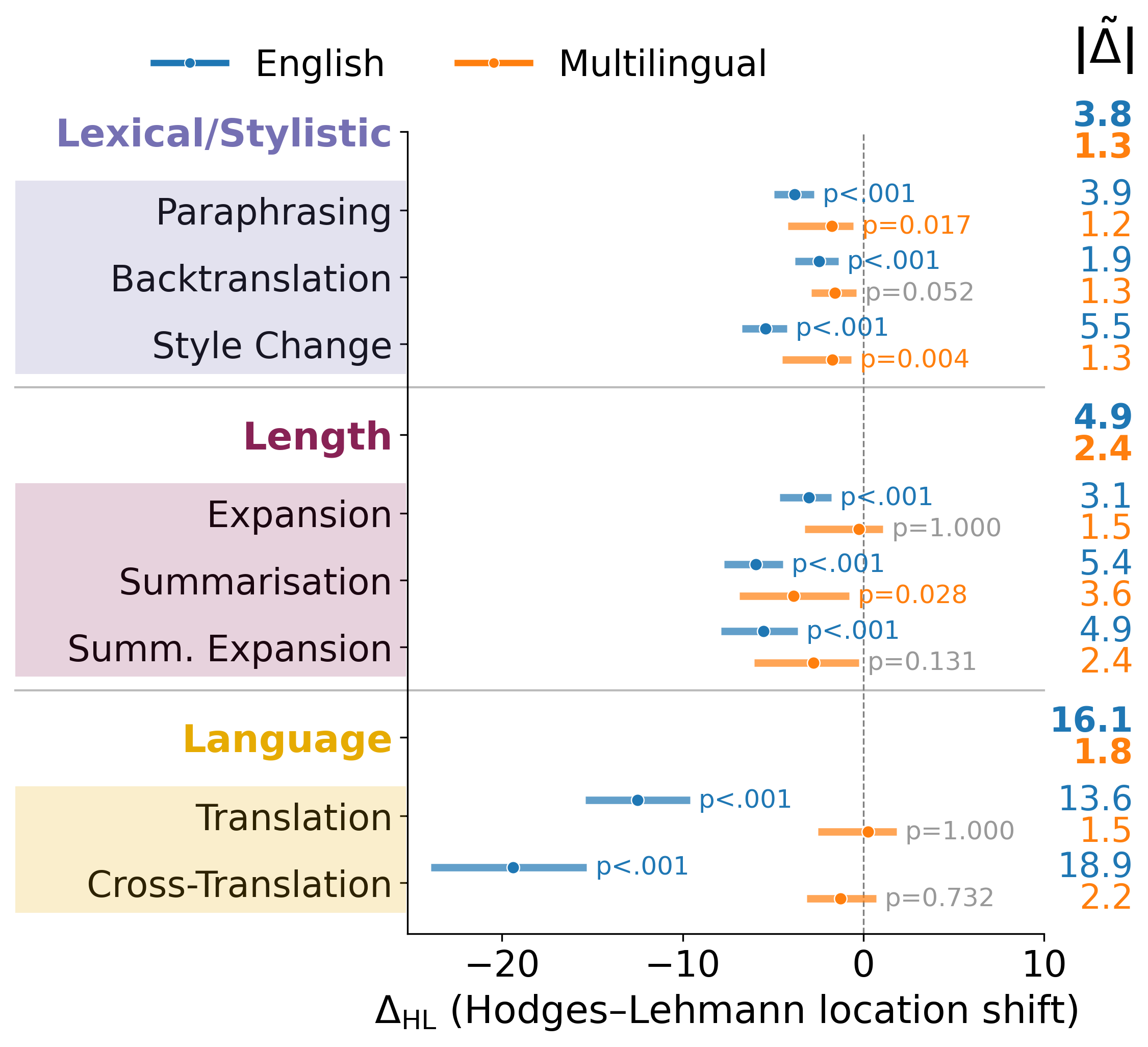}
    \caption{HL location shift with per-transformation 95\% CIs for English and multilingual benchmarks. CIs not adjusted for multiple comparisons. P-values are Holm-corrected with $p \geq .05$ greyed out. $|\tilde{\Delta}|$ reports the median absolute score difference.}
    \label{fig:hl_forest_colored}
\end{figure}

\autoref{fig:hl_forest_colored} reports the aggregated Hodges-Lehmann location shift per transformation type across all models, with Holm-corrected Wilcoxon signed-rank tests. The results are consistent with the preceding findings: all eight transformations produce statistically significant performance decreases ($p < .001$) for the English datasets. 

For multilingual datasets, the median location shifts are smaller and vary in direction, with positive deltas on some transformations partially offsetting negative ones; hence, only 3 of 8 transformations reach statistical significance for shifts in a single direction. Rather than indicating uniform multilingual robustness, this pattern reflects heterogeneous model responses: the median absolute differences remain practically meaningful (English: $1.9$ to $18.9$ points; multilingual: $1.2$ to $3.6$ points).

To cross-check if the results depend on using Gemma-3 as the transformation model, we use a subset of the English data with Qwen3-8B-AWQ and find broadly comparable overall disruption, with the main difference being weaker Length-axis disruption under Qwen3 (see \autoref{app_gen_model_confounding}).

\begin{table}[t]
    \small
    \centering
    \setlength{\tabcolsep}{4pt}
    \begin{tabular}{lcc}
    \toprule
    \textbf{Transformation} & \textcolor{englishblue}{\textbf{English}} & \textcolor{multiorange}{\textbf{Multilingual}} \\
    \midrule
    \textbf{Lexical/Stylistic} & \textbf{0.742}~\tiny{$\pm$0.191} & \textbf{0.816}~\tiny{$\pm$0.144} \\
    \quad Paraphrasing & 0.706~\tiny{$\pm$0.210} & 0.782~\tiny{$\pm$0.161} \\
    \quad Backtranslation & 0.802~\tiny{$\pm$0.146} & 0.858~\tiny{$\pm$0.160} \\
    \quad Style Change & 0.669~\tiny{$\pm$0.223} & 0.782~\tiny{$\pm$0.141} \\
    \midrule
    \textbf{Length} & \textbf{0.619}~\tiny{$\pm$0.221} & \textbf{0.637}~\tiny{$\pm$0.211} \\
    \quad Expansion & 0.486~\tiny{$\pm$0.273} & 0.439~\tiny{$\pm$0.348} \\
    \quad Summarisation & 0.629~\tiny{$\pm$0.335} & 0.786~\tiny{$\pm$0.183} \\
    \quad Summ.\ Expansion & 0.522~\tiny{$\pm$0.268} & 0.590~\tiny{$\pm$0.196} \\
    \midrule
    \textbf{Language} & \textbf{0.373}~\tiny{$\pm$0.205} & \textbf{0.658}~\tiny{$\pm$0.214} \\
    \quad Translation & 0.386~\tiny{$\pm$0.166} & 0.650~\tiny{$\pm$0.200} \\
    \quad Cross-Translation & 0.369~\tiny{$\pm$0.217} & 0.654~\tiny{$\pm$0.187} \\
    \midrule
    \textbf{Overall} & \textbf{0.555}~\tiny{$\pm$0.232} & \textbf{0.761}~\tiny{$\pm$0.135} \\
    \bottomrule
    \end{tabular}
    \caption{
    Kendall's $\tau$ rank correlation between Original and HTEB model rankings per robustness axis and transformation. Each cell shows the mean $\pm$ sample standard
    deviation across per-dataset taus.
    }
    \label{tab:kendall_tau}
    \end{table}

\subsubsection{Model Ranking Stability}

Analysing Kendall's $\tau$ per axis and transformation shows that HTEB leads to substantial model reranking at the dataset level (\autoref{tab:kendall_tau}). 

Mean $\tau$ is lower on English (0.555) than on multilingual data (0.761), so HTEB's transformations frequently alter model orderings, more so on English. Lexical/Stylistic transformations preserve the most rank agreement in both settings. On English, rank agreement is lowest for Language, while on multilingual data Length and Language are close, with Length slightly lower. Within the Length axis, Expansion induces the most reshuffling and Summarisation the least in both settings. Backtranslation yields the highest $\tau$ among individual transformations, which is consistent with its role as a near-identity anchor transformation.

To assess whether rank instability is driven by stochastic variation between transformation runs, we calculate pairwise Kendall's $\tau$ between model rankings across the three runs (within-$\tau$). For all transformations the average within-$\tau$ is larger than the original $\tau$ (see \autoref{tab:within_tau} in \autoref{app_within_tau}). This indicates that stochastic generation is not the main source of rank instability.

\section{Discussion}

\subsection{English–Multilingual Sensitivity Gap}
\label{sec:discussion_lingual_gap}

An emerging pattern is that models degrade more under transformations applied to English than multilingual data (8.41 versus 2.04 points; \autoref{fig:english_multilingual_total_results}). Because this aggregate comparison may be affected by model composition, the Language axis, and differences in task and dataset composition, we conduct several complementary checks. First, excluding the Language axis reduces the English-multilingual gap to 1.89 points. Second, restricting the comparison to the same four multilingual models controls for model composition and still yields larger relative full-HTEB drops on English than multilingual data (7.16\% versus 2.30\%; \autoref{tab:shared_model_relative_drops} in the Appendix). Third, to hold the task, items, and models constant, we repeat the evaluation on a German translation of STS-B and again observe a larger drop on the English version (\autoref{tab:stsb_german_per_model_grouped}). Finally, to assess whether known training overlap explains the pattern, we restrict the comparison to model-dataset pairs marked as zero-shot in MTEB metadata; the average drops remain 8.84 points on English and 2.10 points on multilingual datasets (see \autoref{app_zero_shot} for details). Together, these results show that the gap is not explained by model composition, the Language axis, or known training overlap alone. They do not, however, isolate language as the only cause, since HTEB's English and multilingual benchmarks are not parallel samples and baseline difficulty and other benchmark-level differences may act as uncontrolled confounders.
 
\subsection{Model-Specific Robustness Profiles}
\label{sec:discussion_robustness_axis_diss}

In line with the English-Multilingual Sensitivity Gap, language transformations induce the largest rank reordering for English and length for multilingual evaluation (\autoref{tab:kendall_tau}). The analysis confirms that the axes are partly decoupled in practice and specific to models and datasets: they do not merely rescale a common robustness signal, but induce different degrees of model reranking.

At the model level, the partial decoupling of axes is more pronounced. Paraphrase-Multilingual-MPNet and, to a lesser extent, Jina-Embeddings-v5-Small and Llama-Nemotron-8B are more robust on the Language axis (\autoref{fig:english_multilingual_total_results}, \autoref{fig:model_transform_heatmap}). Consequently, Paraphrase-Multilingual-MPNet is one of two models whose overall HTEB score is effectively unchanged relative to the original score. In contrast, Llama-Nemotron-8B's original score is well below those of other 8B models, yet it maintains original performance under HTEB in the multilingual evaluation, driven primarily by gains on length-increasing transformations (\autoref{fig:model_transform_heatmap}). 

Together, these observations show that absolute benchmark performance and robustness are not interchangeable: static rankings can miss model-specific weaknesses under transformation. The within-model axis orderings reported in \autoref{sec:overall_hteb_results} quantify these differences: robustness profiles vary across models rather than being repetitions of a shared pattern. The split-half analysis (\autoref{tab:half_split_per_axis_compact}) further supports this interpretation: per-axis profiles reflect reproducible model properties rather than artefacts of HTEB's dataset selection.

\subsection{Scaling Shifts Robustness Profiles}
\label{sec:discussion_scale_isnt_robustness}

Within the three model families studied here, scaling does not generally improve HTEB robustness on the English subset. Instead, it leads to similar absolute improvements in original data and HTEB. However, larger parameter counts are associated with smaller language-axis degradation. The ablation patterns reported in \autoref{sec:english_eval} carry two implications for embedding model evaluation. First, improvements on untransformed data do not necessarily translate into smaller transformation-induced degradation, so original-data leaderboards should not be treated as robustness proxies. Second, the family-specific axis profiles suggest that robustness is not influenced by model size alone. 
 
\subsection{Robustness Profiles, Not Just Performance Degradation under Perturbations}
\label{sec:discussion_robustness_profiles}

Prior work, notably PTEB~\citep{frank-afli-2026-pteb}, established that embedding models degrade under paraphrase stress. As the findings discussed in this section show, HTEB is not simply an extension of this result with additional transformations. Rather, it decomposes robustness into interpretable, partly decoupled axes and separates absolute score drops from ranking stability. This has been illustrated with examples, such as the language-robustness of Paraphrase-Multilingual-MPNet and the length-robustness of Llama-Nemotron-8B. 

A one-dimensional benchmark can show whether a model improves; HTEB can show where and how robust the improvement is.

\section{Conclusion}
\label{sec:conclusion}
We introduced the Harder Text Embedding Benchmark (HTEB), which challenges embedding models with dynamically and stochastically LLM-generated text variants in three deployment-motivated dimensions: Lexical/Stylistic, Length, and Language. The quality and validity of these transformations for English data have been assessed through error analysis, LLM judges, and human evaluation. HTEB thereby establishes a dynamic, multidimensional robustness framework that reveals models' axis-specific robustness. Across 16 embedding models on 32 datasets covering 42 languages, we find that HTEB's transformations produce significant performance shifts and substantial ranking instability. Our findings demonstrate that robustness along the three axes is partially decoupled, that increasing scale does not generally increase robustness for the three model families studied, and that the resulting model-specific robustness profiles provide insights relevant to embedding model selection both between and within model families. 

Importantly, HTEB's axes are not intended as universal robustness requirements. For a strictly English deployment, for example, the Language axis should be omitted or reported separately. Conversely, when multilingual or cross-lingual input is relevant, the Language axis can extend otherwise English-only benchmark datasets with generated non-English variants, enabling multilingual stress tests when no suitable parallel benchmark version is available. As a minimal robustness check in practical applications, we recommend evaluating Paraphrasing (as an anchor) plus either Expansion or Summarisation depending on which is more relevant to the deployment setting and dataset. If multilingual input is relevant, Translation should be added and, similarly, Cross-Translation when cross-lingual capabilities are required.

\FloatBarrier
\section*{Limitations}
Our work has several limitations:

\paragraph{What HTEB does and does not measure.} HTEB covers a broad range of transformations, but does not reflect all changes that may occur in real-world deployment. Its transformations depend on the design choices and the LLMs used to generate these. The resulting scores also depend on the benchmark composition, the task-specific metrics, and the aggregation function for overall scores. It does not provide general guarantees on model performance for real-world distribution shifts.

\paragraph{LLM-generated transformations as a potential confound.} Because HTEB's transformations are produced by a generative LLM, embedding models whose training data includes substantial amounts of LLM-generated text may be implicitly adapted to the distribution of such transformations, potentially inflating their robustness scores. Conversely, models trained exclusively on human-authored text may be disproportionately penalised. As synthetic training data becomes increasingly common in (embedding) model development, this circularity risk grows. We note that this concern applies to any LLM-based evaluation framework, including PTEB and LLM-as-judge paradigms, and is not unique to HTEB. Future work could investigate the extent of this effect by comparing robustness scores of models with known synthetic and non-synthetic training compositions.

\paragraph{Limited language and task coverage of human validation.} The principal study uses English-source items; the exploratory Spanish-source study covers one additional source language, and pair-level preservation is assessed only on STS-B for Paraphrasing, Summarised Expansion, and same-target Translation. Cross-Translation was not directly validated at the pair level; on paired tasks it changes same-language similarity into cross-lingual alignment and should therefore be interpreted as a deployment-specific cross-lingual capability.

\paragraph{LLM-contamination in human evaluation.} Despite filtering Amazon MTurk crowd workers for high acceptance rates ($\geq95\%$) and at least 1,000 accepted HITs and having them confirm they did not use AI, we observed indicators of likely LLM use in some submissions: characteristic phrasings and punctuation (e.g., em-dashes), overly wordy explanations inconsistent with our time estimates, and positive signals from AI-detection tools (which we treat as suggestive rather than definitive given their known limitations). We therefore recruited crowd workers whose quality we could assess independently of Amazon MTurk's own metrics wherever possible. Generally, we expect LLM contamination to become an increasingly serious challenge for crowd-sourced human evaluation.

\paragraph{Transformation-task interactions.} Certain transformation-task combinations may test something beyond what they are intended to test. For example, summarising already short sentences may result in minimal change. While we apply all transformations uniformly across tasks for consistency, not all pairings are equally informative as robustness tests, and future work could investigate which combinations provide the strongest signal. Because of this and for runtime reasons (see below), we do not expect practical applications to always apply all of HTEB's transformations.  

\paragraph{Scope of the scale ablation.} Our scale analysis is limited to three embedding model families and the English subset of HTEB. The result should therefore not be interpreted as a universal claim about scaling in embedding models. Broader conclusions would require including additional model families and multilingual evaluation.

\paragraph{Prompt dependence of instruction-tuned embedding models.}
HTEB varies the input texts but keeps the embedding-model instruction fixed for instruction embedding models. Recent work shows that instruction-tuned embedding models can be highly sensitive to the phrasing of this instruction~\citep{Kostiuk2026_OnePromptNotEnough}. We did not evaluate this source of variation.

\paragraph{Runtime.} Performing eight transformations each with three random seeds significantly increases the runtime, particularly since inference with generative models often has much higher runtime than with embedding models. This limitation is particularly relevant for large datasets. We therefore provide practical recommendations regarding the use of HTEB-like evaluation in \autoref{sec:conclusion}. 

\paragraph{Hyperparameters.} We did not explore the effect of different hyperparameter settings, e.g. temperature. This remains to be explored in future work.    

%
%
\section*{Acknowledgments}
We thank the anonymous reviewers and human annotators for their valuable contributions. We also thank the following people for their valuable contribution as annotators: Naciye Bekmezci, Nils Birkhan, Sandra Kristina Döring, Sara Vecino Garcia, Kamil Kabakçılı, Manuel Quintela Pumares, Ümmü Gülsüm Şen, and Alejandro Alvarez Varela.

This research was partially supported by the Horizon Europe project GenDAI (Grant Agreement ID: 101182801) and by the ADAPT Research Centre at Munster Technological University. ADAPT is funded by Taighde Éireann – Research Ireland through the Research Centres Programme and co-funded under the European Regional Development Fund (ERDF) via Grant 13/RC/2106\_P2.

\section*{Ethics Statement}
We did not identify critical ethical problems within this work. 

\paragraph{Crowd worker compensation.} Since our work involved crowd workers recruited via Amazon Mechanical Turk, we ensured that they were fairly compensated. Based on initial tests, we estimated that rating a single transformation on both quality and fluency would take approximately 30 seconds on average. Based on this, we set a minimum hourly compensation rate of \$24.00. For the Expansion task, we increased the minimum hourly compensation by 25\% and for Summarised Expansion by 10\%, to account for the longer texts and corresponding processing times. 

\paragraph{AI Usage.} AI was used to assist with coding and to improve the language of our text. In both cases, the authors take full accountability for the output.

\bibliography{references}
\appendix
\section{Generative Model Selection}
\label{app_gen_model_selection}
\subsection{Generative Models and Hyperparameters}
\label{app_gen_models}
Since we use short names for the models throughout this paper, \autoref{tab:generative_model_revision_ids} provides the full model names including revision IDs. 
\begin{table*}[t]
\centering
\scriptsize
\setlength{\tabcolsep}{2pt}
\begin{tabular}{@{}p{0.58\textwidth}p{0.38\textwidth}@{}}
\toprule
\textbf{Generative model checkpoint} & \textbf{Revision ID} \\
\midrule
\href{https://huggingface.co/allenai/Olmo-3-7B-Instruct}{\texttt{allenai/\allowbreak{}Olmo-\allowbreak{}3-\allowbreak{}7B-\allowbreak{}Instruct}} & \texttt{51c1f0b2\allowbreak{}8546f9cf\allowbreak{}537583ea\allowbreak{}5b9f5da4\allowbreak{}fed23ebe} \\
\href{https://huggingface.co/cpatonn/Apriel-1.5-15b-Thinker-AWQ-4bit}{\texttt{cpatonn/\allowbreak{}Apriel-\allowbreak{}1.\allowbreak{}5-\allowbreak{}15b-\allowbreak{}Thinker-\allowbreak{}AWQ-\allowbreak{}4bit}} & \texttt{0e414fe9\allowbreak{}d1eac914\allowbreak{}ff100cf7\allowbreak{}11e3260f\allowbreak{}932d608f} \\
\href{https://huggingface.co/cpatonn/Apriel-1.5-15b-Thinker-AWQ-8bit}{\texttt{cpatonn/\allowbreak{}Apriel-\allowbreak{}1.\allowbreak{}5-\allowbreak{}15b-\allowbreak{}Thinker-\allowbreak{}AWQ-\allowbreak{}8bit}} & \texttt{9f14e8f5\allowbreak{}a1fa3111\allowbreak{}f26663b9\allowbreak{}01e7737e\allowbreak{}972b7d30} \\
\href{https://huggingface.co/cyankiwi/Apriel-1.6-15b-Thinker-AWQ-4bit}{\texttt{cyankiwi/\allowbreak{}Apriel-\allowbreak{}1.\allowbreak{}6-\allowbreak{}15b-\allowbreak{}Thinker-\allowbreak{}AWQ-\allowbreak{}4bit}} & \texttt{95519242\allowbreak{}a2c600de\allowbreak{}c0537c64\allowbreak{}3d8bd5e5\allowbreak{}fe5d45c5} \\
\href{https://huggingface.co/cyankiwi/Olmo-3-32B-Think-AWQ-4bit}{\texttt{cyankiwi/\allowbreak{}Olmo-\allowbreak{}3-\allowbreak{}32B-\allowbreak{}Think-\allowbreak{}AWQ-\allowbreak{}4bit}} & \texttt{0c93f53d\allowbreak{}48a79cce\allowbreak{}eac8ae52\allowbreak{}02abab19\allowbreak{}71dcf40d} \\
\href{https://huggingface.co/gaunernst/gemma-3-12b-it-int4-awq}{\texttt{gaunernst/\allowbreak{}gemma-\allowbreak{}3-\allowbreak{}12b-\allowbreak{}it-\allowbreak{}int4-\allowbreak{}awq}} & \texttt{3b0c67b9\allowbreak{}8eee8fb9\allowbreak{}0633ef1b\allowbreak{}fbf3d39f\allowbreak{}43b9cf9d} \\
\href{https://huggingface.co/gaunernst/gemma-3-27b-it-int4-awq}{\texttt{gaunernst/\allowbreak{}gemma-\allowbreak{}3-\allowbreak{}27b-\allowbreak{}it-\allowbreak{}int4-\allowbreak{}awq}} & \texttt{7cf8bdc8\allowbreak{}1343c635\allowbreak{}390dd1e1\allowbreak{}bf590ab2\allowbreak{}2dd6f366} \\
\href{https://huggingface.co/gaunernst/gemma-3-4b-it-int4-awq}{\texttt{gaunernst/\allowbreak{}gemma-\allowbreak{}3-\allowbreak{}4b-\allowbreak{}it-\allowbreak{}int4-\allowbreak{}awq}} & \texttt{8f28faf0\allowbreak{}5c382a2d\allowbreak{}d81a4710\allowbreak{}90acdb23\allowbreak{}156eb354} \\
\href{https://huggingface.co/mistralai/Ministral-3-14B-Instruct-2512}{\texttt{mistralai/\allowbreak{}Ministral-\allowbreak{}3-\allowbreak{}14B-\allowbreak{}Instruct-\allowbreak{}2512}} & \texttt{1e4bed9a\allowbreak{}74c1d8af\allowbreak{}713dd9e2\allowbreak{}545d6902\allowbreak{}0ced05dc} \\
\href{https://huggingface.co/mistralai/Ministral-3-3B-Instruct-2512}{\texttt{mistralai/\allowbreak{}Ministral-\allowbreak{}3-\allowbreak{}3B-\allowbreak{}Instruct-\allowbreak{}2512}} & \texttt{b28a99a0\allowbreak{}5583a633\allowbreak{}d139fe03\allowbreak{}3225ad26\allowbreak{}0a0062c6} \\
\href{https://huggingface.co/mistralai/Ministral-3-8B-Instruct-2512}{\texttt{mistralai/\allowbreak{}Ministral-\allowbreak{}3-\allowbreak{}8B-\allowbreak{}Instruct-\allowbreak{}2512}} & \texttt{172a4a40\allowbreak{}656d3e29\allowbreak{}d31473e1\allowbreak{}25a9f435\allowbreak{}b5d84455} \\
\href{https://huggingface.co/Qwen/Qwen3-14B-AWQ}{\texttt{Qwen/\allowbreak{}Qwen3-\allowbreak{}14B-\allowbreak{}AWQ}} & \texttt{31c69efc\allowbreak{}29464b6b\allowbreak{}b0aee139\allowbreak{}8b5a7b50\allowbreak{}a99340c3} \\
\href{https://huggingface.co/Qwen/Qwen3-4B}{\texttt{Qwen/\allowbreak{}Qwen3-\allowbreak{}4B}} & \texttt{1cfa9a72\allowbreak{}08912126\allowbreak{}459214e8\allowbreak{}b0432160\allowbreak{}3b3df60c} \\
\href{https://huggingface.co/Qwen/Qwen3-4B-AWQ}{\texttt{Qwen/\allowbreak{}Qwen3-\allowbreak{}4B-\allowbreak{}AWQ}} & \texttt{74d4bd2b\allowbreak{}d4bff9ca\allowbreak{}fc934522\allowbreak{}1320bffb\allowbreak{}08b406a3} \\
\href{https://huggingface.co/Qwen/Qwen3-8B}{\texttt{Qwen/\allowbreak{}Qwen3-\allowbreak{}8B}} & \texttt{b968826d\allowbreak{}9c46dd60\allowbreak{}66d109ea\allowbreak{}bc625518\allowbreak{}8de91218} \\
\href{https://huggingface.co/Qwen/Qwen3-8B-AWQ}{\texttt{Qwen/\allowbreak{}Qwen3-\allowbreak{}8B-\allowbreak{}AWQ}} & \texttt{4da05a8e\allowbreak{}db55c604\allowbreak{}6cce9585\allowbreak{}86c33b61\allowbreak{}da07bb79} \\
\bottomrule
\end{tabular}
\caption{Generative model revisions IDs.}
\label{tab:generative_model_revision_ids}
\end{table*}

All models were run with \verb|temperature=0| and \verb|top_p=1| to ensure reproducibility. Transformations were generated using the following random seeds: 1337, 1338, 1339.

\subsection{Error Rate Analysis}
\label{app_err_rate}
To select a generative model we conducted an analysis of the models listed in \autoref{tab:error_rates} and evaluated the errors made over all transformations using STS-B as a dataset. Our analysis included the following error types:
\begin{itemize}
    \item \textbf{Identical Rate:} The transformed output is identical to the original input text (case-insensitive comparison), indicating that the transformation failed to modify the text.
    
    \item \textbf{Empty Rate:} The model produces an empty string or whitespace-only output, indicating complete generation failure.
    
    \item \textbf{Ellipsis Rate:} The model outputs only ellipsis characters
    (\texttt{...}, \texttt{..}, or the Unicode ellipsis \texttt{…}),
    indicating that the model refused or failed to generate a transformation.
    
    \item \textbf{JSON Fragment Rate:} The output begins with a JSON delimiter (\texttt{\{} or \texttt{[}), indicating the model produced structured data rather than natural language text.
    
    \item \textbf{Reasoning Leak Rate:} The output contains chain-of-thought (CoT) reasoning markers such as "Here are my reasoning", "Let me think", "I'll", or "Step N:", indicating the model leaked its internal reasoning process instead of providing only the transformed text.
    
    \item \textbf{Prefix Leak Rate:} The output begins with an instruction prefix such as "Translated text:", "Paraphrased text:", "Summary:", "Translation:", or "Paraphrase:", indicating the model included the task instruction in its output.
    
    \item \textbf{Wrong Language:} The output is in the wrong language. For translation and cross-translation this includes sentences that do not match the selected target language. For other transformations it detects outputs that do not match the input language. 
    
    \item \textbf{Runaway Generation Rate:} The output is excessively long relative to the input. For all transformations except Expansion and Summarised Expansion, this occurs when the output word count exceeds 5 times the input word count.
    
    \item \textbf{Truncated Rate:} The output is significantly shorter than the input. For all transformations except Summarisation with inputs longer than 3 words, this occurs when the output word count is less than 20\% of the input word count.
    
    \item \textbf{Summary Too Long Rate:} For Summarisation, this occurs when the summary output contains more words than the original input, defeating the purpose of summarisation.
\end{itemize}
Please note that these error types are not mutually exclusive. The "total error rate", however, counts only a maximum of one error per sample.

An outstanding observation is the very high error rates of the Apriel family of generative models. We did not investigate this systematically, but three factors plausibly contribute: (i) our hyperparameter settings (\verb|temperature=0|, \verb|top_p=1|) deviate from the recommendations of the model cards\footnote{see \url{https://huggingface.co/ServiceNow-AI/Apriel-1.5-15b-Thinker} and \url{https://huggingface.co/ServiceNow-AI/Apriel-1.6-15b-Thinker}} which suggest \verb|temperature=0.6|; (ii) Apriel-Thinker variants emit a \verb|[BEGIN FINAL RESPONSE]| envelope after a reasoning prefix, which our general output post-processor does not extract, so reasoning text is conflated with the transformation; and (iii) we evaluated community-built AWQ quantisations, and quantisation could explain the error rates to some extent. 

\begin{table*}[ht]
    \centering
    \resizebox{\textwidth}{!}{%
        \begin{tabular}{@{}lrrrrrrrrrr|r@{}}
            \toprule
            \textbf{Model} & \makecell{\textbf{Identical}\\\textbf{Rate}} & \makecell{\textbf{Empty}\\\textbf{Rate}} & \makecell{\textbf{Ellipsis}\\\textbf{Rate}} & \makecell{\textbf{JSON Frag.}\\\textbf{Rate}} & \makecell{\textbf{Reasoning}\\\textbf{Leak Rate}} & \makecell{\textbf{Prefix}\\\textbf{Leak Rate}} & \makecell{\textbf{Wrong}\\\textbf{Lang. Rate}} & \makecell{\textbf{Runaway}\\\textbf{Gen. Rate}} & \makecell{\textbf{Truncated}\\\textbf{Rate}} & \makecell{\textbf{Summary}\\\textbf{Too Long}} & \makecell{\textbf{Total Error}\\\textbf{Rate}} \\
            \midrule
            Apriel-1.5-15b-Thinker-AWQ-4bit  & 3.6 & 12.5 & 46.0 & 0.2 & 0.2 & \textbf{0.0} & 2.0 & \textbf{0.0} & 41.3 & 0.3 & 66.6 \\
            Apriel-1.5-15b-Thinker-AWQ-8bit  & 2.3 & \textbf{0.0} & 32.0 & 1.1 & 8.7 & \textbf{0.0} & 2.2 & \textbf{0.0} & 31.2 & 0.4 & 48.5 \\
            Apriel-1.6-15b-Thinker-AWQ-4bit  & \textbf{0.4} & \textbf{0.0} & 97.1 & 1.4 & \textbf{0.0} & \textbf{0.0} & 0.4 & \textbf{0.0} & 73.3 & \textbf{0.0} & 99.2 \\
            Gemma-3-12B-int4-AWQ             & 1.1 & 0.1 & \textbf{0.0} & 0.3 & \textbf{0.0} & \textbf{0.0} & 33.4 & \textbf{0.0} & 0.2 & \textbf{0.0} & 35.5 \\
            \rowcolor[gray]{0.9} Gemma-3-27B-int4-AWQ             & 1.7 & \textbf{0.0} & \textbf{0.0} & \textbf{0.1} & \textbf{0.0} & \textbf{0.0} & 0.6 & 0.2 & \textbf{0.0} & 0.4 & \textbf{3.0} \\
            Gemma-3-4B-int4-AWQ              & \textbf{0.4} & 0.3 & \textbf{0.0} & 7.0 & \textbf{0.0} & \textbf{0.0} & 3.3 & 3.3 & 0.5 & 0.3 & 11.5 \\
            \rowcolor[gray]{0.9} Ministral-3-14B-Instruct-2512    & 1.4 & \textbf{0.0} & \textbf{0.0} & 0.4 & \textbf{0.0} & \textbf{0.0} & 3.9 & \textbf{0.0} & 0.2 & 0.8 & 6.5 \\
            Ministral-3-3B-Instruct-2512     & 0.9 & \textbf{0.0} & \textbf{0.0} & 2.1 & \textbf{0.0} & \textbf{0.0} & 9.5 & 0.1 & 1.0 & 3.0 & 14.2 \\
            Ministral-3-8B-Instruct-2512     & 1.3 & \textbf{0.0} & \textbf{0.0} & 6.7 & \textbf{0.0} & \textbf{0.0} & \textbf{0.2} & \textbf{0.0} & 5.5 & 1.2 & 9.5 \\
            Olmo-3-32B-Think-AWQ-4bit        & 3.8 & \textbf{0.0} & \textbf{0.0} & \textbf{0.1} & \textbf{0.0} & \textbf{0.0} & 11.1 & \textbf{0.0} & 0.1 & 1.4 & 15.5 \\
            \rowcolor[gray]{0.9} Olmo-3-7B-Instruct               & 1.0 & \textbf{0.0} & \textbf{0.0} & 3.8 & \textbf{0.0} & \textbf{0.0} & 1.2 & \textbf{0.0} & 2.0 & 0.9 & 7.0 \\
            Qwen3-14B-AWQ                    & 6.2 & \textbf{0.0} & \textbf{0.0} & 2.2 & \textbf{0.0} & \textbf{0.0} & 0.4 & \textbf{0.0} & 2.0 & 0.3 & 9.3 \\
            Qwen3-4B                         & 2.6 & 0.1 & \textbf{0.0} & 4.1 & 0.1 & \textbf{0.0} & 0.8 & \textbf{0.0} & 3.4 & 0.2 & 7.9 \\
            Qwen3-4B-AWQ                     & 2.4 & \textbf{0.0} & \textbf{0.0} & 5.2 & 0.1 & \textbf{0.0} & 0.6 & \textbf{0.0} & 4.8 & 0.3 & 9.0 \\
            \rowcolor[gray]{0.9} Qwen3-8B                         & 2.5 & \textbf{0.0} & \textbf{0.0} & 2.9 & \textbf{0.0} & \textbf{0.0} & 0.6 & \textbf{0.0} & 2.4 & 0.9 & 7.1 \\
            \rowcolor[gray]{0.9} Qwen3-8B-AWQ                     & 2.5 & \textbf{0.0} & \textbf{0.0} & 1.9 & \textbf{0.0} & \textbf{0.0} & 0.5 & \textbf{0.0} & 1.6 & 0.8 & 5.8 \\
            \bottomrule
        \end{tabular}%
    }
    \caption{Average error rates in \% over all transformations for different generative models on STS-B (best scores: bold; top 5 models highlighted grey).}
    \label{tab:error_rates}
\end{table*}

\subsection{LLM Judge Eval}
\label{app_llm_judge_eval}

\autoref{tab:llm_judge_scores} shows the scores for each generative model and \autoref{tab:llm_judge_by_transformation} the LLM judge scores per transformation. 

\begin{table*}[ht]
    \small
    \centering
    \begin{tabular}{@{}lccccc@{}}
        \toprule
        & \multicolumn{5}{c}{\textbf{Transformation Model}} \\
        \cmidrule(lr){2-6}
        \textbf{LLM Judge} & \makecell{\textbf{Gemma-3-27B}} & \makecell{\textbf{Ministral-3-14B}} & \makecell{\textbf{Olmo-3-7B-Instr.}} & \makecell{\textbf{Qwen3-8B}} & \makecell{\textbf{Qwen3-8B-AWQ}} \\
        \midrule
        Gemma-3-27B      & \textbf{4.64} & 4.52 & 4.06 & 4.62 & 4.58 \\
        Ministral-3-14B  & \textbf{4.08} & 3.98 & 3.38 & 4.01 & 3.97 \\
        Olmo-3-7B-Instr. & 4.36 & 4.34 & 4.10 & \textbf{4.37} & 4.34 \\
        Qwen3-8B         & 4.66 & 4.60 & 4.03 & \textbf{4.68} & 4.65 \\
        Qwen3-8B-AWQ     & \textbf{4.52} & 4.48 & 3.87 & 4.47 & 4.46 \\
        \midrule
        Average [95\% CI] & \makecell{\textbf{4.45}$^a$\\{[4.44, 4.46]}} & \makecell{4.38$^c$\\{[4.37, 4.40]}} & \makecell{3.89$^d$\\{[3.87, 3.91]}} & \makecell{4.43$^b$\\{[4.41, 4.44]}} & \makecell{4.40$^c$\\{[4.38, 4.41]}} \\
        \bottomrule
    \end{tabular}
    \caption{LLM judge cross-evaluation scores for transformation quality (scale 1--5). 
        Superscripts indicate significance groups (bootstrapped, Holm-corrected, $\alpha$=0.05): 
        models sharing the same superscript letter are not statistically significantly different. All statistically significant pairwise differences have Holm-corrected $p<.001$; the only non-significant comparison is Ministral-3-14B vs. Qwen3-8B-AWQ.}
    \label{tab:llm_judge_scores}
\end{table*}

\begin{table*}[ht]
    \small
    \centering
    \begin{tabular}{@{}lccccc|c@{}}
        \toprule
        & \multicolumn{5}{c}{\textbf{Transformation Model}} & \\
        \cmidrule(lr){2-6}
        \textbf{Transformation} & \makecell{\textbf{Gemma-3-27B}} & \makecell{\textbf{Ministral-3-14B}} & \makecell{\textbf{Olmo-3-7B-Instr.}} & \makecell{\textbf{Qwen3-8B}} & \makecell{\textbf{Qwen3-8B-AWQ}} & \textbf{Total} \\
        \midrule
        Paraphrasing     & 4.64 & 3.70 & 4.38 & 4.71 & \textbf{4.73} & 4.43 \\
        Backtranslation  & \textbf{4.29} & 4.21 & 3.15 & 3.98 & 3.93 & 3.91 \\
        Style Change     & 3.58 & 3.78 & 3.95 & 4.12 & \textbf{4.13} & 3.91 \\
        Expansion        & 4.84 & \textbf{4.89} & 4.68 & 4.73 & 4.71 & 4.77 \\
        Summarisation    & 4.14 & 4.41 & 4.30 & \textbf{4.61} & 4.57 & 4.41 \\
        Summ. Expansion  & 4.55 & \textbf{4.75} & 4.28 & 4.56 & 4.48 & 4.52 \\
        Translation      & \textbf{4.71} & 4.47 & 2.54 & 4.10 & 4.06 & 3.97 \\
        Cross-Translation & \textbf{4.88} & 4.86 & 3.81 & 4.62 & 4.57 & 4.55 \\
        \midrule
        Total            & \textbf{4.45} & 4.38 & 3.89 & 4.43 & 4.40 & 4.31 \\
        \bottomrule
    \end{tabular}
    \caption{LLM judge quality scores by transformation (scale 1 to 5). Cells are
  averaged across judge models; Total row and column are arithmetic means computed
  before rounding.}
    \label{tab:llm_judge_by_transformation}
\end{table*}

\subsection{Human Eval}
\label{app_human_eval}

\paragraph{Quality and Fluency for English datasets.} Human ratings were collected via Amazon Mechanical Turk (MTurk) or from qualified volunteers who were native speakers of the relevant language. For MTurk, we required a minimum acceptance rate of 95\%, a minimum of 1,000 approved HITs, and the location filters in \autoref{tab:mturk_location_filters}. Language requirements were stated in the task descriptions, and workers confirmed their proficiency before beginning the survey. Following a data-minimisation policy, we did not collect age, gender, or other information not strictly needed for the study.

\begin{figure}
    \centering
    \includegraphics[width=1\linewidth]{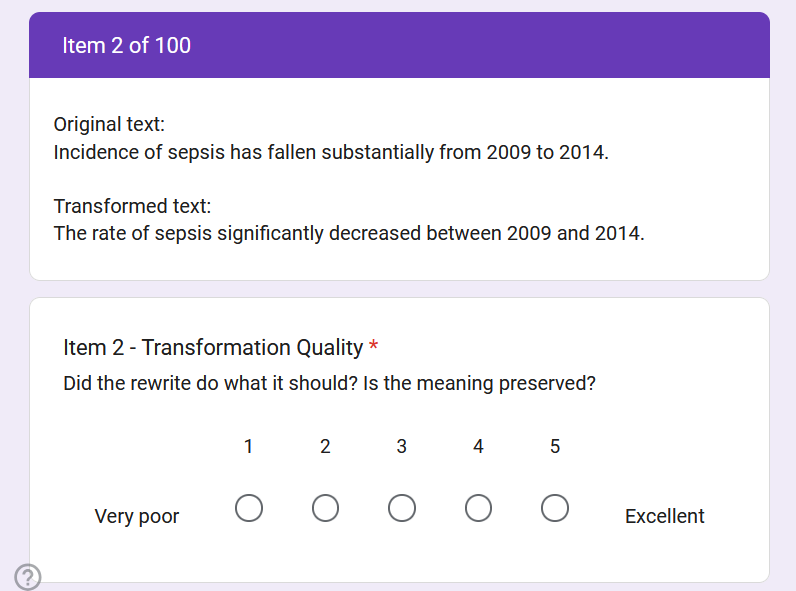}
    \caption{Google Form to collect the human quality rating for a paraphrase.}
    \label{fig:google_form}
\end{figure}

\begin{table}[t]
\centering
\small
\begin{tabularx}{\columnwidth}{
    @{}
    >{\raggedright\arraybackslash}p{0.48\columnwidth}
    >{\raggedright\arraybackslash}X
    @{}
}
\toprule
\textbf{Transformation / target}
& \textbf{MTurk location filter (ISO country codes)} \\
\midrule
Backtranslation
& AU, IE, GB, US \\
Paraphrasing
& AU, IE, GB, US \\
Style Change
& AU, IE, GB, US \\
Expansion
& AU, IE, GB, US \\
Summarisation
& AU, IE, GB, US \\
Summ.\ Exp.
& AU, IE, GB, US \\
(Cross-)Translation AR
& DZ, EG, MA, TN \\
(Cross-)Translation DE
& DE \\
(Cross-)Translation ES
& AR, CL, CO, MX, PE, ES, US \\
(Cross-)Translation FR
& BE, CA, FR \\
(Cross-)Translation TR
& TR \\
\bottomrule
\end{tabularx}
\caption{MTurk location filters used for each transformation and target language.}
\label{tab:mturk_location_filters}
\end{table}

\paragraph{Spanish-source evaluation.} We sampled 20 individual sentences and their transformed versions for each of the six non-language transformations. For the language axis, we sampled 20 sentences for each of the four non-Spanish target languages: Arabic, French, German, and Turkish. Each item was independently rated by three annotators using the same transformation-quality and fluency criteria and the same 1-to-5 scales as in the main human evaluation survey. We also applied the same annotator-eligibility and response-quality checks as in the main study. For this study 1,200 individual human ratings were collected (600 quality ratings + 600 fluency ratings). Given its restriction to Spanish, we treat this study as exploratory with respect to the broader application of HTEB.

\paragraph{STS-B label-preservation study.} We sampled 50 unique original sentence pairs from STS-B. From these sentence pairs, we created 140 transformed sentence pairs using the following transformations: Paraphrasing, Summarised Expansion, or Translation (including Arabic, French, German, and Spanish as target languages). Paraphrasing and Summarised Expansion were applied to all 50 unique original sentence pairs and translation to 40 such sentence pairs. This resulted in matched pairs of original and transformed sentence pairs. Each worker then rated a shuffled set of such matched sentence pairs for a given transformation so that the similarity ratings of original sentence pairs and transformed sentence pairs were generated under comparable conditions (that is, each worker rated an original sentence pair and the transformations of that sentence pair). For each sentence pair, ratings from two to three MTurk workers were collected. This resulted in 790 individual similarity ratings (395 ratings of original sentence pairs + ratings of 395 matched transformed sentence pairs). Detailed results can be found in \autoref{tab:construct_validity_detailed_results}. Due to the limited size and scope, we note that this is an exploratory survey with regard to direct evidence of label preservation on HTEB's datasets through its transformations.

\begin{table*}[t]
\centering
\small
\setlength{\tabcolsep}{6pt}
\begin{tabular*}{\textwidth}{@{\extracolsep{\fill}}lccc@{}}
\toprule
\textbf{Transformation} & \textbf{Transformed} & \textbf{Original} & \textbf{Gold} \\
\midrule
Paraphrasing & 2.28 & 2.36 & 2.37 \\
Summ.\ Expansion & 2.75 & 2.81 & 2.37 \\
Translation & 2.39 & 2.49 & 2.38 \\
\midrule
\multicolumn{4}{l}{\textbf{Pairwise mean absolute difference / source pairs within one point}} \\
\addlinespace[2pt]
\textbf{Transformation} & \textbf{Transformed vs. Original} & \textbf{Transformed vs. Gold} & \textbf{Original vs. Gold} \\
\midrule
Paraphrasing & 0.65 / 88\% & 0.74 / 84\% & 0.79 / 82\% \\
Summ.\ Expansion & 0.41 / 96\% & 0.81 / 68\% & 0.85 / 72\% \\
Translation & 0.45 / 90\% & 1.07 / 65\% & 0.99 / 67.5\% \\
\bottomrule
\end{tabular*}
\caption{Detailed construct-validity results for the STS-B label-preservation study. The upper section reports equal-weight means of source-pair consensus ratings on the 0--5 scale. In the lower section, each cell reports the source-pair mean absolute difference followed by the percentage of source pairs differing by no more than one point. Translation is pooled across Arabic, French, German, and Spanish.}
\label{tab:construct_validity_detailed_results}
\end{table*}

\section{Hardware}
\label{app_hardware}
All experiments were conducted on the following hardware: 2 x NVIDIA RTX 5090 GPUs, AMD Ryzen 9 9950X CPU, 64GB DDR5-RAM.

\section{Embedding Models}
\label{app_emb_models}

\autoref{tab:embedding_models} provides an overview of all embedding models used in the core HTEB evaluation with references (excluding the models in the size ablation). The revision IDs for all checkpoints (including the models in the size ablation) are listed in \autoref{tab:embedding_model_revision_ids}. While for some models we use short names in the main text, these tables include the full model and checkpoints names respectively. Evaluations were conducted using the following random seeds: 1337, 1338, 1339. We used custom prompts for different tasks and models based on small-scale pilot runs and taking the recommendations from the model card into considerations if available.

\begin{table*}[t]
    \centering
    \small
    \newcommand{\blt}{\makebox[1em][l]{\textbullet}}
    \begin{tabularx}{\textwidth}{>{\raggedright\arraybackslash}X >{\raggedright\arraybackslash}p{0.16\textwidth} | >{\raggedright\arraybackslash}p{0.16\textwidth} >{\raggedright\arraybackslash}X}
        \toprule
        \multicolumn{2}{>{\centering\arraybackslash\columncolor{englishbluebg}}p{0.465\textwidth}|}{\textcolor{englishblue}{\textbf{English}}} & \multicolumn{2}{>{\centering\arraybackslash\columncolor{multiorangebg}}p{0.48\textwidth}}{\textcolor{multiorange}{\textbf{Multilingual}}} \\
        \midrule
        \multicolumn{1}{>{\raggedright\arraybackslash}X|}{%
            \blt All-MiniLM-L12-v2~\citep{SBERT_Reimers2019}\newline
            \blt All-MPNet-Base-v2~\citep{SBERT_Reimers2019}\newline
            \blt MXBAI-Embed-Large-v1~\citep{mxbai_embed_large_v1_1,mxbai-embed-large-v1_2}\newline
            \blt Stella-EN-400M-v5~\citep{Zhang2025_JasperStella}\newline
            \blt Jasper-Token-Compression-600M~\citep{JasperTokenCompression600M}\newline
            \blt E5-Mistral-7B-Instruct~\citep{e5-mistral-7b-instruct_2024a, Wang2024_e5_7b_ImprovingTextEmbeddingsLargeLanguageModels}\newline
            \blt NV-Embed-v2~\citep{Lee2025_NVEmbed}%
        } &
        \multicolumn{2}{>{\raggedright\arraybackslash}p{0.35\textwidth}|}{%
            \blt Jina-Embeddings-v5-Text-Small~\citep{jina-embeddings-v5_v2}\newline
            \blt F2LLM-v2-4B~\citep{f2llm-v2}\newline
            \blt Llama-Nemotron-Embed-8B~\citep{llama_nemotron_embed_8} \newline
            \blt Qwen3-Emb-8B~\citep{Qwen3Embedding}%
        } &
        \blt Paraphrase-Multilingual-MPNet-base-v2~\citep{SBERT_Reimers2019}\newline
        \blt Harrier-OSS-270M~\citep{Harrier-OSS-270M}\newline
        \blt Multilingual-E5-Large-Instruct~\citep{Wang2024_e5_MultilingualE5Text}\newline
        \blt BGE-M3~\citep{Chen2024_M3_embedding_ACL}\newline
        \blt GTE-Multilingual-Base~\citep{zhang-etal-2024-mgte}
    \end{tabularx}
    \caption{Embedding models evaluated in core HTEB evaluation (e.g., excluding model size ablation). Centre column lists models used in both \textcolor{englishblue}{\textbf{English}} and \textcolor{multiorange}{\textbf{Multilingual}} experiments. Note that this table does not specify whether a model was trained English or non-English data.}
    \label{tab:embedding_models}
\end{table*}

\begin{table*}[t]
\centering
\scriptsize
\setlength{\tabcolsep}{2pt}
\begin{tabular}{@{}p{0.58\textwidth}p{0.38\textwidth}@{}}
\toprule
\textbf{Model checkpoint} & \textbf{Revision ID} \\
\midrule
\href{https://huggingface.co/Alibaba-NLP/gte-multilingual-base}{\texttt{Alibaba-\allowbreak{}NLP/\allowbreak{}gte-\allowbreak{}multilingual-\allowbreak{}base}} & \texttt{9bbca17d\allowbreak{}9273fd0d\allowbreak{}03d5725c\allowbreak{}7a4b0f6b\allowbreak{}45142062} \\
\href{https://huggingface.co/BAAI/bge-m3}{\texttt{BAAI/\allowbreak{}bge-\allowbreak{}m3}} & \texttt{5617a9f6\allowbreak{}1b028005\allowbreak{}a4858fda\allowbreak{}c845db40\allowbreak{}6aefb181} \\
\href{https://huggingface.co/codefuse-ai/F2LLM-v2-0.6B}{\texttt{codefuse-\allowbreak{}ai/\allowbreak{}F2LLM-\allowbreak{}v2-\allowbreak{}0.\allowbreak{}6B}} & \texttt{2b415909\allowbreak{}1278275a\allowbreak{}3b00d2c3\allowbreak{}9095754d\allowbreak{}59d7d7de} \\
\href{https://huggingface.co/codefuse-ai/F2LLM-v2-1.7B}{\texttt{codefuse-\allowbreak{}ai/\allowbreak{}F2LLM-\allowbreak{}v2-\allowbreak{}1.\allowbreak{}7B}} & \texttt{3585dd8f\allowbreak{}7593a4fd\allowbreak{}c554f8c6\allowbreak{}01c31f59\allowbreak{}7c002409} \\
\href{https://huggingface.co/codefuse-ai/F2LLM-v2-14B}{\texttt{codefuse-\allowbreak{}ai/\allowbreak{}F2LLM-\allowbreak{}v2-\allowbreak{}14B}} & \texttt{1284bb8b\allowbreak{}8995abc0\allowbreak{}9d3ee8a7\allowbreak{}0bc2e88c\allowbreak{}9854fd35} \\
\href{https://huggingface.co/codefuse-ai/F2LLM-v2-160M}{\texttt{codefuse-\allowbreak{}ai/\allowbreak{}F2LLM-\allowbreak{}v2-\allowbreak{}160M}} & \texttt{60229594\allowbreak{}f9498ae4\allowbreak{}4c553f1f\allowbreak{}8cebf325\allowbreak{}59bc8577} \\
\href{https://huggingface.co/codefuse-ai/F2LLM-v2-330M}{\texttt{codefuse-\allowbreak{}ai/\allowbreak{}F2LLM-\allowbreak{}v2-\allowbreak{}330M}} & \texttt{1239cdd5\allowbreak{}44b24c24\allowbreak{}7ed75df2\allowbreak{}ae22e5a4\allowbreak{}01ac4659} \\
\href{https://huggingface.co/codefuse-ai/F2LLM-v2-4B}{\texttt{codefuse-\allowbreak{}ai/\allowbreak{}F2LLM-\allowbreak{}v2-\allowbreak{}4B}} & \texttt{2c0231f9\allowbreak{}3661c526\allowbreak{}0cb71efb\allowbreak{}2097a3f7\allowbreak{}210b5996} \\
\href{https://huggingface.co/codefuse-ai/F2LLM-v2-80M}{\texttt{codefuse-\allowbreak{}ai/\allowbreak{}F2LLM-\allowbreak{}v2-\allowbreak{}80M}} & \texttt{f4a16a11\allowbreak{}c9f5c8c7\allowbreak{}e2269465\allowbreak{}3de6ce75\allowbreak{}430f4538} \\
\href{https://huggingface.co/codefuse-ai/F2LLM-v2-8B}{\texttt{codefuse-\allowbreak{}ai/\allowbreak{}F2LLM-\allowbreak{}v2-\allowbreak{}8B}} & \texttt{03c66983\allowbreak{}47b4115c\allowbreak{}4a4500a5\allowbreak{}e7368cca\allowbreak{}8ea518f5} \\
\href{https://huggingface.co/infgrad/Jasper-Token-Compression-600M}{\texttt{infgrad/\allowbreak{}Jasper-\allowbreak{}Token-\allowbreak{}Compression-\allowbreak{}600M}} & \texttt{e1373068\allowbreak{}c35fba2a\allowbreak{}37568da4\allowbreak{}eb410537\allowbreak{}7e17eae9} \\
\href{https://huggingface.co/intfloat/e5-mistral-7b-instruct}{\texttt{intfloat/\allowbreak{}e5-\allowbreak{}mistral-\allowbreak{}7b-\allowbreak{}instruct}} & \texttt{d62cc441\allowbreak{}7be0e328\allowbreak{}9c03dd45\allowbreak{}9d1ef2d1\allowbreak{}37e2554e} \\
\href{https://huggingface.co/intfloat/multilingual-e5-large-instruct}{\texttt{intfloat/\allowbreak{}multilingual-\allowbreak{}e5-\allowbreak{}large-\allowbreak{}instruct}} & \texttt{274baa43\allowbreak{}b0e13e37\allowbreak{}fafa6428\allowbreak{}dbc7938e\allowbreak{}62e5c439} \\
\href{https://huggingface.co/jinaai/jina-embeddings-v5-text-small}{\texttt{jinaai/\allowbreak{}jina-\allowbreak{}embeddings-\allowbreak{}v5-\allowbreak{}text-\allowbreak{}small}} & \texttt{dd76d535\allowbreak{}f5447ca3\allowbreak{}897a9c89\allowbreak{}3fb1e612\allowbreak{}ead98192} \\
\href{https://huggingface.co/microsoft/harrier-oss-v1-0.6b}{\texttt{microsoft/\allowbreak{}harrier-\allowbreak{}oss-\allowbreak{}v1-\allowbreak{}0.\allowbreak{}6b}} & \texttt{f9b9dc8d\allowbreak{}367d443f\allowbreak{}2479d27a\allowbreak{}a5d8d285\allowbreak{}0c0774ee} \\
\href{https://huggingface.co/microsoft/harrier-oss-v1-270m}{\texttt{microsoft/\allowbreak{}harrier-\allowbreak{}oss-\allowbreak{}v1-\allowbreak{}270m}} & \texttt{31de22b6\allowbreak{}73913c7d\allowbreak{}658c0f03\allowbreak{}f792d77c\allowbreak{}2dcf8ebd} \\
\href{https://huggingface.co/microsoft/harrier-oss-v1-27b}{\texttt{microsoft/\allowbreak{}harrier-\allowbreak{}oss-\allowbreak{}v1-\allowbreak{}27b}} & \texttt{0c0fc62f\allowbreak{}6d8af9e8\allowbreak{}604cb818\allowbreak{}c412301b\allowbreak{}103a0093} \\
\href{https://huggingface.co/mixedbread-ai/mxbai-embed-large-v1}{\texttt{mixedbread-\allowbreak{}ai/\allowbreak{}mxbai-\allowbreak{}embed-\allowbreak{}large-\allowbreak{}v1}} & \texttt{b33106f5\allowbreak{}85b9ce46\allowbreak{}904ad744\allowbreak{}3a3b52b7\allowbreak{}a63e231c} \\
\href{https://huggingface.co/NovaSearch/stella_en_400M_v5}{\texttt{NovaSearch/\allowbreak{}stella\_\allowbreak{}en\_\allowbreak{}400M\_\allowbreak{}v5}} & \texttt{ffeb2b7e\allowbreak{}e715c226\allowbreak{}d4ffe5e4\allowbreak{}619f7dbb\allowbreak{}48624c20} \\
\href{https://huggingface.co/nvidia/llama-embed-nemotron-8b}{\texttt{nvidia/\allowbreak{}llama-\allowbreak{}embed-\allowbreak{}nemotron-\allowbreak{}8b}} & \texttt{aa3b43a4\allowbreak{}95a9b280\allowbreak{}d1bdb716\allowbreak{}da37c54b\allowbreak{}b495d630} \\
\href{https://huggingface.co/nvidia/NV-Embed-v2}{\texttt{nvidia/\allowbreak{}NV-\allowbreak{}Embed-\allowbreak{}v2}} & \texttt{3fa59658\allowbreak{}547db50a\allowbreak{}1e8e3346\allowbreak{}cf057fd0\allowbreak{}c77ed6ef} \\
\href{https://huggingface.co/Qwen/Qwen3-Embedding-0.6B}{\texttt{Qwen/\allowbreak{}Qwen3-\allowbreak{}Embedding-\allowbreak{}0.\allowbreak{}6B}} & \texttt{c54f2e6e\allowbreak{}80b2d7b7\allowbreak{}de06f51c\allowbreak{}ec4959f6\allowbreak{}b3e03418} \\
\href{https://huggingface.co/Qwen/Qwen3-Embedding-4B}{\texttt{Qwen/\allowbreak{}Qwen3-\allowbreak{}Embedding-\allowbreak{}4B}} & \texttt{5cf2132a\allowbreak{}bc99cad0\allowbreak{}20ac570b\allowbreak{}19d031ef\allowbreak{}ec650f2b} \\
\href{https://huggingface.co/Qwen/Qwen3-Embedding-8B}{\texttt{Qwen/\allowbreak{}Qwen3-\allowbreak{}Embedding-\allowbreak{}8B}} & \texttt{1d8ad4ca\allowbreak{}9b3dd805\allowbreak{}9ad90a75\allowbreak{}d4983776\allowbreak{}a23d44af} \\
\href{https://huggingface.co/sentence-transformers/all-MiniLM-L12-v2}{\texttt{sentence-\allowbreak{}transformers/\allowbreak{}all-\allowbreak{}MiniLM-\allowbreak{}L12-\allowbreak{}v2}} & \texttt{a50ef001\allowbreak{}43b4d539\allowbreak{}1434df20\allowbreak{}ae116325\allowbreak{}88ac25be} \\
\href{https://huggingface.co/sentence-transformers/all-mpnet-base-v2}{\texttt{sentence-\allowbreak{}transformers/\allowbreak{}all-\allowbreak{}mpnet-\allowbreak{}base-\allowbreak{}v2}} & \texttt{e8c3b32e\allowbreak{}df5434bc\allowbreak{}2275fc9b\allowbreak{}ab85f826\allowbreak{}40a19130} \\
\href{https://huggingface.co/sentence-transformers/paraphrase-multilingual-mpnet-base-v2}{\texttt{sentence-\allowbreak{}transformers/\allowbreak{}paraphrase-\allowbreak{}multilingual-\allowbreak{}mpnet-\allowbreak{}base-\allowbreak{}v2}} & \texttt{4328cf26\allowbreak{}390c98c5\allowbreak{}e3c738b4\allowbreak{}460a05b9\allowbreak{}5f4911f5} \\
\bottomrule
\end{tabular}
\caption{Embedding model checkpoint revision IDs.}
\label{tab:embedding_model_revision_ids}
\end{table*}

\section{Prompts}
\label{app_prompts}

In the following we provide the verbatim prompts used to generate each of the eight text transformations evaluated in HTEB. The placeholder \texttt{\{target\_language\}} is substituted at inference time with the name of the target language (e.g.\ "English", "Spanish"). While we did not systematically optimise prompts, we tested different prompts on a hold-out dataset. In some cases, this resulted in model-specific prompts. This section only includes the prompts used for the selected HTEB transformation model Gemma-3-27B-int4-AWQ.

\subsection{Single-step transformations}

This section includes the prompts that are used for transformations that require a single LLM call. 

\begin{promptbox}{Paraphrasing}
\ttfamily\small
Rephrase the following text while keeping its original meaning. IMPORTANT: You MUST answer in \{target\_language\}. Do NOT translate the text to English or any other language. Only reply with the paraphrased text and only provide a single response -- no alternatives! Do not include explanations or notes.
\end{promptbox}

\begin{promptbox}{Style Change}
\ttfamily\small
Change the style of the following text. If the text is casual or informal, rewrite it in a formal style. If the text is formal or scientific, rewrite it in a casual or informal style. Preserve the original meaning while changing only the tone, style, and vocabulary. IMPORTANT: You MUST answer in \{target\_language\}. Do NOT translate the text to English or any other language. Only reply with the rewritten text -- no explanations or notes!
\end{promptbox}

\begin{promptbox}{Expansion}
\ttfamily\small
Expand the following text by adding more details, context, and elaboration while preserving the same core meaning. IMPORTANT: If the input is a question, your output must also be a question -- do NOT answer it. If the input is a statement, your output must also be a statement. IMPORTANT: You MUST answer in \{target\_language\}. Do NOT translate the text to English or any other language. Use similar words and maintain the original message. Only reply with the expanded text -- no explanations or notes!
\end{promptbox}

\begin{promptbox}{Summarise}
\ttfamily\small
Your task is to SUMMARIZE the text below -- make it shorter while keeping the same meaning. DO NOT change the sentence type:
\\
- If input is a STATEMENT, output must be a STATEMENT
\\
- If input is a QUESTION, output must be a QUESTION (do NOT answer it!)
\\[4pt]
Examples:
\\
- Statement `I loved the movie' \textrightarrow{} `Enjoyed the movie' (NOT `Did you enjoy the movie?')
\\
- Question `Where is my card?' \textrightarrow{} `Card location?' (NOT `Check your wallet')
\\[4pt]
IMPORTANT: You MUST answer in \{target\_language\}. Do NOT translate the text to English or any other language. Only output the summary, nothing else.
\\[4pt]
Text to summarize:
\end{promptbox}

\begin{promptbox}{Translation}
\ttfamily\small
Translate the following text to \{target\_language\}. Provide only the translation -- no explanations or notes!
\end{promptbox}

\begin{promptbox}{Cross-Translation}
\ttfamily\small
Translate the following text to \{target\_language\}. Provide only the translation -- no explanations or notes!
\end{promptbox}

For Translation, the target language is drawn deterministically from \{Spanish, French, German, Turkish, Arabic\}; if the source language is known it is excluded from the candidate set. For a given run, the same target language is used for every example in the batch. For Cross-Translation, the prompt template is identical to translation, but the target language is resampled \emph{per text}. 

\subsection{Multi-step transformations}

This section contains the prompts for transformations that use two consecutive LLM calls.

\paragraph{Backtranslation.} The input is translated into a
random intermediate language drawn from
\{English, Spanish, French, German, Turkish, Arabic\} (excluding the
source language), then translated back into the source language.

\begin{promptbox}{Backtranslation -- Step 1 (forward translation)}
\ttfamily\small
Translate the following text to \{target\_language\}. Provide only the translation -- no explanations or notes!\\[4pt]
\textrm{\textit{\{target\_language\} is the randomly chosen
intermediate language.}}
\end{promptbox}

\begin{promptbox}{Backtranslation -- Step 2 (backward translation)}
\ttfamily\small
Translate the following text to \{target\_language\}. Provide only the translation -- no explanations or notes!\\[4pt]
\textrm{\textit{\{target\_language\} is the original source
language of the input.}}
\end{promptbox}

\paragraph{Summarised Expansion.} The input is first expanded with the
\emph{Expansion} prompt, then summarised with the \emph{Summarise}
prompt. The second call takes the output of the first as its input.

\begin{promptbox}{Summarised Expansion -- Step 1 (Expansion)}
\ttfamily\small
Expand the following text by adding more details, context, and elaboration while preserving the same core meaning. IMPORTANT: If the input is a question, your output must also be a question -- do NOT answer it. If the input is a statement, your output must also be a statement. IMPORTANT: You MUST answer in \{target\_language\}. Do NOT translate the text to English or any other language. Use similar words and maintain the original message. Only reply with the expanded text -- no explanations or notes!
\end{promptbox}

\begin{promptbox}{Summ. Expansion -- Step 2 (Summarisation)}
\ttfamily\small
Your task is to SUMMARIZE the text below -- make it shorter while keeping the same meaning. DO NOT change the sentence type:
\\
- If input is a STATEMENT, output must be a STATEMENT
\\
- If input is a QUESTION, output must be a QUESTION (do NOT answer it!)
\\[4pt]
Examples:
\\
- Statement `I loved the movie' \textrightarrow{} `Enjoyed the movie' (NOT `Did you enjoy the movie?')
\\
- Question `Where is my card?' \textrightarrow{} `Card location?' (NOT `Check your wallet')
\\[4pt]
IMPORTANT: You MUST answer in \{target\_language\}. Do NOT translate the text to English or any other language. Only output the summary, nothing else.
\\[4pt]
Text to summarize:
\end{promptbox}

\section{Datasets}
\label{app_datasets}
\autoref{tab:english_datasets} and \autoref{tab:multilingual_datasets} provide an overview of all datasets. Note that for Reranking and Retrieval only the queries are transformed which reflects the common use case of varying user requests towards a relatively more stable knowledge base. Moreover, transforming documents would drastically increase the runtime of HTEB. 

\begin{table*}[ht]
    \small
    \centering
    \begin{tabularx}{\textwidth}{p{1.7cm}>{\raggedright\arraybackslash}X>{\raggedright\arraybackslash}p{2.7cm}c}
        \toprule
        \textbf{Task} & \textbf{Dataset} & \textbf{Domains} & \textbf{Languages} \\
        \midrule
        \multirow{3}{*}[0pt]{Classification}
            & AmazonCounterfactualClassification \citep{AmazonCounterFactuals_Classification} & Product reviews & eng \\
            & Banking77Classification \citep{Casanueva2020_Banking77Classification} & Finance & eng \\
            & MassiveIntentClassification \citep{FitzGerald2023_MASSIVE1MExampleMultilingualNaturalLanguageUnderstandingDataset} & Voice assistant & eng \\
        \cmidrule(lr){1-4}
        \multirow{3}{*}[0pt]{Clustering}
            & BiorxivClusteringS2S \citep{Enevoldsen2024_MMTEB_ICLR} & Academic (biology) & eng \\
            & MedrxivClusteringS2S.v2 \citep{Enevoldsen2024_MMTEB_ICLR} & Academic (medical) & eng \\
            & TwentyNewsgroupsClustering.v2 \citep{Lang1995_TwentyNewsGroupsClustering} & News & eng \\
        \cmidrule(lr){1-4}
        \multirow{3}{*}[0pt]{\shortstack[lt]{Pair\\Classification}}
            & SprintDuplicateQuestions \citep{Shah2018_SprintDuplicateQuestions} & Programming & eng \\
            & TwitterSemEval2015 \citep{Xu2015_TwitterSemEval2015PairClassification} & Social media & eng \\
            & TwitterURLCorpus \citep{lan-etal-2017-continuously} & Social media & eng \\
        \cmidrule(lr){1-4}
        \multirow{3}{*}[0pt]{Reranking}
            & AskUbuntuDupQuestions \citep{Wang2021_TSDAE_AskUbuntuDupQPair_Reranking} & Q\&A forum & eng \\
            & SciDocsRR \citep{Cohan2020_SciDocsRR} & Academic & eng \\
            & StackOverflowDupQuestions \citep{Liu2018_StackOverflowDupQuestions} & Programming & eng \\
        \cmidrule(lr){1-4}
        \multirow{3}{*}[0pt]{Retrieval}
            & ArguAna \citep{wachsmuth-etal-2018-retrieval} & Argumentative essays & eng \\
            & CQADupstackGamingRetrieval \citep{Hoogeveen2015_CQADupStack} & Q\&A forum & eng \\
            & SciFact \citep{Wadden2020_SciFact} & Academic & eng \\
        \cmidrule(lr){1-4}
        STR & SemRel2024 \citep{Ousidhoum_SemRel2024} & News, social media & eng \\
        \cmidrule(lr){1-4}
        \multirow{2}{*}[0pt]{STS}
            & BIOSSES \citep{Sogancıoglu2017_BIOSSES} & Academic (biomedical) & eng \\
            & STS-B \citep{SemEval2017_Task1} & News, blog & eng \\
        \cmidrule(lr){1-4}
        Summarisation & SummEval \citep{fabbri-etal-2021-summeval} & News & eng \\
        \bottomrule
    \end{tabularx}
    \caption{English datasets used in HTEB. Language codes follow ISO 639-3 (see \autoref{tab:language_codes}). Please refer to \citet{Muennighoff2023_MTEB_EACL} and \citet{Enevoldsen2024_MMTEB_ICLR} for further details on the datasets.}
    \label{tab:english_datasets}
\end{table*}

\begin{table*}[ht]
    \small
    \centering
    \begin{tabularx}{\textwidth}{p{1.7cm}>{\raggedright\arraybackslash}p{4.7cm}>{\raggedright\arraybackslash}p{2.5cm}>{\raggedright\arraybackslash}X}
        \toprule
        \textbf{Task} & \textbf{Dataset} & \textbf{Domains} & \textbf{Languages} \\
        \midrule
        \multirow{2}{*}[0pt]{Classification}
            & IsiZuluNewsClassification \citep{madodonga_izindaba_tindzaba_2023} & News & zul \\
            & SentimentAnalysisHindi \citep{Parida2023_SentimentAnalysisHindi} & Social media & hin \\
        \cmidrule(lr){1-4}
        \multirow{2}{*}[0pt]{Clustering}
            & MasakhaNEWSClusteringS2S \citep{Adelani2023_MasakhaNEWS} & News & amh, eng, fra, hau, ibo, lin, lug, orm, pcm, run, sna, som, swa, tir, xho, yor \\
            & PlscClusteringS2S.v2 \citep{PlscClusteringS2S} & Academic & pol \\
        \cmidrule(lr){1-4}
        \multirow{2}{*}[0pt]{\shortstack[lt]{Pair\\Classification}}
            & RTE3 \citep{Giampiccolo2007_RTE3} & News, encyclopedic & deu, eng, fra, ita \\
            & IndonLI \citep{mahendra-etal-2021-indonli} & General & ind \\
        \cmidrule(lr){1-4}
        \multirow{2}{*}[0pt]{Reranking}
            & VoyageMMarcoReranking \citep{Clavie2024_VoyageMMarcoReranking} & Web & jpn \\ 
            & NamaaMrTydiReranking \citep{Enevoldsen2024_MMTEB_ICLR} & Web & ara \\ 
        \cmidrule(lr){1-4}
        \multirow{2}{*}[0pt]{Retrieval}
            & TwitterHjerneRetrieval \citep{Holm2025_TwitterHernje} & Social media & dan \\
            & Ko-StrategyQA \citep{geva-etal-2021-aristotle} & Q\&A & kor \\
        \cmidrule(lr){1-4}
        STR & SemRel2024 \citep{Ousidhoum_SemRel2024} & News, social media & afr, amh, arb, arq, ary, eng, hau, hin, ind, kin, mar, pan, spa, tel \\
        \cmidrule(lr){1-4}
        \multirow{2}{*}[0pt]{STS}
            & STS17 \citep{SemEval2017_Task1} & News, web & ara, deu, eng, fra, ita, kor, nld, spa, tur \\
            & IndicCrosslingualSTS \citep{Ramesh2022_IndicCrosslingual} & General & asm, ben, eng, guj, hin, kan, mal, mar, ory, pan, tam, tel, urd \\
        \bottomrule
    \end{tabularx}
    \caption{Multilingual datasets used in HTEB. Language codes follow ISO 639-3 (see \autoref{tab:language_codes}). Please refer to \citet{Muennighoff2023_MTEB_EACL} and \citet{Enevoldsen2024_MMTEB_ICLR} for further details on the datasets.}
    \label{tab:multilingual_datasets}
\end{table*}

\section{Languages}
See \autoref{tab:language_codes} for a full list of languages and their respective language codes. Languages are abbreviated using ISO 639-3 codes\footnote{\url{https://www.loc.gov/standards/iso639-3/php/code_list.php}}.

\begin{table*}[t]
    \centering
    \footnotesize
    \setlength{\tabcolsep}{3pt}
    \renewcommand{\arraystretch}{0.92}
    \begin{tabular}{@{}ll@{\hspace{0.8em}}!{\vrule width 0.4pt}@{\hspace{0.8em}}ll@{\hspace{0.8em}}!{\vrule width 0.4pt}@{\hspace{0.8em}}ll@{\hspace{0.8em}}!{\vrule width 0.4pt}@{\hspace{0.8em}}ll@{}}
        \toprule
        \textbf{Code} & \textbf{Language} &
        \textbf{Code} & \textbf{Language} &
        \textbf{Code} & \textbf{Language} &
        \textbf{Code} & \textbf{Language} \\
        \midrule
        afr & Afrikaans          & guj & Gujarati      & lug & Luganda          & som & Somali \\
        amh & Amharic           & hau & Hausa         & mal & Malayalam        & spa & Spanish \\
        ara & Arabic            & hin & Hindi         & mar & Marathi          & swa & Swahili \\
        arb & Arabic (MSA)      & ibo & Igbo          & nld & Dutch            & tam & Tamil \\
        arq & Arabic (Algerian) & ind & Indonesian    & orm & Oromo            & tel & Telugu \\
        ary & Arabic (Moroccan) & ita & Italian       & ory & Odia             & tir & Tigrinya \\
        asm & Assamese          & jpn & Japanese      & pan & Punjabi          & tur & Turkish \\
        ben & Bengali           & kan & Kannada       & pcm & Nigerian Pidgin  & urd & Urdu \\
        dan & Danish            & kin & Kinyarwanda   & pol & Polish           & xho & isiXhosa \\
        deu & German            & kor & Korean        & run & Rundi            & yor & Yorùbá \\
        eng & English           & lin & Lingala       & sna & Shona            & zul & Zulu \\
        fra & French            &     &               &     &                  &     & \\
        \midrule
        \multicolumn{8}{@{}l}{\textbf{Total: 45 codes (42 languages with Arabic variants grouped)}} \\
        \bottomrule
    \end{tabular}
    \caption{Languages and ISO 639-3 language codes used in HTEB's datasets.}
    \label{tab:language_codes}
\end{table*}

\section{Metrics}
\label{app_metrics}

HTEB uses the following metrics which are the main metrics used by MTEB:

\begin{itemize}
    \item \textbf{Classification}: Accuracy
    \item \textbf{Clustering}: V-measure
    \item \textbf{Pair Classification}: Average Precision (AP)
    \item \textbf{Reranking}: MAP (Mean Average Precision)
    \item \textbf{Retrieval}: nDCG@10 (Normalised Discounted Cumulative Gain at 10)
    \item \textbf{STS}: Spearman's $\rho$
    \item \textbf{STR}: Spearman's $\rho$
    \item \textbf{Summarisation}: Spearman's $\rho$
\end{itemize}

\section{Transformation Examples}
\label{app_examples}

\autoref{tab:transformation_examples} provides a randomly selected example for each transformation.  
\begin{table*}[htbp]
    \centering
    \small
    \begin{tabular}{p{3cm}p{6cm}p{6cm}}
        \toprule
         & \textbf{Sentence 1} & \textbf{Sentence 2} \\
        \midrule
        \textbf{Original} & A man is singing and playing a guitar. & A man is playing a guitar. \\
        \midrule
        \textbf{Paraphrasing} & Someone is performing music by both vocalizing and strumming a guitar. & Someone is strumming a guitar. \\
        \midrule
        \textbf{Backtranslation} & A man sings and plays guitar. & A man is playing guitar. \\
        \midrule
        \textbf{Expansion} & A solitary figure, a man immersed in melody, is presently engaged in both vocal performance and instrumental accompaniment upon an acoustic guitar. His voice carries through the air, perhaps filled with joy, sorrow, or simply storytelling, weaving tales only he truly understands. The gentle strumming of his six strings resonates with each note sung, creating a harmonious blend that draws attention – whether intentionally sought or merely offered to the passing breeze. He might be performing for an audience, large or small, or lost in private contemplation, serenading himself under the open sky. Perhaps he's busking in a bustling city square, hoping to earn a few coins, or nestled comfortably within the walls of a cozy tavern, providing entertainment for patrons enjoying their evening. & A solitary man sits immersed in music, his fingers deftly moving across the fretboard of an acoustic guitar. Sunlight streams through a nearby window, illuminating dust motes dancing in the air around him as he strums chords – perhaps melancholic blues riffs or bright, cheerful folk melodies. His eyes are closed in concentration, lost in the world created by the vibrations emanating from the instrument's soundhole. It's unclear whether he plays for himself, seeking solace in the rhythm, or if he hopes to share his musical expression with others; perhaps he practices diligently preparing for a performance in front of an audience someday. \\
        \midrule
        \textbf{Summarisation} & Man sings with guitar. & Man plays guitar. \\
        \midrule
        \textbf{Summarised Expansion} & A musician sings and plays guitar, sharing stories through sound in a peaceful moment. & A man plays guitar in quiet contemplation, lost in music within sunlit solitude. \\
        \midrule
        \textbf{Translation} & Ein Mann singt und spielt Gitarre. & Ein Mann spielt Gitarre. \\
        \midrule
        \textbf{Cross-Translation} & Ein Mann singt und spielt Gitarre. & Un hombre está tocando la guitarra. \\
        \midrule
        \textbf{Style Change} & An individual is vocalizing musical notes accompanied by strumming a six-string instrument. & An individual is engaged in performing music on a guitar. \\
        \bottomrule
    \end{tabular}
    \caption{Transformations for an STS-B sentence pair. For STS, cross-translation translates each sentence to a different target language.}
    \label{tab:transformation_examples}
\end{table*}

\section{Statistical Analysis}
\label{app_statistical_analysis}

\autoref{tab:stats_overview} shows an overview of the applied statistical methods.

\begin{table*}[t]
  \centering
  \small
  \begin{tabular}{p{1.8cm} p{3.3cm} p{2.2cm} p{3.1cm} p{3.2cm}}
  \toprule
  \textbf{Area of Analysis} & \textbf{Analysis} & \textbf{Unit of Observation} &
  \textbf{Aggregation} & \textbf{Statistical Method} \\
  \midrule

  Transformation model selection
  & LLM judge scores across candidate generators (\autoref{tab:llm_judge_by_transformation},
  \autoref{tab:llm_judge_scores})
  & Sample-level judge score per generator
  & For each source sample and generator, mean over paired texts, transformations, and judges
  & Percentile bootstrap 95\% CI for the mean; pairwise paired bootstrap tests with Holm
  correction \\

  \midrule
  \addlinespace

  Human evaluation
  & Quality \& fluency ratings per transformation, with Translation and Cross-Translation pooled
  (\autoref{tab:human_eval_per_language})
  & Individual rater-item rating
  & None before reporting; ratings are pooled within each transformation
  & Mean $\pm$ sample SD \\

  \midrule
  \addlinespace

  Human evaluation
  & Quality \& fluency ratings per robustness axis (\autoref{tab:human_eval_per_language})
  & Individual rater-item rating
  & None before reporting; ratings are pooled by axis membership
  & Mean $\pm$ sample SD \\

  \midrule
  \addlinespace

  Human evaluation
  & Inter-rater agreement (\autoref{tab:human_eval_per_language})
  & Rating per item and rater
  & None; ratings are arranged as item-by-rater matrices within each reporting subset
  & Gwet's AC$_2$ with ordinal weights \\

  \midrule
  \addlinespace

  Human evaluation
  & Embedding drift vs.\ human quality (\autoref{fig:quality_vs_drift})
  & Rated item / text pair
  & Mean quality rating across raters; drift $= 1 - \cos(\mathrm{emb}_{orig}, \mathrm{emb}
  _{trans})$
  & Spearman's $\rho$ with 95\% percentile bootstrap CI \\

  \midrule
  \addlinespace

  Ranking stability
  & Ranking stability per transformation (\autoref{tab:kendall_tau})
  & Dataset-level pair of model score vectors
  & Mean over runs per model-dataset-transformation
  & Kendall's $\tau$; reported as mean $\pm$ sample SD across datasets \\

  \midrule
  \addlinespace

  Ranking stability
  & Ranking stability per axis (\autoref{tab:kendall_tau})
  & Dataset-level pair of model score vectors
  & Mean over runs, then mean transformed score over axis transformations per model-dataset
  & Kendall's $\tau$; reported as mean $\pm$ sample SD across datasets \\

  \midrule
  \addlinespace

  Ranking stability
  & Split-half reliability of per-axis profiles (\autoref{tab:half_split_per_axis_compact})
  & Per-axis model score vector for one dataset split half
  & Mean over runs, datasets in the split half, and transformations within axis
  & Spearman's $\rho$ across models per axis; median over $N{=}1000$ stratified dataset splits \\

  \midrule
  \addlinespace

  Transformation-level comparison
  & Score deltas per transformation (\autoref{fig:hl_forest_colored})
  & Dataset-level paired score difference
  & Mean over embedding models and runs per dataset-transformation
  & Wilcoxon signed-rank test with Holm correction; Hodges--Lehmann shift with 95\% CI from
  Walsh averages \\

  \midrule
  \addlinespace

  Axis-level comparison
  & Absolute score deltas per axis (\autoref{fig:hl_forest_colored})
  & Dataset-transformation absolute paired score difference
  & Mean over embedding models and runs per dataset-transformation
  & Median absolute paired difference, pooled over transformations within axis \\

  \midrule
  \addlinespace

  Axis-level visualization
  & Original, total HTEB, and per-axis HTEB scores (\autoref{fig:english_multilingual_total_results})
  & Embedding-model score
  & Original: mean over datasets. Per-axis HTEB: mean over transformations within each axis, with scores averaged over runs and datasets. Total HTEB: unweighted mean of the three per-axis HTEB scores.
  & Descriptive mean visualization \\

  \bottomrule
  \end{tabular}
  \caption{Overview of statistical analyses in HTEB. The unit of observation is what the statistical method operates on. The aggregation explains how that unit of observation is derived before the statistical method or descriptive summary is computed.}
  \label{tab:stats_overview}
  \end{table*}

\section{Overall HTEB Results}
\label{app_overall_results_detailed}

\autoref{tab:robustness-axis-dumbbell-joint-horizontal-english} and \autoref{tab:robustness-axis-dumbbell-joint-horizontal-multilingual} show the overall results for the English and multilingual evaluation respectively. This is the data underlying \autoref{fig:english_multilingual_total_results}. 

\section{Zero-Shot Model Analysis}
\label{app_zero_shot}

As an exploratory analysis, we used MTEB metadata to classify each evaluated model-dataset combination as zero-shot, non-zero-shot, or unknown with respect to known training overlap, and computed the score drop as the Original score minus the HTEB score. Overlap status was unavailable for 19 of 209 embedding model-dataset combinations in HTEB's English eval datasets and for 13 of 117 combinations in multilingual evaluation. These combinations were excluded from the two status-based comparisons below (valid combinations are denoted as n in the lists below). On the English datasets, non-zero-shot combinations do not show larger drops than zero-shot combinations; their average drop is 0.49 points smaller. When restricting the English-multilingual comparison to zero-shot combinations, English datasets still show substantially larger drops, with a gap of 6.74 points. Thus, the English-multilingual sensitivity gap described in \autoref{sec:discussion_lingual_gap} also appears among combinations marked as zero-shot in MTEB metadata. Because this metadata captures only known benchmark overlap, it does not rule out undocumented fine-tuning or broader pretraining exposure.\\

\noindent\textcolor{englishblue}{\textbf{English:}}
\begin{itemize}
  \item zero-shot drop: 8.84 (n=157)
  \item non-zero-shot drop: 8.35 (n=33)
  \item Difference (non-zero-shot minus zero-shot): -0.49.
\end{itemize}

\noindent\textcolor{englishblue}{\textbf{English}} vs. \textcolor{multiorange}{\textbf{Multilingual}:}
\begin{itemize}
  \item English zero-shot drop: 8.84 (n=157)
  \item Multilingual zero-shot drop: 2.10 (n=103)
  \item Difference (English minus multilingual): 6.74.
\end{itemize}

\section{Within-Transformation Kendall’s Tau}
\label{app_within_tau}

The within-$\tau$ (the rank correlation between different runs within the same transformation) is shown in \autoref{tab:within_tau}. 

\begin{table}[t]
\small
\centering
\setlength{\tabcolsep}{4pt}
\begin{tabular}{lcc}
\toprule
\textbf{Transformation} & \textcolor{englishblue}{\textbf{English}} & \textcolor{multiorange}{\textbf{Multilingual}} \\
\midrule
\textbf{Lexical/Stylistic} & \textbf{0.945}~\tiny{$\pm$0.090} & \textbf{0.937}~\tiny{$\pm$0.057} \\
\quad Paraphrasing & 0.951~\tiny{$\pm$0.057} & 0.949~\tiny{$\pm$0.065} \\
\quad Backtranslation & 0.890~\tiny{$\pm$0.131} & 0.915~\tiny{$\pm$0.076} \\
\quad Style Change & 0.938~\tiny{$\pm$0.073} & 0.957~\tiny{$\pm$0.064} \\
\midrule
\textbf{Length} & \textbf{0.959}~\tiny{$\pm$0.053} & \textbf{0.943}~\tiny{$\pm$0.056} \\
\quad Expansion & 0.928~\tiny{$\pm$0.062} & 0.922~\tiny{$\pm$0.076} \\
\quad Summarisation & 0.953~\tiny{$\pm$0.075} & 0.941~\tiny{$\pm$0.072} \\
\quad Summ.\ Expansion & 0.886~\tiny{$\pm$0.113} & 0.928~\tiny{$\pm$0.061} \\
\midrule
\textbf{Language} & \textbf{0.892}~\tiny{$\pm$0.072} & \textbf{0.909}~\tiny{$\pm$0.073} \\
\quad Translation & 0.782~\tiny{$\pm$0.130} & 0.806~\tiny{$\pm$0.140} \\
\quad Cross-Translation & 0.938~\tiny{$\pm$0.079} & 0.895~\tiny{$\pm$0.081} \\
\midrule
\textbf{Overall} & \textbf{0.907}~\tiny{$\pm$0.066} & \textbf{0.941}~\tiny{$\pm$0.055} \\
\bottomrule
\end{tabular}
\caption{
Within-transformation Kendall's $\tau$ rank correlation between HTEB model rankings across transformation runs, per robustness axis and transformation. Each cell shows the mean $\pm$ sample standard deviation across per-dataset taus after averaging the pairwise run correlations. Axis and overall rows first average scores over their constituent transformations within each run.
}
\label{tab:within_tau}
\end{table}

\begin{table*}[t]
\centering
\small
\setlength{\tabcolsep}{4pt}
\begin{tabular}{lrcccc}
\toprule
\textbf{Model} & \textbf{Original} & \multicolumn{4}{c}{\textbf{HTEB}} \\
\cmidrule(lr){3-6}
 & & \textbf{Lex./Styl.} & \textbf{Length} & \textbf{Language} & \textbf{Total} \\
\midrule
All-MiniLM-L12-v2 & 66.36 & 62.64 (-3.72) & 61.86 (-4.50) & 37.14 (-29.22) & 53.88 (-12.48) \\
All-MPNet-Base-v2 & 67.09 & 62.84 (-4.25) & 62.57 (-4.52) & 36.34 (-30.75) & 53.92 (-13.17) \\
MXBAI-Emb-Large-v1 & 70.75 & 66.34 (-4.41) & 65.32 (-5.43) & 40.72 (-30.03) & 57.46 (-13.29) \\
Stella-EN-400M-v5 & 71.04 & 66.67 (-4.37) & 65.23 (-5.81) & 45.46 (-25.58) & 59.12 (-11.92) \\
Jasper-600M & 70.78 & 67.10 (-3.67) & 65.14 (-5.63) & 56.69 (-14.08) & 62.98 (-7.80) \\
Jina-Emb-v5-Small & 71.50 & 67.64 (-3.85) & 65.75 (-5.75) & 66.12 (-5.38) & 66.50 (-5.00) \\
F2LLM-v2-4B & 73.00 & 68.42 (-4.58) & 66.93 (-6.07) & 67.76 (-5.24) & 67.70 (-5.30) \\
E5-Mistral-7B-it & 67.28 & 63.32 (-3.96) & 62.71 (-4.57) & 56.47 (-10.82) & 60.83 (-6.45) \\
Llama-Nemotron-8B & 63.17 & 59.91 (-3.26) & 62.49 (-0.68) & 54.16 (-9.01) & 58.85 (-4.32) \\
Qwen3-Emb-8B & 72.77 & 68.12 (-4.65) & 66.62 (-6.14) & 67.09 (-5.68) & 67.28 (-5.49) \\
NV-Embed-v2 & 73.34 & 69.26 (-4.08) & 66.99 (-6.35) & 61.84 (-11.50) & 66.03 (-7.31) \\
\midrule
\textbf{Average} & \textbf{69.73} & \textbf{65.66 (-4.07)} & \textbf{64.69 (-5.04)} & \textbf{53.62 (-16.12)} & \textbf{61.32 (-8.41)} \\
\bottomrule
\end{tabular}
\caption{
Exact values used in the robustness-axis dumbbell plot for English data; scores are percentages. HTEB is the mean of the three per-axis scores. Parenthesised values are differences vs.\ Original.
}
\label{tab:robustness-axis-dumbbell-joint-horizontal-english}
\end{table*}

\begin{table*}[t]
\centering
\small
\setlength{\tabcolsep}{4pt}
\begin{tabular}{lrcccc}
\toprule
\textbf{Model} & \textbf{Original} & \multicolumn{4}{c}{\textbf{HTEB}} \\
\cmidrule(lr){3-6}
 & & \textbf{Lex./Styl.} & \textbf{Length} & \textbf{Language} & \textbf{Total} \\
\midrule
Harrier-270M & 58.87 & 55.59 (-3.27) & 54.64 (-4.23) & 55.47 (-3.39) & 55.23 (-3.63) \\
Paraphr.-ML-MPNet & 55.52 & 54.35 (-1.17) & 53.40 (-2.12) & 58.75 (+3.23) & 55.50 (-0.02) \\
GTE-ML-Base & 62.40 & 60.11 (-2.29) & 59.46 (-2.95) & 61.43 (-0.97) & 60.33 (-2.07) \\
ML-E5-Large-it & 61.65 & 58.63 (-3.02) & 57.88 (-3.78) & 57.61 (-4.04) & 58.04 (-3.61) \\
BGE-M3 & 62.99 & 60.32 (-2.66) & 58.95 (-4.04) & 60.62 (-2.36) & 59.97 (-3.02) \\
Jina-Emb-v5-Small & 63.91 & 61.54 (-2.37) & 60.86 (-3.04) & 64.74 (+0.83) & 62.38 (-1.53) \\
F2LLM-v2-4B & 67.90 & 65.09 (-2.81) & 63.74 (-4.16) & 67.91 (+0.01) & 65.58 (-2.32) \\
Llama-Nemotron-8B & 56.37 & 54.86 (-1.51) & 57.82 (+1.45) & 56.63 (+0.27) & 56.44 (+0.07) \\
Qwen3-Emb-8B & 64.43 & 61.79 (-2.64) & 61.02 (-3.41) & 63.71 (-0.72) & 62.17 (-2.26) \\
\midrule
\textbf{Average} & \textbf{61.56} & \textbf{59.14 (-2.42)} & \textbf{58.64 (-2.92)} & \textbf{60.76 (-0.79)} & \textbf{59.52 (-2.04)} \\
\bottomrule
\end{tabular}
\caption{
Exact values used in the robustness-axis dumbbell plot for Multilingual data; scores are percentages. HTEB is the mean of the three per-axis scores. Parenthesised values are differences vs.\ Original.
}
\label{tab:robustness-axis-dumbbell-joint-horizontal-multilingual}
\end{table*}

\section{Results on the German Translation of STS-B}
\label{sec:sts_deu}

\autoref{tab:stsb_german_per_model_grouped} shows the results for STS-B (original English version) and the translated German version for all models used in the English evaluation setting (i.e. the same models as on the left in \autoref{fig:english_multilingual_total_results}). The table groups these models as English-only or multilingual since English-only models are expected to behave differently than multilingual models on the generated STS-B-German. Translations were generated using Gemma-3-27B-int4-AWQ and selectively validated by a native German speaker.

\begin{table*}[h]
\centering
\small
\begin{tabular}{lrrrrrr}
\toprule
\textbf{Model} & \multicolumn{3}{c}{\textbf{STSB English}} & \multicolumn{3}{c}{\textbf{STSB German}} \\
\cmidrule(lr){2-4} \cmidrule(lr){5-7}
 & \textbf{Original} & \textbf{HTEB} & \textbf{$\Delta$} & \textbf{Original} & \textbf{HTEB} & \textbf{$\Delta$} \\
\midrule
\multicolumn{7}{l}{\textbf{English-only}} \\
\quad All-MiniLM-L12-v2 & 83.1 & 57.8 & $-$25.3 & 64.1 & 48.0 & $-$16.1 \\
\quad All-MPNet-Base-v2 & 83.4 & 57.9 & $-$25.5 & 62.5 & 46.5 & $-$16.0 \\
\quad MXBAI-Emb-Large-v1 & 89.3 & 64.0 & $-$25.3 & 64.8 & 51.4 & $-$13.4 \\
\quad Stella-EN-400M-v5 & 87.5 & 64.4 & $-$23.2 & 68.5 & 55.6 & $-$12.9 \\
\quad Jasper-600M & 87.2 & 72.8 & $-$14.3 & 78.7 & 66.3 & $-$12.4 \\
\quad NV-Embed-v2 & 87.2 & 76.1 & $-$11.1 & 83.9 & 72.6 & $-$11.3 \\
\textit{Subtotal (English-only)} & \textit{86.3} & \textit{65.5} & \textit{$-$20.8} & \textit{70.4} & \textit{56.7} & \textit{$-$13.7} \\
\midrule
\multicolumn{7}{l}{\textbf{Multilingual}} \\
\quad Jina-Emb-v5-Small & 94.8 & 83.6 & $-$11.2 & 90.2 & 82.7 & $-$7.6 \\
\quad F2LLM-v2-4B & 87.3 & 78.2 & $-$9.1 & 83.8 & 77.2 & $-$6.6 \\
\quad E5-Mistral-7B-it & 84.6 & 76.8 & $-$7.8 & 83.6 & 75.1 & $-$8.5 \\
\quad Llama-Nemotron-8B & 72.5 & 71.3 & $-$1.3 & 70.7 & 69.9 & $-$0.9 \\
\quad Qwen3-Emb-8B & 88.5 & 79.2 & $-$9.3 & 90.3 & 83.3 & $-$7.0 \\
\textit{Subtotal (Multilingual)} & \textit{85.5} & \textit{77.8} & \textit{$-$7.7} & \textit{83.7} & \textit{77.6} & \textit{$-$6.1} \\
\midrule
\textbf{Total (avg)} & \textbf{85.9} & \textbf{71.1} & \textbf{$-$14.9} & \textbf{76.5} & \textbf{66.2} & \textbf{$-$10.2} \\
\bottomrule
\end{tabular}
\caption{Per-model Original and HTEB scores on STSBenchmark (EN) vs STSBenchmark-DEU, grouped by scope (English-only vs multilingual) with per-group subtotals and an overall total. HTEB is axis-balanced. $\Delta$ = HTEB $-$ Original per language.}
\label{tab:stsb_german_per_model_grouped}
\end{table*}

\begin{table*}[t]
\centering
\small
\begin{tabular*}{\textwidth}{@{\extracolsep{\fill}}lrrrrrr@{}}
\toprule
& \multicolumn{3}{c}{\textcolor{englishblue}{\textbf{English datasets}}} & \multicolumn{3}{c}{\textcolor{multiorange}{\textbf{Multilingual datasets}}} \\
\cmidrule(lr){2-4} \cmidrule(lr){5-7}
\textbf{Model} & \textbf{Original} & \textbf{HTEB} & \textbf{Rel.\ drop (\%)} & \textbf{Original} & \textbf{HTEB} & \textbf{Rel.\ drop (\%)} \\
\midrule
Jina-Emb-v5-Small & 71.50 & 66.50 & 6.99 & 63.91 & 62.38 & 2.39 \\
F2LLM-v2-4B & 73.00 & 67.70 & 7.25 & 67.90 & 65.58 & 3.41 \\
Llama-Nemotron-8B & 63.17 & 58.85 & 6.83 & 56.37 & 56.44 & -0.12 \\
Qwen3-Emb-8B & 72.77 & 67.28 & 7.55 & 64.43 & 62.17 & 3.51 \\
\midrule
\textbf{Average across four models} & \textbf{70.11} & \textbf{65.08} & \textbf{7.16} & \textbf{63.15} & \textbf{61.64} & \textbf{2.30} \\
\bottomrule
\end{tabular*}
\caption{Original and HTEB scores for the four multilingual models evaluated on both English and multilingual datasets. Relative drops are calculated per model from unrounded scores. The final row reports their mean.}
\label{tab:shared_model_relative_drops}
\end{table*}

\section{HTEB Performance differences for English vs. Multilingual Evaluation}
\label{app_eng_vs_multilingual}

\begin{figure}
    \centering
    \includegraphics[width=1\linewidth]{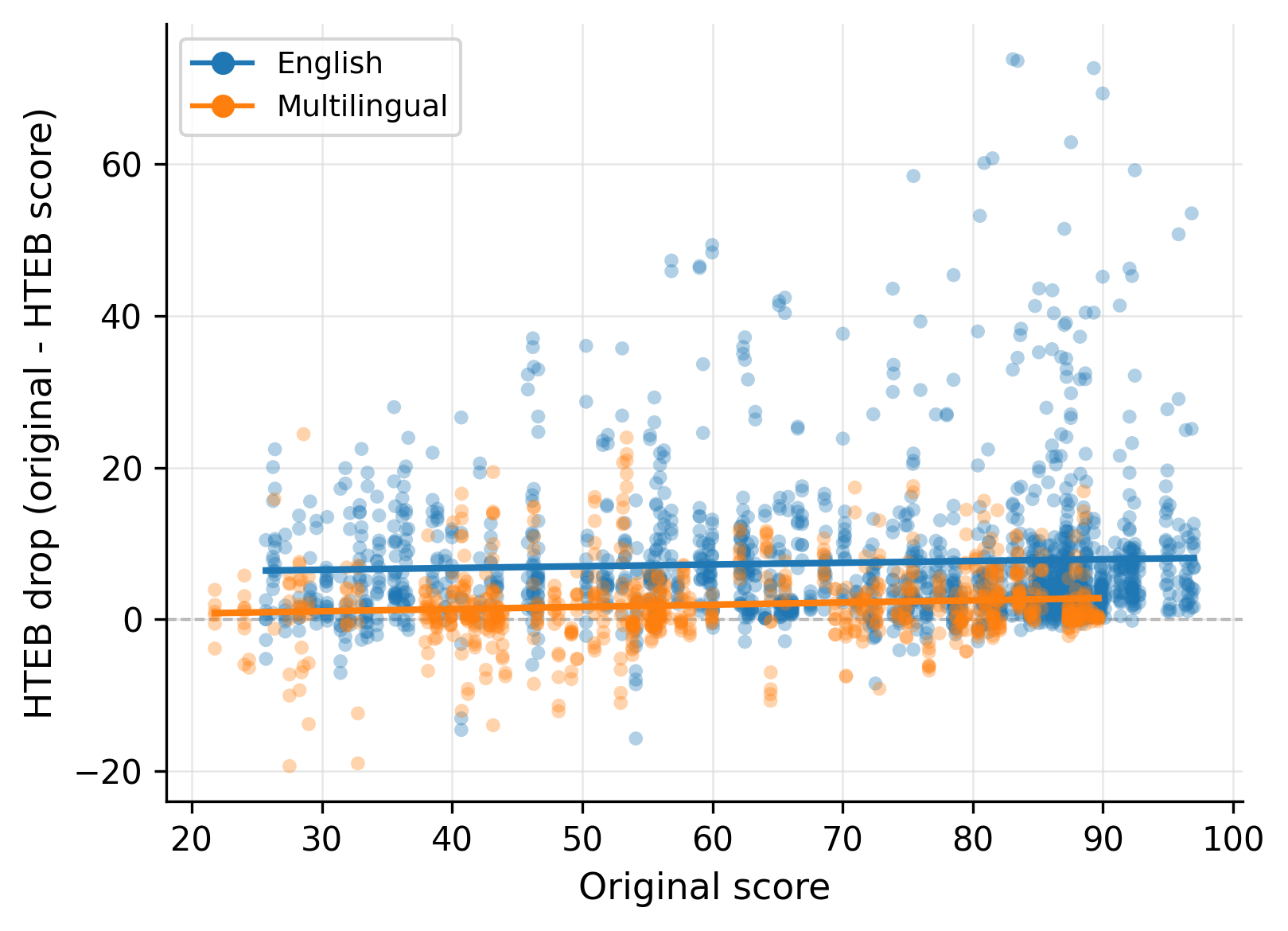}
    \caption{HTEB performance drop vs. Original score for English and multilingual results. Each point is one pair of dataset–model–transformation scores. The x-axis shows the Original score, and the y-axis shows HTEB performance drop, defined as Original - transformed score. Horizontal bars show the mean drop for each setting.}
    \label{fig:hteb_drop_vs_original_score}
\end{figure}

\autoref{fig:hteb_drop_vs_original_score} plots the performance differences on HTEB vs. original data for each dataset-model-transformation combination for English and multilingual evaluations.

\section{Within-Model Axis Ranking by Disruption}
\label{app_axis_ranking}
\begin{figure*}
    \centering
    \includegraphics[width=1\linewidth]{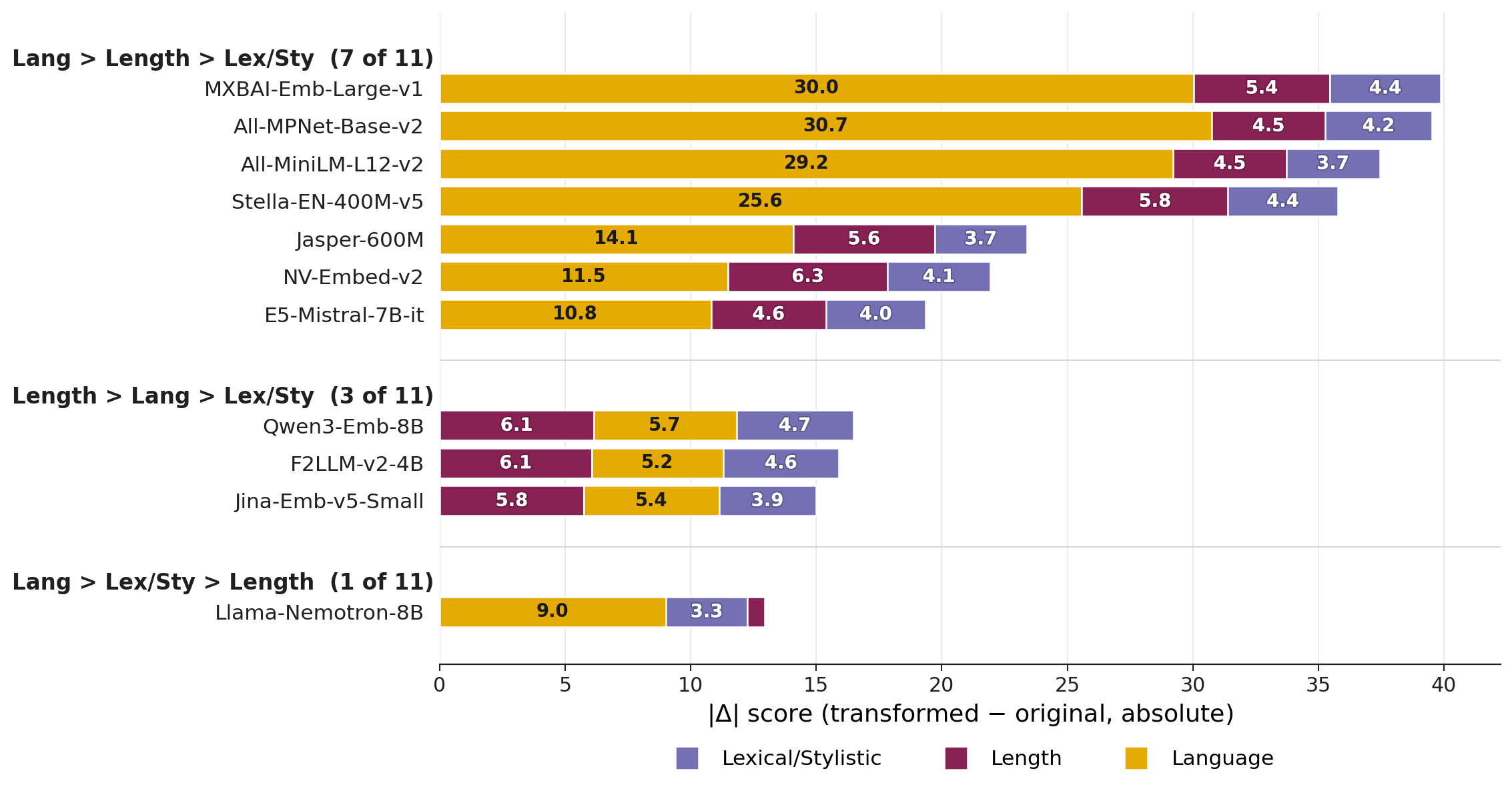}
    \caption{\textcolor{englishblue}{\textbf{English}} benchmark ($n=11$ models). Per-model $|\Delta|$ score by robustness axis, stacked left-to-right in order of within-model disruption and grouped by axis ordering.}
    \label{fig:app_axis_ranking_eng}
\end{figure*}

\begin{figure*}
    \centering
    \includegraphics[width=1\linewidth]{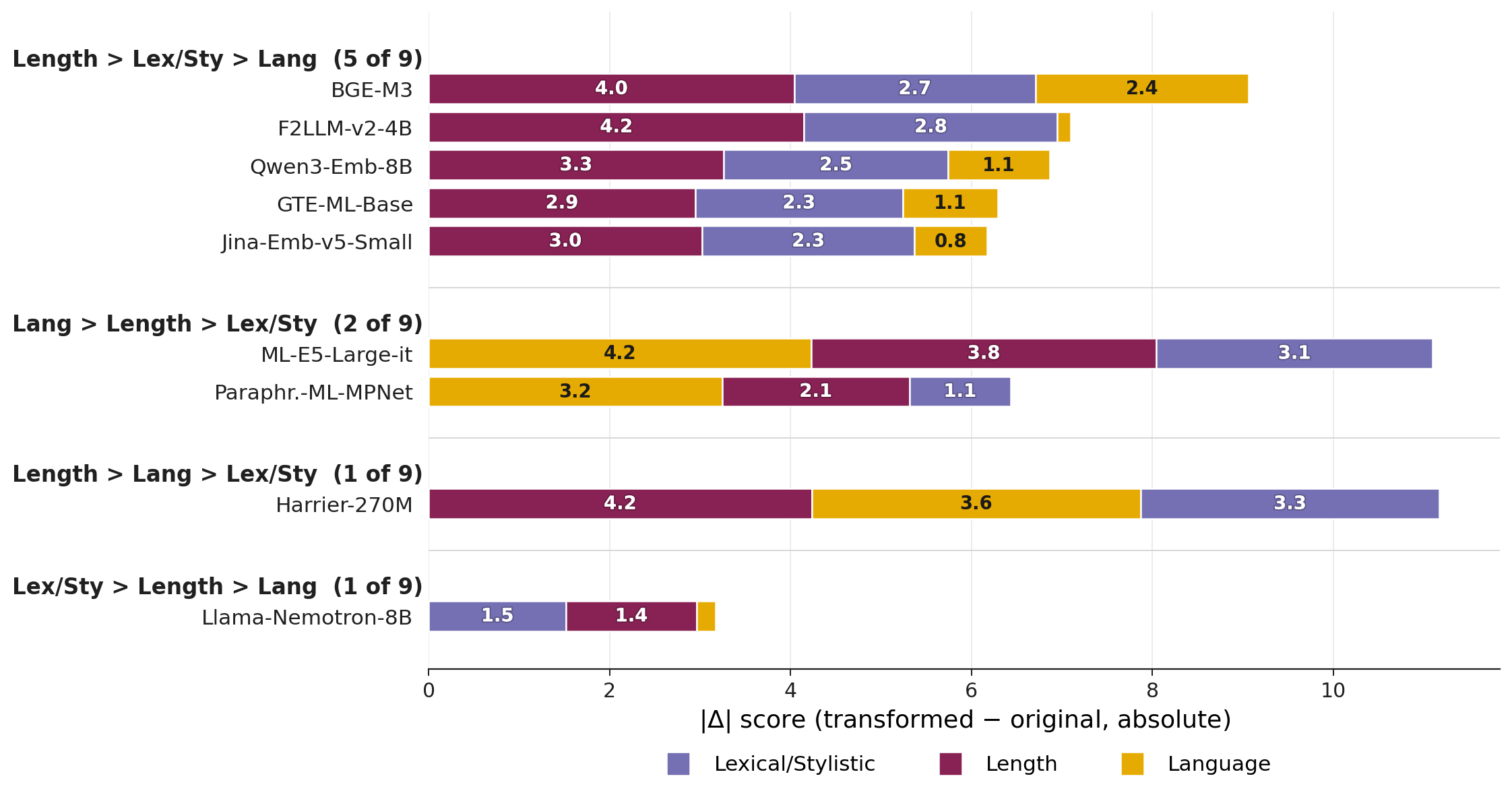}
    \caption{\textcolor{multiorange}{\textbf{Multilingual}} benchmark ($n=9$ models). Per-model $|\Delta|$ score by robustness axis, stacked left-to-right in order of within-model disruption and grouped by axis ordering.}
    \label{fig:app_axis_ranking_multilingual}
\end{figure*}

\autoref{fig:app_axis_ranking_eng} and \autoref{fig:app_axis_ranking_multilingual} provide a visualisation of the axis profiles per embedding model.

\section{HTEB Results on Dataset-Level}
\label{app_dataset_level_results}

\autoref{tab:english_results_per_dataset_1} to \autoref{tab:multilingual_results_per_dataset_2} show the results per dataset, including the standard deviation over runs. Exact total scores can differ from \autoref{fig:english_multilingual_total_results} due to rounding.  

\begin{table*}[ht]
\small
\centering
\begin{tabularx}{\textwidth}{
  >{\centering\arraybackslash}p{0.03cm}
  >{\raggedright\arraybackslash}p{1.9cm}
  >{\centering\arraybackslash}X
  @{\hspace{0pt}}
  >{\centering\arraybackslash}X
  @{\hspace{8pt}}
  >{\centering\arraybackslash}X
  @{\hspace{0pt}}
  >{\centering\arraybackslash}X
  @{\hspace{8pt}}
  >{\centering\arraybackslash}X
  @{\hspace{0pt}}
  >{\centering\arraybackslash}X
  @{\hspace{8pt}}
  >{\centering\arraybackslash}X
  @{\hspace{0pt}}
  >{\centering\arraybackslash}X
}
\toprule
 &  & \multicolumn{2}{c}{\textbf{All-MiniLM-}} & \multicolumn{2}{c}{\textbf{All-MPNet-}} & \multicolumn{2}{c}{\textbf{MXBAI-Em-}} & \multicolumn{2}{c}{\textbf{Stella-EN-}} \\
 &  & \multicolumn{2}{c}{\textbf{L12-v2}} & \multicolumn{2}{c}{\textbf{Base-v2}} & \multicolumn{2}{c}{\textbf{bed-Large-v1}} & \multicolumn{2}{c}{\textbf{400M-v5}} \\
\cmidrule(l{2pt}r{2pt}){3-4}\cmidrule(l{2pt}r{2pt}){5-6}\cmidrule(l{2pt}r{2pt}){7-8}\cmidrule(l{2pt}r{2pt}){9-10}
 & \textbf{Dataset} & \textbf{Orig.} & \textbf{HTEB} & \textbf{Orig.} & \textbf{HTEB} & \textbf{Orig.} & \textbf{HTEB} & \textbf{Orig.} & \textbf{HTEB} \\
\midrule
\multirow{3}{*}{\rotatebox[origin=c]{90}{\textbf{CLS}}} & AmazonCounterf. & 87.46 & \red{85.10}~{\small $\pm$0.07} & 85.37 & \red{84.33}~{\small $\pm$0.16} & 89.85 & \red{85.73}~{\small $\pm$0.26} & 89.85 & \red{85.42}~{\small $\pm$0.21} \\
 & Banking77 & 91.29 & \red{76.83}~{\small $\pm$3.20} & 92.23 & \red{76.52}~{\small $\pm$3.12} & 92.04 & \red{75.68}~{\small $\pm$3.03} & 87.03 & \red{67.32}~{\small $\pm$3.78} \\
 & MassiveIntent & 85.07 & \red{73.47}~{\small $\pm$2.42} & 86.08 & \red{73.55}~{\small $\pm$2.51} & 86.79 & \red{74.02}~{\small $\pm$2.72} & 88.23 & \red{73.17}~{\small $\pm$3.91} \\
\midrule
\multirow{3}{*}{\rotatebox[origin=c]{90}{\textbf{CLU}}} & BiorxivS2S & 31.80 & \red{25.89}~{\small $\pm$1.63} & 33.51 & \red{26.62}~{\small $\pm$3.49} & 36.30 & \red{28.35}~{\small $\pm$3.93} & 42.14 & \red{33.69}~{\small $\pm$2.37} \\
 & MedrxivS2S & 28.25 & \red{23.53}~{\small $\pm$1.45} & 29.58 & \red{24.68}~{\small $\pm$1.31} & 27.19 & \red{23.56}~{\small $\pm$1.60} & 32.16 & \red{27.08}~{\small $\pm$2.17} \\
 & 20Newsgroups & 46.62 & \red{36.21}~{\small $\pm$1.51} & 50.30 & \red{37.55}~{\small $\pm$1.54} & 53.06 & \red{40.52}~{\small $\pm$2.39} & 55.54 & \red{43.77}~{\small $\pm$2.58} \\
\midrule
\multirow{3}{*}{\rotatebox[origin=c]{90}{\textbf{PC}}} & SprintDupQues. & 92.44 & \red{74.18}~{\small $\pm$0.49} & 89.97 & \red{67.94}~{\small $\pm$0.30} & \underline{96.81} & \red{80.44}~{\small $\pm$0.56} & 95.81 & \red{77.93}~{\small $\pm$0.47} \\
 & TwitterSemEval & 70.02 & \red{56.35}~{\small $\pm$0.04} & 73.85 & \red{57.06}~{\small $\pm$0.06} & \textbf{78.52} & \red{59.56}~{\small $\pm$0.07} & 75.98 & \red{59.69}~{\small $\pm$0.13} \\
 & TwitterURL & 84.78 & \red{70.82}~{\small $\pm$0.08} & 85.09 & \red{70.06}~{\small $\pm$0.05} & 86.22 & \red{73.35}~{\small $\pm$0.09} & \underline{87.17} & \red{74.59}~{\small $\pm$0.05} \\
\midrule
\multirow{3}{*}{\rotatebox[origin=c]{90}{\textbf{RR}}} & AskUbuntu & 64.04 & \red{59.16}~{\small $\pm$1.00} & 65.82 & \red{59.84}~{\small $\pm$1.26} & 65.20 & \red{58.73}~{\small $\pm$1.21} & 63.03 & \red{58.84}~{\small $\pm$1.80} \\
 & SciDocsRR & 87.20 & \red{75.11}~{\small $\pm$3.78} & \underline{88.64} & \red{76.83}~{\small $\pm$3.83} & 87.53 & \red{77.64}~{\small $\pm$4.20} & 86.39 & \red{78.74}~{\small $\pm$4.42} \\
 & StackOverflow & 51.58 & \red{40.44}~{\small $\pm$1.26} & 51.96 & \red{40.91}~{\small $\pm$1.40} & 55.20 & \red{43.61}~{\small $\pm$1.62} & 51.90 & \red{44.18}~{\small $\pm$2.14} \\
\midrule
\multirow{3}{*}{\rotatebox[origin=c]{90}{\textbf{RT}}} & ArguAna & 46.21 & \red{31.76}~{\small $\pm$1.61} & 45.83 & \red{32.58}~{\small $\pm$2.66} & 65.11 & \red{45.32}~{\small $\pm$3.92} & 63.29 & \red{50.83}~{\small $\pm$4.77} \\
 & CQADupstack & 56.85 & \red{34.00}~{\small $\pm$1.24} & 59.98 & \red{36.55}~{\small $\pm$1.50} & 59.01 & \red{35.51}~{\small $\pm$1.62} & 62.36 & \red{42.48}~{\small $\pm$3.48} \\
 & SciFact & 62.49 & \red{49.77}~{\small $\pm$3.12} & 65.56 & \red{50.43}~{\small $\pm$2.66} & 73.91 & \red{61.68}~{\small $\pm$4.23} & 78.00 & \red{66.45}~{\small $\pm$4.95} \\
\midrule
\rotatebox[origin=c]{90}{\textbf{STR}} & SemRel2024 & 81.51 & \red{66.36}~{\small $\pm$1.44} & 80.87 & \red{65.49}~{\small $\pm$1.31} & 80.54 & \red{66.96}~{\small $\pm$2.56} & \textbf{83.64} & \red{72.66}~{\small $\pm$2.03} \\
\midrule
\multirow{2}{*}{\rotatebox[origin=c]{90}{\textbf{STS}}} & BIOSSES & 83.70 & \red{72.76}~{\small $\pm$1.88} & 80.39 & \red{68.86}~{\small $\pm$1.87} & 86.11 & \red{73.48}~{\small $\pm$1.47} & 81.17 & \red{74.97}~{\small $\pm$0.80} \\
 & STSBenchmark & 83.07 & \red{57.76}~{\small $\pm$0.56} & 83.43 & \red{57.94}~{\small $\pm$0.14} & \underline{89.29} & \red{64.02}~{\small $\pm$2.38} & 87.54 & \red{64.36}~{\small $\pm$1.18} \\
\midrule
\rotatebox[origin=c]{90}{\textbf{Sum}} & SummEval & 26.40 & \red{14.24}~{\small $\pm$1.06} & 26.25 & \red{16.72}~{\small $\pm$0.82} & 35.55 & \red{23.60}~{\small $\pm$0.93} & \underline{38.50} & \red{27.05}~{\small $\pm$1.62} \\
\midrule
 & \textbf{Average} & 66.36 & \red{53.88}~{\small $\pm$1.47} & 67.09 & \red{53.92}~{\small $\pm$1.58} & 70.75 & \red{57.46}~{\small $\pm$2.04} & 71.04 & \red{59.12}~{\small $\pm$2.26} \\
\bottomrule
\end{tabularx}
\caption{
\textcolor{englishblue}{\textbf{English (part 1 of 3)}} embedding model performance on original data and axis-balanced HTEB scores based on 3 runs with $\pm$ denoting the standard deviation (in \%). \textbf{Bold}: Row-wise best score across all table parts. \underline{Underline}: Row-wise second-best score across all table parts. \green{Green}: Performance increase. \red{Red}: Performance decrease. Task abbreviations: CLS = Classification, CLU = Clustering, PC = Pair Classification, RR = Reranking, RT = Retrieval, STR = Semantic Textual Relatedness, STS = Semantic Textual Similarity, Sum = Summarisation.
}
\label{tab:english_results_per_dataset_1}
\end{table*}

\begin{table*}[ht]
\small
\centering
\begin{tabularx}{\textwidth}{
  >{\centering\arraybackslash}p{0.03cm}
  >{\raggedright\arraybackslash}p{1.9cm}
  >{\centering\arraybackslash}X
  @{\hspace{0pt}}
  >{\centering\arraybackslash}X
  @{\hspace{8pt}}
  >{\centering\arraybackslash}X
  @{\hspace{0pt}}
  >{\centering\arraybackslash}X
  @{\hspace{8pt}}
  >{\centering\arraybackslash}X
  @{\hspace{0pt}}
  >{\centering\arraybackslash}X
  @{\hspace{8pt}}
  >{\centering\arraybackslash}X
  @{\hspace{0pt}}
  >{\centering\arraybackslash}X
}
\toprule
 &  & \multicolumn{2}{c}{\textbf{Jasper-Token-}} & \multicolumn{2}{c}{\textbf{Jina-Emb-v5-}} & \multicolumn{2}{c}{\textbf{F2LLM-}} & \multicolumn{2}{c}{\textbf{E5-Mistral-}} \\
 &  & \multicolumn{2}{c}{\textbf{Compr.-600M}} & \multicolumn{2}{c}{\textbf{Text-Small}} & \multicolumn{2}{c}{\textbf{v2-4B}} & \multicolumn{2}{c}{\textbf{7B-it}} \\
\cmidrule(l{2pt}r{2pt}){3-4}\cmidrule(l{2pt}r{2pt}){5-6}\cmidrule(l{2pt}r{2pt}){7-8}\cmidrule(l{2pt}r{2pt}){9-10}
 & \textbf{Dataset} & \textbf{Orig.} & \textbf{HTEB} & \textbf{Orig.} & \textbf{HTEB} & \textbf{Orig.} & \textbf{HTEB} & \textbf{Orig.} & \textbf{HTEB} \\
\midrule
\multirow{3}{*}{\rotatebox[origin=c]{90}{\textbf{CLS}}} & AmazonCounterf. & 90.15 & \red{86.46}~{\small $\pm$0.34} & \textbf{92.24} & \red{\textbf{88.31}}~{\small $\pm$0.07} & 89.85 & \red{86.99}~{\small $\pm$0.08} & 88.36 & \red{86.90}~{\small $\pm$0.32} \\
 & Banking77 & \underline{92.39} & \red{83.94}~{\small $\pm$0.69} & 92.04 & \red{79.56}~{\small $\pm$1.28} & 92.30 & \red{\underline{87.55}}~{\small $\pm$0.21} & 90.93 & \red{84.62}~{\small $\pm$0.53} \\
 & MassiveIntent & 86.92 & \red{82.03}~{\small $\pm$0.29} & \textbf{88.74} & \red{82.25}~{\small $\pm$0.08} & 88.26 & \red{\underline{84.73}}~{\small $\pm$0.25} & 86.42 & \red{83.08}~{\small $\pm$0.47} \\
\midrule
\multirow{3}{*}{\rotatebox[origin=c]{90}{\textbf{CLU}}} & BiorxivS2S & 36.64 & \red{29.22}~{\small $\pm$2.51} & 39.47 & \red{35.49}~{\small $\pm$0.45} & \textbf{55.98} & \red{\textbf{41.28}}~{\small $\pm$1.12} & 33.05 & \red{25.30}~{\small $\pm$2.22} \\
 & MedrxivS2S & 30.41 & \red{26.58}~{\small $\pm$0.50} & 33.39 & \red{\underline{31.88}}~{\small $\pm$0.26} & \textbf{42.98} & \red{\textbf{35.26}}~{\small $\pm$0.42} & 29.10 & \red{23.27}~{\small $\pm$0.61} \\
 & 20Newsgroups & 59.28 & \red{44.89}~{\small $\pm$1.21} & 54.20 & \red{\underline{50.65}}~{\small $\pm$0.47} & \underline{59.71} & \red{\textbf{52.73}}~{\small $\pm$0.74} & 46.28 & \red{33.87}~{\small $\pm$0.65} \\
\midrule
\multirow{3}{*}{\rotatebox[origin=c]{90}{\textbf{PC}}} & SprintDupQues. & 96.35 & \red{87.44}~{\small $\pm$0.20} & 96.54 & \red{\textbf{92.30}}~{\small $\pm$0.11} & 95.11 & \red{\underline{89.56}}~{\small $\pm$0.20} & 88.67 & \red{74.53}~{\small $\pm$0.21} \\
 & TwitterSemEval & 72.35 & \red{64.21}~{\small $\pm$0.02} & 72.30 & \red{\underline{68.76}}~{\small $\pm$0.04} & \underline{77.49} & \red{\textbf{70.25}}~{\small $\pm$0.09} & 75.40 & \red{67.43}~{\small $\pm$0.09} \\
 & TwitterURL & 85.79 & \red{78.45}~{\small $\pm$0.06} & 85.99 & \red{81.51}~{\small $\pm$0.07} & 86.98 & \red{\textbf{82.82}}~{\small $\pm$0.03} & 85.16 & \red{80.45}~{\small $\pm$0.08} \\
\midrule
\multirow{3}{*}{\rotatebox[origin=c]{90}{\textbf{RR}}} & AskUbuntu & 62.76 & \red{60.00}~{\small $\pm$0.59} & 66.07 & \red{64.44}~{\small $\pm$0.30} & \underline{66.14} & \red{\underline{64.66}}~{\small $\pm$0.27} & 60.05 & \red{58.56}~{\small $\pm$0.32} \\
 & SciDocsRR & 84.38 & \red{80.40}~{\small $\pm$0.96} & 86.11 & \red{84.13}~{\small $\pm$0.27} & 86.90 & \red{\underline{84.51}}~{\small $\pm$0.34} & 82.10 & \red{78.57}~{\small $\pm$0.81} \\
 & StackOverflow & 50.39 & \red{43.76}~{\small $\pm$0.86} & 55.23 & \red{50.38}~{\small $\pm$0.49} & \underline{57.34} & \red{\underline{53.41}}~{\small $\pm$0.51} & 46.57 & \red{43.27}~{\small $\pm$0.51} \\
\midrule
\multirow{3}{*}{\rotatebox[origin=c]{90}{\textbf{RT}}} & ArguAna & \underline{68.70} & \red{\underline{60.29}}~{\small $\pm$2.55} & 64.46 & \red{59.33}~{\small $\pm$0.20} & 62.15 & \red{58.85}~{\small $\pm$0.06} & 53.29 & \red{44.55}~{\small $\pm$1.99} \\
 & CQADupstack & 66.55 & \red{50.23}~{\small $\pm$1.84} & 62.16 & \red{52.17}~{\small $\pm$0.81} & 62.42 & \red{\underline{54.01}}~{\small $\pm$0.57} & 56.30 & \red{41.52}~{\small $\pm$1.90} \\
 & SciFact & 75.00 & \red{69.25}~{\small $\pm$1.98} & 76.53 & \red{73.70}~{\small $\pm$0.92} & 71.05 & \red{68.11}~{\small $\pm$0.39} & 74.81 & \red{68.48}~{\small $\pm$2.45} \\
\midrule
\rotatebox[origin=c]{90}{\textbf{STR}} & SemRel2024 & 74.36 & \red{71.88}~{\small $\pm$0.09} & 78.69 & \red{77.46}~{\small $\pm$0.09} & 82.20 & \red{\textbf{79.81}}~{\small $\pm$0.01} & 78.47 & \red{76.85}~{\small $\pm$0.10} \\
\midrule
\multirow{2}{*}{\rotatebox[origin=c]{90}{\textbf{STS}}} & BIOSSES & \textbf{88.71} & \red{80.12}~{\small $\pm$1.02} & 85.07 & \red{81.23}~{\small $\pm$0.18} & \underline{87.14} & \red{\textbf{83.90}}~{\small $\pm$0.28} & 84.61 & \red{78.40}~{\small $\pm$0.74} \\
 & STSBenchmark & 87.15 & \red{72.83}~{\small $\pm$0.70} & \textbf{94.83} & \red{\textbf{83.59}}~{\small $\pm$0.07} & 87.30 & \red{78.16}~{\small $\pm$0.10} & 84.58 & \red{76.82}~{\small $\pm$0.43} \\
\midrule
\rotatebox[origin=c]{90}{\textbf{Sum}} & SummEval & 36.46 & \red{24.63}~{\small $\pm$0.05} & 34.40 & \red{26.38}~{\small $\pm$0.30} & 35.67 & \red{\textbf{29.76}}~{\small $\pm$0.51} & 34.24 & \red{\underline{29.37}}~{\small $\pm$0.49} \\
\midrule
 & \textbf{Average} & 70.78 & \red{62.98}~{\small $\pm$0.87} & 71.50 & \red{66.50}~{\small $\pm$0.34} & \underline{73.00} & \red{\textbf{67.70}}~{\small $\pm$0.32} & 67.28 & \red{60.83}~{\small $\pm$0.79} \\
\bottomrule
\end{tabularx}
\caption{
\textcolor{englishblue}{\textbf{English (part 2 of 3)}} embedding model performance on original data and axis-balanced HTEB scores based on 3 runs with $\pm$ denoting the standard deviation (in \%). \textbf{Bold}: Row-wise best score across all table parts. \underline{Underline}: Row-wise second-best score across all table parts. \green{Green}: Performance increase. \red{Red}: Performance decrease. Task abbreviations: CLS = Classification, CLU = Clustering, PC = Pair Classification, RR = Reranking, RT = Retrieval, STR = Semantic Textual Relatedness, STS = Semantic Textual Similarity, Sum = Summarisation.
}
\label{tab:english_results_per_dataset_2}
\end{table*}

\begin{table*}[ht]
\small
\centering
\begin{tabularx}{\textwidth}{
  >{\centering\arraybackslash}p{0.03cm}
  >{\raggedright\arraybackslash}p{1.9cm}
  >{\centering\arraybackslash}X
  @{\hspace{0pt}}
  >{\centering\arraybackslash}X
  @{\hspace{8pt}}
  >{\centering\arraybackslash}X
  @{\hspace{0pt}}
  >{\centering\arraybackslash}X
  @{\hspace{8pt}}
  >{\centering\arraybackslash}X
  @{\hspace{0pt}}
  >{\centering\arraybackslash}X
}
\toprule
 &  & \multicolumn{2}{c}{\textbf{Llama-Emb-}} & \multicolumn{2}{c}{\textbf{Qwen3-Em-}} & \multicolumn{2}{c}{\textbf{NV-Embed-}} \\
 &  & \multicolumn{2}{c}{\textbf{Nemotron-8B}} & \multicolumn{2}{c}{\textbf{bedding-8B}} & \multicolumn{2}{c}{\textbf{v2}} \\
\cmidrule(l{2pt}r{2pt}){3-4}\cmidrule(l{2pt}r{2pt}){5-6}\cmidrule(l{2pt}r{2pt}){7-8}
 & \textbf{Dataset} & \textbf{Orig.} & \textbf{HTEB} & \textbf{Orig.} & \textbf{HTEB} & \textbf{Orig.} & \textbf{HTEB} \\
\midrule
\multirow{3}{*}{\rotatebox[origin=c]{90}{\textbf{CLS}}} & AmazonCounterf. & \underline{91.34} & \red{87.72}~{\small $\pm$0.05} & 88.96 & \red{87.53}~{\small $\pm$0.07} & \textbf{92.24} & \red{\underline{88.30}}~{\small $\pm$0.36} \\
 & Banking77 & 87.42 & \red{82.80}~{\small $\pm$0.42} & \textbf{92.78} & \red{\textbf{87.82}}~{\small $\pm$0.18} & 91.61 & \red{84.10}~{\small $\pm$0.44} \\
 & MassiveIntent & 84.40 & \red{82.68}~{\small $\pm$0.14} & \underline{88.33} & \red{\textbf{84.96}}~{\small $\pm$0.13} & 87.36 & \red{82.13}~{\small $\pm$0.38} \\
\midrule
\multirow{3}{*}{\rotatebox[origin=c]{90}{\textbf{CLU}}} & BiorxivS2S & 31.43 & \red{29.30}~{\small $\pm$0.52} & 39.98 & \red{31.81}~{\small $\pm$2.25} & \underline{46.63} & \red{\underline{36.71}}~{\small $\pm$1.83} \\
 & MedrxivS2S & 25.71 & \red{24.24}~{\small $\pm$0.31} & 30.34 & \red{25.83}~{\small $\pm$0.30} & \underline{35.64} & \red{30.41}~{\small $\pm$0.78} \\
 & 20Newsgroups & 40.71 & \red{37.10}~{\small $\pm$0.39} & 55.68 & \red{43.87}~{\small $\pm$1.12} & \textbf{62.72} & \red{50.25}~{\small $\pm$1.36} \\
\midrule
\multirow{3}{*}{\rotatebox[origin=c]{90}{\textbf{PC}}} & SprintDupQues. & 75.44 & \red{57.12}~{\small $\pm$0.16} & \textbf{96.97} & \red{89.03}~{\small $\pm$0.11} & 94.94 & \red{83.07}~{\small $\pm$0.23} \\
 & TwitterSemEval & 54.12 & \green{56.32}~{\small $\pm$0.08} & 72.05 & \red{68.66}~{\small $\pm$0.08} & 77.16 & \red{67.33}~{\small $\pm$0.08} \\
 & TwitterURL & 85.65 & \red{77.19}~{\small $\pm$0.06} & 86.29 & \red{\underline{81.64}}~{\small $\pm$0.01} & \textbf{88.26} & \red{81.04}~{\small $\pm$0.07} \\
\midrule
\multirow{3}{*}{\rotatebox[origin=c]{90}{\textbf{RR}}} & AskUbuntu & 55.62 & \red{54.78}~{\small $\pm$0.21} & \textbf{67.54} & \red{\textbf{65.21}}~{\small $\pm$0.23} & 65.23 & \red{64.07}~{\small $\pm$0.44} \\
 & SciDocsRR & 80.77 & \red{77.94}~{\small $\pm$0.50} & \textbf{89.70} & \red{\textbf{87.65}}~{\small $\pm$0.31} & 85.77 & \red{82.51}~{\small $\pm$1.20} \\
 & StackOverflow & 43.46 & \red{39.24}~{\small $\pm$0.28} & \textbf{59.85} & \red{\textbf{54.28}}~{\small $\pm$0.45} & 56.00 & \red{51.79}~{\small $\pm$0.87} \\
\midrule
\multirow{3}{*}{\rotatebox[origin=c]{90}{\textbf{RT}}} & ArguAna & 59.07 & \red{54.66}~{\small $\pm$0.17} & \textbf{75.26} & \red{\textbf{69.39}}~{\small $\pm$2.78} & 68.60 & \red{60.21}~{\small $\pm$2.68} \\
 & CQADupstack & 46.16 & \red{37.75}~{\small $\pm$0.93} & \textbf{70.17} & \red{\textbf{59.29}}~{\small $\pm$0.66} & \underline{66.88} & \red{53.74}~{\small $\pm$2.17} \\
 & SciFact & 75.71 & \red{71.71}~{\small $\pm$0.99} & \underline{78.56} & \red{\underline{75.92}}~{\small $\pm$0.52} & \textbf{81.23} & \red{\textbf{76.02}}~{\small $\pm$2.17} \\
\midrule
\rotatebox[origin=c]{90}{\textbf{STR}} & SemRel2024 & 79.12 & \red{77.16}~{\small $\pm$0.24} & 76.42 & \red{76.10}~{\small $\pm$0.07} & \underline{83.02} & \red{\underline{79.21}}~{\small $\pm$0.14} \\
\midrule
\multirow{2}{*}{\rotatebox[origin=c]{90}{\textbf{STS}}} & BIOSSES & 78.67 & \red{76.21}~{\small $\pm$0.78} & 86.29 & \red{\underline{81.67}}~{\small $\pm$0.48} & 86.76 & \red{79.51}~{\small $\pm$1.16} \\
 & STSBenchmark & 72.52 & \red{71.26}~{\small $\pm$0.03} & 88.50 & \red{\underline{79.22}}~{\small $\pm$0.02} & 87.19 & \red{76.10}~{\small $\pm$0.66} \\
\midrule
\rotatebox[origin=c]{90}{\textbf{Sum}} & SummEval & 32.91 & \red{23.03}~{\small $\pm$0.53} & \textbf{38.87} & \red{28.36}~{\small $\pm$0.44} & 36.19 & \red{28.08}~{\small $\pm$0.47} \\
\midrule
 & \textbf{Average} & 63.17 & \red{58.85}~{\small $\pm$0.36} & 72.77 & \red{\underline{67.28}}~{\small $\pm$0.54} & \textbf{73.34} & \red{66.03}~{\small $\pm$0.92} \\
\bottomrule
\end{tabularx}
\caption{
\textcolor{englishblue}{\textbf{English (part 3 of 3)}} embedding model performance on original data and axis-balanced HTEB scores based on 3 runs with $\pm$ denoting the standard deviation (in \%). \textbf{Bold}: Row-wise best score across all table parts. \underline{Underline}: Row-wise second-best score across all table parts. \green{Green}: Performance increase. \red{Red}: Performance decrease. Task abbreviations: CLS = Classification, CLU = Clustering, PC = Pair Classification, RR = Reranking, RT = Retrieval, STR = Semantic Textual Relatedness, STS = Semantic Textual Similarity, Sum = Summarisation.
}
\label{tab:english_results_per_dataset_3}
\end{table*}

\begin{table*}[ht]
\small
\centering
\begin{tabularx}{\textwidth}{
  >{\centering\arraybackslash}p{0.03cm}
  >{\raggedright\arraybackslash}p{1.9cm}
  >{\centering\arraybackslash}X
  @{\hspace{0pt}}
  >{\centering\arraybackslash}X
  @{\hspace{8pt}}
  >{\centering\arraybackslash}X
  @{\hspace{0pt}}
  >{\centering\arraybackslash}X
  @{\hspace{8pt}}
  >{\centering\arraybackslash}X
  @{\hspace{0pt}}
  >{\centering\arraybackslash}X
  @{\hspace{8pt}}
  >{\centering\arraybackslash}X
  @{\hspace{0pt}}
  >{\centering\arraybackslash}X
  @{\hspace{8pt}}
  >{\centering\arraybackslash}X
  @{\hspace{0pt}}
  >{\centering\arraybackslash}X
}
\toprule
 &  & \multicolumn{2}{c}{\textbf{Harrier-OSS-}} & \multicolumn{2}{c}{\textbf{Para-ML-}} & \multicolumn{2}{c}{\textbf{GTE-Multi-}} & \multicolumn{2}{c}{\textbf{Multilingual-E5-}} & \multicolumn{2}{c}{\textbf{BGE-M3}} \\
 &  & \multicolumn{2}{c}{\textbf{v1-270M}} & \multicolumn{2}{c}{\textbf{MPNet-Base-v2}} & \multicolumn{2}{c}{\textbf{lingual-Base}} & \multicolumn{2}{c}{\textbf{Large-it}} & \multicolumn{2}{c}{} \\
\cmidrule(l{2pt}r{2pt}){3-4}\cmidrule(l{2pt}r{2pt}){5-6}\cmidrule(l{2pt}r{2pt}){7-8}\cmidrule(l{2pt}r{2pt}){9-10}\cmidrule(l{2pt}r{2pt}){11-12}
 & \textbf{Dataset} & \textbf{Orig.} & \textbf{HTEB} & \textbf{Orig.} & \textbf{HTEB} & \textbf{Orig.} & \textbf{HTEB} & \textbf{Orig.} & \textbf{HTEB} & \textbf{Orig.} & \textbf{HTEB} \\
\midrule
\multirow{2}{*}{\rotatebox[origin=c]{90}{\textbf{CLS}}} & IsiZuluNews & 42.87 & \green{43.15}~{\small $\pm$0.49} & 41.22 & \green{43.98}~{\small $\pm$0.23} & \underline{47.87} & \red{45.86}~{\small $\pm$0.16} & 41.62 & \red{40.66}~{\small $\pm$0.38} & 46.54 & \red{\underline{46.02}}~{\small $\pm$0.57} \\
 & SentimentHindi & 80.76 & \green{80.86}~{\small $\pm$1.74} & 82.08 & \green{82.12}~{\small $\pm$0.12} & 80.42 & \green{81.66}~{\small $\pm$0.14} & 81.64 & \red{81.15}~{\small $\pm$0.21} & \underline{84.72} & \red{\underline{83.54}}~{\small $\pm$0.08} \\
\midrule
\multirow{2}{*}{\rotatebox[origin=c]{90}{\textbf{CLU}}} & MasakhaNEWS & 31.65 & \red{23.77}~{\small $\pm$0.28} & 22.44 & \green{24.28}~{\small $\pm$0.32} & 31.91 & \red{29.92}~{\small $\pm$0.20} & 27.80 & \red{23.33}~{\small $\pm$0.27} & 24.05 & \red{23.34}~{\small $\pm$0.19} \\
 & PlscS2S.v2 & 39.88 & \red{35.16}~{\small $\pm$0.57} & 38.83 & \green{39.23}~{\small $\pm$0.09} & 38.64 & \red{37.92}~{\small $\pm$0.27} & 40.10 & \red{35.68}~{\small $\pm$0.61} & 37.96 & \red{35.89}~{\small $\pm$0.43} \\
\midrule
\multirow{2}{*}{\rotatebox[origin=c]{90}{\textbf{PC}}} & RTE3 & 86.50 & \red{86.50}~{\small $\pm$0.18} & \underline{89.14} & \red{\underline{88.68}}~{\small $\pm$0.01} & 88.03 & \red{87.41}~{\small $\pm$0.12} & 87.62 & \red{87.45}~{\small $\pm$0.08} & 88.20 & \red{87.89}~{\small $\pm$0.10} \\
 & IndonLI & 50.00 & \green{52.51}~{\small $\pm$0.15} & \underline{57.60} & \green{\underline{57.74}}~{\small $\pm$0.18} & 55.32 & \green{56.55}~{\small $\pm$0.18} & 53.85 & \green{55.44}~{\small $\pm$0.12} & 55.96 & \green{56.11}~{\small $\pm$0.20} \\
\midrule
\multirow{2}{*}{\rotatebox[origin=c]{90}{\textbf{RR}}} & VoyageMMarco & 60.92 & \red{52.05}~{\small $\pm$0.53} & 41.12 & \red{40.55}~{\small $\pm$0.70} & 57.78 & \red{52.29}~{\small $\pm$0.65} & 62.10 & \red{53.23}~{\small $\pm$0.49} & 64.15 & \red{54.22}~{\small $\pm$0.35} \\
 & NamaaMrTydi & 80.83 & \red{73.65}~{\small $\pm$0.10} & 70.94 & \red{67.35}~{\small $\pm$0.07} & \textbf{88.62} & \red{\textbf{87.11}}~{\small $\pm$0.05} & \underline{88.52} & \red{80.36}~{\small $\pm$0.07} & 79.50 & \red{78.03}~{\small $\pm$0.10} \\
\midrule
\multirow{2}{*}{\rotatebox[origin=c]{90}{\textbf{RT}}} & TwitterHjerne & 55.73 & \green{55.87}~{\small $\pm$1.01} & 58.26 & \red{57.77}~{\small $\pm$1.00} & 69.44 & \red{68.18}~{\small $\pm$0.22} & 72.61 & \red{70.95}~{\small $\pm$0.75} & 71.76 & \red{67.74}~{\small $\pm$0.60} \\
 & Ko-StratQA & 72.04 & \red{67.20}~{\small $\pm$0.64} & 52.96 & \green{55.85}~{\small $\pm$2.14} & 75.05 & \red{71.16}~{\small $\pm$1.15} & 75.42 & \red{66.41}~{\small $\pm$1.00} & 79.42 & \red{74.29}~{\small $\pm$0.56} \\
\midrule
\rotatebox[origin=c]{90}{\textbf{STR}} & SemRel2024 & 54.16 & \red{51.57}~{\small $\pm$0.16} & 50.92 & \red{50.37}~{\small $\pm$0.12} & \underline{55.99} & \red{\textbf{54.45}}~{\small $\pm$0.17} & 54.24 & \red{52.50}~{\small $\pm$0.08} & 55.41 & \red{\underline{54.33}}~{\small $\pm$0.28} \\
\midrule
\multirow{2}{*}{\rotatebox[origin=c]{90}{\textbf{STS}}} & STS17 & 66.01 & \red{62.92}~{\small $\pm$0.09} & 83.46 & \red{78.12}~{\small $\pm$0.14} & 81.42 & \red{76.13}~{\small $\pm$0.13} & 72.82 & \red{70.24}~{\small $\pm$0.12} & 80.22 & \red{74.69}~{\small $\pm$0.10} \\
 & IndicCrossling. & 43.88 & \red{32.83}~{\small $\pm$0.29} & 32.77 & \green{35.44}~{\small $\pm$0.34} & 40.74 & \red{35.70}~{\small $\pm$0.33} & 43.16 & \red{37.15}~{\small $\pm$0.18} & 50.96 & \red{\textbf{43.47}}~{\small $\pm$0.26} \\
\midrule
 & \textbf{Average} & 58.87 & \red{55.23}~{\small $\pm$0.48} & 55.52 & \red{55.50}~{\small $\pm$0.42} & 62.40 & \red{60.33}~{\small $\pm$0.29} & 61.65 & \red{58.04}~{\small $\pm$0.34} & 62.99 & \red{59.97}~{\small $\pm$0.29} \\
\bottomrule
\end{tabularx}
\caption{
\textcolor{multiorange}{\textbf{Multilingual (part 1 of 2)}} embedding model performance on original data and axis-balanced HTEB scores based on 3 runs with $\pm$ denoting the standard deviation (in \%). \textbf{Bold}: Row-wise best score across all table parts. \underline{Underline}: Row-wise second-best score across all table parts. \green{Green}: Performance increase. \red{Red}: Performance decrease. Task abbreviations: CLS = Classification, CLU = Clustering, PC = Pair Classification, RR = Reranking, RT = Retrieval, STR = Semantic Textual Relatedness, STS = Semantic Textual Similarity.
}
\label{tab:multilingual_results_per_dataset_1}
\end{table*}

\begin{table*}[ht]
\small
\centering
\begin{tabularx}{\textwidth}{
  >{\centering\arraybackslash}p{0.03cm}
  >{\raggedright\arraybackslash}p{1.9cm}
  >{\centering\arraybackslash}X
  @{\hspace{0pt}}
  >{\centering\arraybackslash}X
  @{\hspace{8pt}}
  >{\centering\arraybackslash}X
  @{\hspace{0pt}}
  >{\centering\arraybackslash}X
  @{\hspace{8pt}}
  >{\centering\arraybackslash}X
  @{\hspace{0pt}}
  >{\centering\arraybackslash}X
  @{\hspace{8pt}}
  >{\centering\arraybackslash}X
  @{\hspace{0pt}}
  >{\centering\arraybackslash}X
}
\toprule
 &  & \multicolumn{2}{c}{\textbf{Jina-Emb-v5-}} & \multicolumn{2}{c}{\textbf{F2LLM-}} & \multicolumn{2}{c}{\textbf{Llama-Emb-}} & \multicolumn{2}{c}{\textbf{Qwen3-}} \\
 &  & \multicolumn{2}{c}{\textbf{Text-Small}} & \multicolumn{2}{c}{\textbf{v2-4B}} & \multicolumn{2}{c}{\textbf{Nemotron-8B}} & \multicolumn{2}{c}{\textbf{Embedding-8B}} \\
\cmidrule(l{2pt}r{2pt}){3-4}\cmidrule(l{2pt}r{2pt}){5-6}\cmidrule(l{2pt}r{2pt}){7-8}\cmidrule(l{2pt}r{2pt}){9-10}
 & \textbf{Dataset} & \textbf{Orig.} & \textbf{HTEB} & \textbf{Orig.} & \textbf{HTEB} & \textbf{Orig.} & \textbf{HTEB} & \textbf{Orig.} & \textbf{HTEB} \\
\midrule
\multirow{2}{*}{\rotatebox[origin=c]{90}{\textbf{CLS}}} & IsiZuluNews & 39.89 & \red{39.84}~{\small $\pm$0.18} & \textbf{49.60} & \red{\textbf{49.22}}~{\small $\pm$0.32} & 41.76 & \green{43.52}~{\small $\pm$0.22} & 43.88 & \green{45.86}~{\small $\pm$0.26} \\
 & SentimentHindi & 70.26 & \green{73.32}~{\small $\pm$0.48} & 84.42 & \red{83.05}~{\small $\pm$0.13} & 81.54 & \green{82.20}~{\small $\pm$0.06} & \textbf{87.35} & \red{\textbf{85.91}}~{\small $\pm$0.13} \\
\midrule
\multirow{2}{*}{\rotatebox[origin=c]{90}{\textbf{CLU}}} & MasakhaNEWS & \textbf{43.36} & \green{\textbf{44.09}}~{\small $\pm$0.42} & \underline{42.60} & \green{\underline{43.72}}~{\small $\pm$0.46} & 28.89 & \red{26.89}~{\small $\pm$0.43} & 28.40 & \red{28.23}~{\small $\pm$0.64} \\
 & PlscS2S.v2 & \textbf{43.56} & \green{\textbf{43.62}}~{\small $\pm$0.12} & \underline{43.41} & \red{\underline{42.32}}~{\small $\pm$0.27} & 38.16 & \red{36.90}~{\small $\pm$0.22} & 40.94 & \red{36.88}~{\small $\pm$0.37} \\
\midrule
\multirow{2}{*}{\rotatebox[origin=c]{90}{\textbf{PC}}} & RTE3 & \textbf{89.63} & \red{\textbf{89.45}}~{\small $\pm$0.08} & 87.88 & \red{87.83}~{\small $\pm$0.11} & 87.36 & \red{87.25}~{\small $\pm$0.17} & 87.35 & \red{87.14}~{\small $\pm$0.13} \\
 & IndonLI & \textbf{59.96} & \red{\textbf{59.29}}~{\small $\pm$0.16} & 55.63 & \green{56.13}~{\small $\pm$0.29} & 49.17 & \green{52.13}~{\small $\pm$0.18} & 53.86 & \green{55.22}~{\small $\pm$0.15} \\
\midrule
\multirow{2}{*}{\rotatebox[origin=c]{90}{\textbf{RR}}} & VoyageMMarco & \underline{68.56} & \red{61.10}~{\small $\pm$0.44} & \textbf{81.89} & \red{\textbf{74.57}}~{\small $\pm$0.24} & 55.81 & \red{52.29}~{\small $\pm$0.49} & 66.18 & \red{\underline{61.72}}~{\small $\pm$0.29} \\
 & NamaaMrTydi & 74.84 & \green{75.36}~{\small $\pm$0.06} & 76.63 & \green{\underline{81.28}}~{\small $\pm$0.12} & 64.47 & \green{68.95}~{\small $\pm$0.17} & 75.70 & \red{75.00}~{\small $\pm$0.84} \\
\midrule
\multirow{2}{*}{\rotatebox[origin=c]{90}{\textbf{RT}}} & TwitterHjerne & 71.54 & \red{70.77}~{\small $\pm$0.48} & \textbf{81.91} & \red{\textbf{78.90}}~{\small $\pm$0.53} & 48.19 & \green{53.67}~{\small $\pm$1.10} & \underline{78.73} & \red{\underline{78.52}}~{\small $\pm$0.26} \\
 & Ko-StratQA & 80.99 & \red{78.18}~{\small $\pm$0.36} & \underline{83.57} & \red{\textbf{81.06}}~{\small $\pm$0.24} & 80.87 & \red{73.40}~{\small $\pm$0.87} & \textbf{83.64} & \red{\underline{80.41}}~{\small $\pm$0.28} \\
\midrule
\rotatebox[origin=c]{90}{\textbf{STR}} & SemRel2024 & 53.99 & \red{52.83}~{\small $\pm$0.11} & \textbf{56.41} & \red{54.30}~{\small $\pm$0.25} & 51.63 & \red{50.78}~{\small $\pm$0.18} & 55.06 & \red{53.07}~{\small $\pm$0.27} \\
\midrule
\multirow{2}{*}{\rotatebox[origin=c]{90}{\textbf{STS}}} & STS17 & \textbf{87.92} & \red{\textbf{82.44}}~{\small $\pm$0.12} & \underline{85.30} & \red{\underline{79.85}}~{\small $\pm$0.08} & 77.42 & \red{74.45}~{\small $\pm$0.25} & 83.35 & \red{77.10}~{\small $\pm$0.10} \\
 & IndicCrossling. & 46.28 & \red{40.65}~{\small $\pm$0.24} & \textbf{53.41} & \red{40.33}~{\small $\pm$0.41} & 27.52 & \green{31.25}~{\small $\pm$0.34} & \underline{53.13} & \red{\underline{43.17}}~{\small $\pm$0.13} \\
\midrule
 & \textbf{Average} & 63.91 & \red{\underline{62.38}}~{\small $\pm$0.25} & \textbf{67.90} & \red{\textbf{65.58}}~{\small $\pm$0.27} & 56.37 & \green{56.44}~{\small $\pm$0.36} & \underline{64.43} & \red{62.17}~{\small $\pm$0.30} \\
\bottomrule
\end{tabularx}
\caption{
\textcolor{multiorange}{\textbf{Multilingual (part 2 of 2)}} embedding model performance on original data and axis-balanced HTEB scores based on 3 runs with $\pm$ denoting the standard deviation (in \%). \textbf{Bold}: Row-wise best score across all table parts. \underline{Underline}: Row-wise second-best score across all table parts. \green{Green}: Performance increase. \red{Red}: Performance decrease. Task abbreviations: CLS = Classification, CLU = Clustering, PC = Pair Classification, RR = Reranking, RT = Retrieval, STR = Semantic Textual Relatedness, STS = Semantic Textual Similarity.
}
\label{tab:multilingual_results_per_dataset_2}
\end{table*}

\section{Sensitivity to the Transformation Model Choice}
\label{app_gen_model_confounding}

Relying on Gemma-3-27B-int4-AWQ as the only transformation model risks conflating general transformation effects with Gemma-specific output characteristics. We therefore repeat the evaluation on the 19 English datasets using Qwen3-8B-AWQ as an alternative transformation model, applied to three embedding models spanning size and family: All-MPNet-Base-v2 (small, classical baseline), Jina-Embeddings-v5-Small (small, recent), and Qwen3-Embedding-8B (large, instruction-tuned). However, Gemma-3 was selected because it achieved lower error rates and higher LLM-judge scores in our transformation-model selection pipeline, whereas Qwen3's transformations were not validated through human evaluation. We therefore use Qwen3-8B-AWQ only as a sensitivity check for transformation-model choice, keeping Gemma-3-27B-int4-AWQ as the main HTEB transformation model; Qwen3-based results should be interpreted with caution.

The overall disruption relative to original scores remains comparable across transformation models (Gemma-3: -7.89, Qwen3: -6.82); see \autoref{tab:qwen3_gemma3_eng_full_per_transformation}. The per-axis pattern is also broadly similar, although Qwen3 is less disruptive on Lexical/Stylistic (-3.28 vs. Gemma's -4.25) and Length (-2.68 vs. Gemma's -5.47), but more disruptive on Language (-14.51 vs. Gemma's -13.94). Individual transformations vary more, especially Summarisation (Gemma-3: -6.36, Qwen3: -1.32). A plausible explanation is that Qwen3's summaries deviate less in length from the original text and therefore disrupt embeddings less: \autoref{tab:qwen3-gemma3-transform-lengths} shows that they retain on average 63\% of the original word count, compared with 37\% for Gemma-3. This pattern is not fully consistent: Gemma-3's expansions are 4.32 times the original length on average compared with 1.93 for Qwen3, yet the disruption gap for Expansion is only about one percentage point. Overall, these comparisons suggest that the main findings are not purely Gemma-specific, while absolute HTEB scores remain conditioned on the chosen transformation model and the quality of its transformations.

\begin{table*}[t]
\centering
\small
\begin{tabular}{l c c c c c c}
\toprule
 & & \multicolumn{2}{c}{\textbf{HTEB Score}} & \multicolumn{2}{c}{\textbf{$\Delta$ vs.\ Original}} & \textbf{$\Delta$ Qwen3 $-$} \\
\cmidrule(lr){3-4} \cmidrule(lr){5-6}
\textbf{Transformation} & \textbf{Original} & \textbf{Gemma-3} & \textbf{Qwen3} & \textbf{Gemma-3} & \textbf{Qwen3} & \textbf{$\Delta$ Gemma-3} \\
\midrule
\textbf{Lexical/Stylistic} & \textbf{70.45} & \textbf{66.20} & \textbf{67.17} & \textbf{-4.25} & \textbf{-3.28} & \textbf{+0.97} \\
\quad Paraphrasing & 70.45 & 66.50 & 68.14 & -3.95 & -2.32 & +1.63 \\
\quad Backtranslation & 70.45 & 67.49 & 66.05 & -2.96 & -4.40 & -1.44 \\
\quad Style Change & 70.45 & 64.61 & 67.33 & -5.85 & -3.12 & +2.72 \\
\midrule
\textbf{Length} & \textbf{70.45} & \textbf{64.98} & \textbf{67.77} & \textbf{-5.47} & \textbf{-2.68} & \textbf{+2.79} \\
\quad Expansion & 70.45 & 66.55 & 67.72 & -3.90 & -2.73 & +1.17 \\
\quad Summarisation & 70.45 & 64.09 & 69.13 & -6.36 & -1.32 & +5.04 \\
\quad Summ.\ Expansion & 70.45 & 64.30 & 66.47 & -6.16 & -3.98 & +2.17 \\
\midrule
\textbf{Language} & \textbf{70.45} & \textbf{56.52} & \textbf{55.95} & \textbf{-13.94} & \textbf{-14.51} & \textbf{-0.57} \\
\quad Translation & 70.45 & 59.11 & 58.28 & -11.34 & -12.18 & -0.83 \\
\quad Cross-Translation & 70.45 & 53.92 & 53.62 & -16.53 & -16.84 & -0.30 \\
\midrule
\textbf{Overall} & \textbf{70.45} & \textbf{62.57} & \textbf{63.63} & \textbf{-7.89} & \textbf{-6.82} & \textbf{+1.07} \\
\bottomrule
\end{tabular}
\caption{HTEB scores by transformation under Qwen3-8B-AWQ vs.\ Gemma-3-27B generation, averaged across the 19 English HTEB datasets, 3 embedding models (All-MPNet-Base-v2, Jina-Emb-v5-Small, Qwen3-Embedding-8B), and 3 runs. Original scores come from the latest Gemma-3 result within each embedding-model and dataset group and are shared by both generator conditions. Overall averages the three robustness-axis subtotals equally. $\Delta$ columns are differences between HTEB and the Original score; the right-hand column shows the difference of the differences between HTEB and original scores (Qwen3 minus Gemma-3).}
\label{tab:qwen3_gemma3_eng_full_per_transformation}
\end{table*}

\begin{table*}[t]
\centering
\small
\begin{tabular}{l c c c c c c c}
\toprule
 & & \multicolumn{2}{c}{\textbf{Transformed (mean words)}} & \multicolumn{2}{c}{\textbf{Length ratio vs.\ Original}} & \multicolumn{2}{c}{\textbf{\% empty}} \\
\cmidrule(lr){3-4} \cmidrule(lr){5-6} \cmidrule(lr){7-8}
\textbf{Transformation} & \textbf{Original} & \textbf{Gemma-3} & \textbf{Qwen3} & \textbf{Gemma-3} & \textbf{Qwen3} & \textbf{Gemma-3} & \textbf{Qwen3} \\
\midrule
\textbf{Lexical/Stylistic} & \textbf{23.5} & \textbf{23.0} & \textbf{23.4} & \textbf{0.98} & \textbf{1.00} & \textbf{0.0} & \textbf{0.0} \\
\quad Paraphrasing & 23.5 & 21.8 & 22.7 & 0.93 & 0.97 & 0.0 & 0.0 \\
\quad Backtranslation & 23.5 & 22.0 & 22.9 & 0.94 & 0.97 & 0.0 & 0.1 \\
\quad Style Change & 23.5 & 25.2 & 24.6 & 1.07 & 1.05 & 0.0 & 0.0 \\
\midrule
\textbf{Length} & \textbf{23.5} & \textbf{45.2} & \textbf{28.5} & \textbf{1.92} & \textbf{1.21} & \textbf{0.0} & \textbf{0.0} \\
\quad Expansion & 23.5 & 101.6 & 45.4 & 4.32 & 1.93 & 0.0 & 0.1 \\
\quad Summarisation & 23.5 & 8.7 & 14.7 & 0.37 & 0.63 & 0.0 & 0.0 \\
\quad Summ.\ Expansion & 23.5 & 25.4 & 25.4 & 1.08 & 1.08 & 0.0 & 0.1 \\
\midrule
\textbf{Language} & \textbf{23.5} & \textbf{22.4} & \textbf{22.7} & \textbf{0.95} & \textbf{0.97} & \textbf{0.0} & \textbf{0.2} \\
\quad Translation & 23.5 & 22.6 & 22.8 & 0.96 & 0.97 & 0.0 & 0.3 \\
\quad Cross-Translation & 23.5 & 22.3 & 22.6 & 0.95 & 0.96 & 0.0 & 0.1 \\
\midrule
\textbf{Overall} & \textbf{23.5} & \textbf{30.2} & \textbf{24.9} & \textbf{1.29} & \textbf{1.06} & \textbf{0.0} & \textbf{0.1} \\
\bottomrule
\end{tabular}
\caption{Word-count comparison between Qwen3-8B-AWQ and Gemma-3-27B transformations, averaged over the 19 English HTEB datasets and three runs. \textit{Original} reports the mean word count of the source text; the \textit{Transformed} columns report the same statistic for the LLM-generated text under each generator. The overall row averages the three transformation-axis rows equally. \textit{Length ratio vs.\ Original} is computed from the displayed averaged word counts. \textit{\% empty} reports the percentage of items whose latest cached transformation is an empty string after all retries.}
\label{tab:qwen3-gemma3-transform-lengths}
\end{table*}

\end{document}